\documentclass{article}
\usepackage{geometry}

\usepackage[T1]{fontenc}
\usepackage[latin9]{inputenc}
\usepackage{bm}
\usepackage{amsmath,amsthm,mathtools}
\usepackage{amssymb}

\usepackage{natbib}

\usepackage[unicode=true,
 bookmarks=false,
 breaklinks=false,pdfborder={0 0 1},colorlinks=false]
 {hyperref}

\hypersetup{colorlinks,citecolor=blue,filecolor=blue,linkcolor=blue,urlcolor=blue}

\makeatletter
\usepackage{comment}

\usepackage{booktabs}

\usepackage{graphicx}

\usepackage[linesnumbered,ruled,vlined]{algorithm2e}

\SetCommentSty{mycommfont}
\usepackage[linesnumbered,ruled,vlined]{algorithm2e}

\usepackage{float}
\usepackage{multirow}
 
\usepackage{dsfont}
\usepackage{tcolorbox}

\usepackage{color}
\definecolor{yxc}{RGB}{255,0,0}
\definecolor{ycz}{RGB}{125,0,0}
\definecolor{ytw}{RGB}{255,69,0}
\definecolor{gen}{RGB}{0,0,200}

\allowdisplaybreaks

\DeclareMathOperator{\ind}{\mathds{1}}  % Indicator

\newcommand{\mprob}{\mathbb{P}}

\newcommand{\mymid}{\,|\,}

\newcommand{\Pdata}{p_{\mathsf{data}}}

\title{Faster Diffusion Models via Higher-Order Approximation}

\author{%
Gen Li\thanks{The first two authors contributed equally.}
\thanks{Department of Statistics, Chinese University of Hong Kong.}
\and
Yuchen Zhou\footnotemark[1]
\thanks{Department of Statistics, University of Illinois Urbana-Champaign.} 
\and Yuting Wei\thanks{Department of Statistics and Data Science, the Wharton School, University of Pennsylvania.} 
 \and Yuxin Chen\footnotemark[3]
}
\date{June 2025;~~ Revised: August 2025}

\begin{document}

\theoremstyle{plain} \newtheorem{lemma}{\textbf{Lemma}}\newtheorem{proposition}{\textbf{Proposition}}\newtheorem{theorem}{\textbf{Theorem}}

\theoremstyle{assumption}\newtheorem{assumption}{\textbf{Assumption}}
\theoremstyle{remark}\newtheorem{remark}{\textbf{Remark}}

\maketitle

\begin{abstract}
	In this paper, we explore provable acceleration of diffusion models without any additional retraining. Focusing on the task of approximating a target data distribution in $\mathbb{R}^d$ to within $\varepsilon$ total-variation distance, we propose a principled,  training-free sampling algorithm that requires only the order of
	$$ d^{1+2/K} \varepsilon^{-1/K} $$
	score function evaluations (up to log factor) in the presence of accurate scores, where $K>0$ is an arbitrary fixed integer. This result applies to a broad class of target data distributions, without the need for assumptions such as smoothness or log-concavity. Our theory is robust vis-a-vis inexact score estimation, degrading gracefully as the score estimation error increases --- without demanding higher-order smoothness on the score estimates as assumed in previous work.  The proposed algorithm draws insight from high-order ODE solvers, leveraging high-order Lagrange interpolation and successive refinement to approximate the integral derived from the probability flow ODE. More broadly, our work develops a theoretical framework towards understanding the efficacy of high-order methods for accelerated sampling. 
\end{abstract}

%These findings unlock the theoretical potential of training-free acceleration, unveiling the acceleration effect for a broad spectrum of data distributions.  


\tableofcontents

\section{Introduction}

Diffusion models, which originally drew inspiration from nonequilibrium thermodynamics \citep{sohl2015deep}, 
have emerged as a powerful driving force in the realm of generative modeling, reshaping the landscape of contemporary generative artificial intelligence \citep{croitoru2023diffusion,yang2023diffusion}.   Yet, despite their incredible sample quality and enhanced stability,  diffusion models face the challenge of slow data generation compared to alternatives like Generative Adversarial Networks (GANs) \citep{goodfellow2020generative}  and  Variational Autoencoders (VAEs) \citep{kingma2013auto}. 
In comparison to GANs and VAEs that generate data in a single step (i.e., through a single forward pass of a neural network), diffusion models require a number of iterations.  Each of these iterations, which oftentimes involves evaluating the pretrained score functions, necessitates at least one pass of a neural network or transformer,  making the entire process more computationally intensive.  In order to fully unleash the capability of diffusion models for real-time data generation,  accelerating their sampling process without compromising sample quality is essential.

\subsection{Diffusion models and score function evaluations}

To set the stage, note that a diffusion model begins with a data-contaminating forward process: 
\begin{subequations}
	\label{eq:forward-process}
	\begin{align}
		(\text{forward process})\qquad X_0 &\sim \Pdata, \quad 
		X_t = \sqrt{1-\beta_{t}}\, X_{t-1} + \sqrt{\beta_t}\, W_t, \quad t = 1, \cdots, T,
	\end{align}
\end{subequations}
where $\Pdata$ indicates the target data distribution in $\mathbb{R}^d$, 
$T$ is the total number of steps, the $W_t$'s are noise vectors independently drawn from $\mathcal{N}(0,I_d)$, and $\{\beta_t\}_{1\leq t\leq T}\subset (0,1) $ denotes some predetermined sequence that governs the variance of the additive noise at each step. In essence, 
this forward process first draws a sample $X_0$ from the distribution of interest, and progressively corrupts it by injecting independent Gaussian noise. When $T$ is sufficiently large, it is often the case that the distribution of $X_T$ becomes exceedingly close to $\mathcal{N}(0,I_d)$. The central task of diffusion generative modeling is to learn a time-reversal of the above forward process, enabling us to start with a pure noise distribution (e.g., $\mathcal{N}(0,I_d)$) and  iteratively turn it back into the target distribution $\Pdata$. 
Mathematically, this can be described as the construction of a backward process 
\begin{align*}
	(\text{backward process}) \qquad 
	Y_T &\sim \mathcal{N}(0,I_d), \qquad
	Y_T \rightarrow Y_{T-1} \rightarrow \dots \rightarrow Y_1 \rightarrow Y_0
\end{align*}
that ensures the distributions of $Y_t$ and $X_t$ stay close for each step $t$, namely, $Y_t \overset{\mathrm{d}}{\approx} X_t$.

As already demonstrated by classical stochastic differential equation (SDE) literature \citep{anderson1982reverse,haussmann1986time}, time-reversal of the forward process \eqref{eq:forward-process} is made possible through the use of the (Stein) score functions $\{\nabla \log p_{X_t}(\cdot)\}$. 
Consequently, 
diffusion generative modeling involves an initial, extensive training phase to learn these score functions, which produces a collection of pretrained score estimates that can be readily invoked in data generation.  
For instance,  two prominent score-based diffusion model paradigms --- DDPM \citep{ho2020denoising} and the probability flow ODE (or DDIM)  \citep{song2020score,song2020denoising} --- both rely upon these pretrained scores for iterative sampling. In practice, however, these mainstream approaches remain relatively slow due to their large iteration complexities; for example, generating high-quality samples with DDPM can take hundreds or even thousands of iterations, with each iteration requiring the evaluation of a neural network or a transformer.  
All this underscores the need to speed up the iterative sampling process while maintaining the outstanding sample quality  diffusion models are known for.

\subsection{Training-free acceleration}

In this work, we explore principled,  training-free acceleration of diffusion models \citep{lu2022dpm,zheng2023dpm,zhao2024unipc,zhangfast,jolicoeur2021gotta,xue2024sa}.  
%
%which seek to expedite the data generation process directly using the obtained score functions.
%, as opposed to the training-based approach such as distillation or consistency models that require additional (often extensive) training processes \citep{song2023consistency,li2024towards}. 
In a nutshell, the training-free approach wraps around and makes more efficient use of the pretrained score functions to expedite data generation, which stands in sharp contrast to the training-based approach --- such as distillation or consistency models --- that performs another round of resource-intensive training  \citep{luhman2021knowledge,salimansprogressive,meng2023distillation,song2023consistency,li2024towards,nichol2021improved}.  Remarkably, training-free acceleration has already seen widespread adoption in practice, 
with prominent examples being the DPM-solver \citep{luhman2021knowledge}, DPM-Solver++ \citep{lu2025dpm}, and UNiPC \citep{zhao2023unipc} that have achieved dramatic empirical speedups compared to the original probability flow ODE algorithm.

Nevertheless, the theoretical underpinnings for accelerated diffusion models remain largely elusive. It was not until recently that researchers began developing non-asymptotic analysis frameworks for the training-free approach,  %\citep{li2024accelerating,wu2024stochastic,huang2024convergence,li2024provable}, 
although significant technical hurdles remain that limit the theoretical performance. More concretely, consider training-free acceleration methods that come with provable guarantees, and suppose that we would like to generate a sample whose distribution is $\varepsilon$-close in total-variation (TV) distance to the target  distribution of interest. \citet{li2024accelerating} proposed a deterministic accelerated approach, akin to the second-order ODE solver, achieving an iteration complexity of $\widetilde{O}(d^3 / \sqrt{\varepsilon})$.\footnote{Throughout this paper, 
	the standard notation $f(d,\varepsilon^{-1}) = O (g(d,\varepsilon^{-1}))$ or $f(d,\varepsilon^{-1}) \lesssim g(d,\varepsilon^{-1})$ means that there exists a numerical constant $c_1 >
	0$ such that $|f(d,\varepsilon^{-1})| \leq c_1|g(d,\varepsilon^{-1})|$; 
	$f(d,\varepsilon^{-1}) \gtrsim g(d,\varepsilon^{-1})$ means that $g(d,\varepsilon^{-1}) \lesssim f(d,\varepsilon^{-1})$; 
	and $f(d,\varepsilon^{-1}) \asymp g(d,\varepsilon^{-1})$ means that both $f(d,\varepsilon^{-1}) \lesssim g(d,\varepsilon^{-1})$ and $f(d,\varepsilon^{-1}) \gtrsim g(d,\varepsilon^{-1})$ hold true. The notation $\widetilde{O}(\cdot)$ is defined analogously to $O(\cdot)$ except that the log dependency is hidden. } \citet{huang2024convergence} analyzed the $p$-th order Runge-Kutta method for solving the probability flow ODE, 
and established an iteration complexity of $\widetilde{O}\big((LDd)^{1+1/p}/\varepsilon^{1/p}\big)$ for any constant order $p$, where $D$ represents the radius of the data support and $L$ bounds certain high-order derivatives of the score estimates. This represents a clear improvement over \citet{li2024accelerating} in terms of the $\varepsilon$ dependency, although the stringent requirements on $L$ and $D$ limit the range of data distributions it can accommodate without compromising the acceleration effect.  
Meanwhile, stochastic training-free methods have been proposed and analyzed in recent literature as well \citep{li2024accelerating,wu2024stochastic,li2024provable}, with the state-of-the-art results achieving an iteration complexity of $\widetilde{O}(d^{5/4} / \sqrt{\varepsilon})$ \citep{li2024provable}. While such stochastic acceleration methods often come with theoretical guarantees under minimal assumptions, their $\varepsilon$ dependency remains considerably worse than that of the deterministic counterpart.

In summary, while the strand of work outlined above has made significant progress towards enriching the theoretical foundation for training-free acceleration, it still leaves room for improvement, given that all of these prior results fell short of optimality. 

%This calls for further investigation to fully unlock the theoretical potential of the training-free approach. 

%although their theoretical guarantees are in general worse than the deterministic counterpart . 

\subsection{Main contributions} 

% {\color{red} Do we need to mention that we provide a theoretical framework for high-order samplers?}

Motivated by the limitations of prior work as outlined above, the present paper seeks to strengthen the theoretical and algorithmic development for training-free acceleration of diffusion models.  More concretely, we propose a novel training-free acceleration algorithm aimed at solving the probability flow ODE. Falling under the category of deterministic samplers (except for its initialization), the proposed algorithm exploits high-order Lagrange polynomial interpolation, in conjunction with suitable successive refinement, to approximately calculate the integral derived from the probability flow ODE. We prove that, to achieve $\varepsilon$-precision in total variation, it takes our algorithm $\widetilde{O}(\frac{ d^{1+2/K} }{  \varepsilon^{1/K} })$ iterations, or equivalently, $$\widetilde{O}\bigg(\frac{ d^{1+2/K} }{  \varepsilon^{1/K} }\bigg) \text{ score function evaluations}$$
in the presence of accurate scores. Our theory does not require any smoothness or log-concavity assumptions on the target distribution, making it broadly applicable. When only inexact score estimates are available, we demonstrate that the sampling quality degrades gracefully as the score estimation error and the associated Jacobian errors increase. Our results outperform 
\citet{li2024accelerating,wu2024stochastic,li2024provable} by offering better scaling in both $\varepsilon$ and $d$, and improve upon \citet{huang2024convergence} by accommodating a broader set of nonsmooth distributions without assuming higher-order smoothness.  More broadly, our theory develops a theoretical framework that proves effective for analyzing higher-order accelerated samplers.

%In this paper, we introduce a novel acceleration method for deterministic samplers. Motivated by the Taylor expansion, at the $t$-th step, we make use of multiple score estimates $\{s_{\tau_{t,i}}(\cdot)\}$ for different $\tau_{t,i}$ to obtain more accurate estimators for the score functions \{$s_{\tau}^\star$\} (whose formal definition is deferred to \eqref{eq:score_form}). To achieve $\varepsilon$-accuracy in total variation, we prove that $\widetilde{O}(\frac{ d^{1+2/K} }{  \varepsilon^{1/K} })$ iterations is sufficient as long as the estimates of the score functions and the corresponding Jacobian matrices are accurate.

\subsection{Other related work}
\label{sec:related-work}

The development of convergence theory for diffusion models --- particularly DDPM and DDIM --- has received much attention during the past few years.  Partial examples include \cite{chen2022sampling,liu2022let,lee2023convergence,chen2023improved,chen2023restoration,li2023towards,cheng2023convergence,chen2023probability,benton2024nearly,tang2024score,liang2024non,li2025dimension,jiao2024instance,cai2025minimax,li2024unified,gao2024convergence,li2024improved,li2024d}; see also the references therein.  The convergence analysis for the DDPM was carried out in \citet{chen2022sampling} by means of Girsanov's theorem, and this analysis framework was subsequently improved to accommodate nonsmooth data distributions \citet{lee2023convergence,chen2023improved}, unveiling an iteration complexity with nearly linear $d$ dependency \citet{benton2024nearly,li2024d}. When it comes to the DDIM sampler or the probability flow ODE, the convergence analysis was originally provided in  \citet{chen2023restoration}, with a set of subsequent work devoted to sharpening the iteration complexity \citep{li2024sharp,huang2024convergence,li2024unified,liang2025low}. 
In addition, a recent line of work \citet{li2024adapting,huang2024denoising,azangulov2024convergence,potaptchik2024linear,liang2025low,tang2025adaptivity}  explored how DDPM and DDIM adapt to unknown low-dimensional structures underlying the target data distribution, leading to substantially improved theoretical guarantees. 
Finally, rather than reducing the number of score evaluations, a strand of recent work (e.g., \citet{shih2023parallel}) accelerates diffusion sampling by running denoising steps in parallel. Interestingly, our proposed method is also amenable to parallel computing to some extent, although this is beyond the main goal of the current paper.  

\section{Preliminaries}
\label{sec:preliminary}

Before proceeding to the description of our algorithm, let us start by gathering several preliminary facts.

\paragraph{The forward process.}
Recall the forward process~\eqref{eq:forward-process}. By taking 
\begin{align}
\label{eq:overline-alpha}
\alpha_t \coloneqq 1- \beta_t 
\qquad \text{and}\qquad 
\overline{\alpha}_t \coloneqq \prod_{i=1}^t \alpha_i, 
%\qquad \text{and} \qquad  
\end{align}
we see that \eqref{eq:forward-process} admits the following distributional characterization:  
\begin{equation}
    \label{eq:forward-alt}
    X_t = \sqrt{\overline{\alpha}_t}X_0 + \sqrt{1 - \overline{\alpha}_t}\,\overline{W}_t 
    \qquad \text{  with  } \overline{W}_t \sim \mathcal{N}(0,I_d).
\end{equation}

Moreover, we find it useful to introduce a continuous-time generalization $(\overline{X}_{\tau})$ of the above process. 
Specifically, for every $\tau \in [0,1]$, construct $\overline{X}_{\tau}$ such that 
\begin{align}
\label{defn:X-tau-process}
    \overline{X}_\tau = \sqrt{1-\tau}X_0 + \sqrt{\tau}Z
    \qquad \text{with }~ X_0 \sim \Pdata ~\text{ and }~Z\sim \mathcal{N}(0,I_d), 
\end{align}
where $X_0$ and $Z$ are independently generated. Clearly, this construction obeys 
\begin{align}
\overline{X}_{1-\overline{\alpha}_t} \overset{\mathrm{d}}{=} X_t.
\label{eq:X-alphat-Xt}
\end{align}
In the sequel, 
we shall often refer to $\tau$ as the time variable too, as long as it is clear from the context.  
We will develop and describe our algorithm largely based on this continuous-time forward process~\eqref{defn:X-tau-process}.

\paragraph{The score function and Tweedie's formula.} 
As mentioned previously, a key object that plays a pivotal role in the sampling procedure is the (Stein) score function, defined as the gradient of the log marginal density of the forward process.  To be precise, the score function associated with $(\overline{X}_{\tau})$ is defined and denoted by
\begin{align}
    \label{eq:defn-score}
    s^\star_{\tau}(X) \coloneqq \nabla \log p_{\overline{X}_{\tau}}(X)
\end{align}
for each $\tau \in [0,1]$. 
The celebrated Tweedie formula \citep{efron2011tweedie} tells us that 
\begin{align} %s_{\tau}^{\star}(x)=\frac{\sqrt{1-\tau}}{\tau}\mathbb{E}\big[X_{0}\mid\overline{X}_{\tau}=x\big]-x
s_{\tau}^{\star}(x)=-\frac{1}{\tau}\Big\{x-\sqrt{1-\tau}\,\mathbb{E}\big[X_{0}\mid\overline{X}_{\tau}=x\big]\Big\}
\end{align}
for any $x\in \mathbb{R}^d$, 
which readily implies that
\begin{align}
\label{eq:score_form}
s_{\tau}^{\star}(\sqrt{1-\tau}x) &= -\frac{\sqrt{1-\tau}}{\tau}\int_{x_0} p_{X_0 \mymid \overline{X}_\tau}(x_0 \mymid \sqrt{1-\tau}x)(x - x_0)\mathrm{d}x_0. 
\end{align}

\paragraph{The probability flow ODE.}
As has been shown by \citet{song2020score}, 
there exists an ordinary differential equation (ODE), called the probability flow ODE, that is able to reverse the forward process \eqref{defn:X-tau-process} (in the sense of yielding matching marginal distributions). Note that the precise form of the probability flow ODE depends on the parameterization of the forward process (e.g., $(X_t)$ and $(\overline{X}_{\tau})$ are associated with different, although closely related, ODEs). In particular,  the probability flow ODE associated with process \eqref{defn:X-tau-process} takes the following form: 
\begin{align} \label{eq:ODE}
	\mathrm{d} \frac{Y_\tau^{\mathsf{ode}}}{\sqrt{1-\tau}} = -\frac{1}{2(1-\tau)^{3/2}}s_{\tau}^{\star}(Y_\tau^{\mathsf{ode}}) \mathrm{d} \tau; 
\end{align}
see, e.g., \citet[Eqn.~(20)]{li2024accelerating}.  
When setting $Y_{\tau_2}^{\mathsf{ode}}=\overline{X}_{\tau_2}$ and running \eqref{eq:ODE} backward from $\tau_2$ to $\tau_1$, one has $Y_{\tau_1}^{\mathsf{ode}} \overset{\mathrm{d}}{=}\overline{X}_{\tau_1}$, allowing one to transport the distribution of $\overline{X}_{\tau_2}$ to that of $\overline{X}_{\tau_1}$.

\paragraph{Notation.}
%\yxc{Can you add important notation here?}
%The notation $f(d, T) \lesssim g(d, T)$ or $f(d, T) = O(g(d, T))$ means that $|f(d, T)| \leq Cg(d, T)$ holds for some numerical constant $C > 0$; we let $f(d, T) \gtrsim g(d, T)$ indicate that $f(d, T) \geq C|g(d, T)|$ for some numerical constant $C > 0$; $f(n_1, n_2) \asymp g(n_1, n_2)$ means that both $f(d, T) \lesssim g(d, T)$ and $f(d, T) \gtrsim g(d, T)$ hold; 
For two probability measures $P$ and $Q$,  the total-variation distance between them is defined as ${\sf TV}(P,Q) \coloneqq \frac{1}{2}\int|{\rm d}P - {\rm d}Q|$. For any random object $X$, we let $p_X$ denote its probability density function. For any matrix $A$, we denote by $\|A\|$ its spectral norm.

%\begin{equation}
%\mathrm{d}Y_{\tau}^{\mathsf{pf}}=-\frac{1}{2}\big(Y_{\tau}^{\mathsf{pf}}+s_{\tau}^{\star}(Y_{\tau}^{\mathsf{pf}})\big)\mathrm{d}\tau,
%\end{equation}
%
%which obeys $Y_{\tau}^{\mathsf{pf}}\overset{\mathrm{d}}{=} \overline{X}_{\tau}$

% and the following ODE \citep[Eqn.~(20)]{li2024accelerating}
%
%\begin{align} \label{eq:ODE}
%\mathrm{d} \frac{x_\tau^{\star}}{\sqrt{1-\tau}} = -\frac{1}{2(1-\tau)^{3/2}}s_{\tau}^{\star}(x_\tau^{\star}) \mathrm{d} \tau
%\end{align}
%can transport the distribution of $X_{\tau_1}$ to the distribution of $X_{\tau_2}$.

% \vspace{1cm}
% \citet{huang2024reverse}

% \paragraph{DDIM.}

\section{Algorithm}
\label{sec:algorithm}

%\subsection{Algorithm description}

In this section, we present the main ideas and detailed procedure of the proposed algorithm. 
At a high level, our algorithm is an iterative procedure that attempts to approximately solve ODE~\eqref{eq:ODE} by approximating the true scores in the integrand with the aid of high-order polynomials.

\paragraph{Approximating an integral.} 
Recall the fact that
$
X_{t} = \overline{X}_{1-\overline{\alpha}_t} $ 
and $
X_{t-1} = \overline{X}_{1-\overline{\alpha}_{t-1}} ,
$
which makes apparent the importance of time points $1-\overline{\alpha}_t$ and $1-\overline{\alpha}_{t-1}$ when generating $Y_{t-1}$. Armed with the probability flow ODE~\eqref{eq:ODE}, an ideal strategy  for step $t-1$ would be to evaluate the following integral: 
\begin{equation}
	\frac{Y_{1-\overline{\alpha}_{t-1}}^{\mathsf{ode}}}{\sqrt{\overline{\alpha}_{t-1}}}=\frac{Y_{1-\overline{\alpha}_{t}}^{\mathsf{ode}}}{\sqrt{\overline{\alpha}_{t}}}- {\displaystyle \int}_{1-\overline{\alpha}_{t}}^{1-\overline{\alpha}_{t-1}}\frac{1}{2(1-\tau)^{3/2}}s_{\tau}^{\star}(Y_{\tau}^{\mathsf{ode}})\mathrm{d}\tau.
	\label{eq:integral-tau1-tau2}
\end{equation}
However, computing \eqref{eq:integral-tau1-tau2} involves evaluation of an infinite number of score functions (i.e., those in the integral), making it computationally intensive. As a result, everything comes down to approximating the integral using a small number of score function evaluations.

\begin{algorithm}[t]
	\caption{HEROISM (\textbf{H}igh-ord\textbf{ER} \textbf{O}de-based d\textbf{I}ffusion \textbf{S}a\textbf{M}pler)} \label{algorithm:high_order}
	\DontPrintSemicolon
	\textbf{input:} $\tau_{t,0} = \tau_{t+1, K-1} = 1 - \overline{\alpha}_t$ for all $1 \leq t \leq T-1$, and $\tau_{T, 0} = 1 - \overline{\alpha}_{T}$. 
	Choose $\tau_{t, i}$ as in \eqref{eqn:brahms-tau-i}.\\
	\textbf{initialization:} 
	%$t = T,\dots, 2$, 
	$ x_{\tau_{T, 0}} \leftarrow Y_{T}\sim \mathcal{N}(0, I_d)$.\\
	\For{$t = T,\dots, 2$} 
	{$x_{\tau_{t, 0}} \leftarrow Y_t$ and $x_{\tau_{t, i}}^{(0)} \leftarrow Y_t$ for all $i = 0, \dots, K-1$\\
		\For {$n = 0, \dots, N-1$}
		{
			%Update $(1-\tau)^{-3/2}s_{\tau}^{(n)}$ via Eqn.~\eqref{eq:score_update} for all $\tau \in [\tau_{t,K-1}, \tau_{t,0}]$\\
			compute $x_{\tau_{t,i}}^{(n+1)}$ via Eqn.~\eqref{eq:update_rule} for all $i = 0, \dots, K-1$
		}
		$Y_{t-1} \leftarrow x_{\tau_{t, K-1}}^{(N)}$
		%\\
		%$t \leftarrow t - 1$
	}
	\textbf{output:} $Y_1$    
\end{algorithm}

\paragraph{High-order polynomials.}
% For any $1 \leq t \leq T$, we set $\tau_{t, 0} = \tau_{t + 1, K-1} = 1 - \overline{\alpha}_{t}$. %for the deterministic sampler and $\tau_{t, 0} = 1 - \overline{\alpha}_{t}, \tau_{t + 1, k-1} = 1 - \overline{\alpha}_{t-1}$ for the stochastic sampler.
% In addition, for any $2 \leq t \leq T$, $0 < i < K - 1$, let 
% \begin{align*}
	% 	\tau_{t, i} = \tau_{t, 0} + \frac{i}{K - 1}(\tau_{t, K - 1} -  \tau_{t, 0}).
	% \end{align*}
% In words, given every $t$, dividing the interval $[1 - \overline{\alpha}_t, 1 - \overline{\alpha}_{t+1})$ into $K-1$ equal intervals, the sequence $\tau_{t,i}$ with $i=1,\ldots, K-2$ takes the value of those end points.  

% \yutingcomment{check the new version:}

In comparison to first-order solvers like DDIM that construct $Y_{t-1}$ solely based on the score estimate $s_{1-\overline{\alpha}_{t}} (\cdot)$, we propose to take advantage of the score functions at $K$ time points --- denoted by $\tau_{t,0},\dots,\tau_{t,K-1}$ --- within the interval $[1 - \overline{\alpha}_{t-1}, 1 - \overline{\alpha}_{t}]$. 
To be precise, 
set 
\begin{subequations}
	\label{eqn:brahms-tau-i}
	\begin{align}
		\tau_{t, 0} = 1 - \overline{\alpha}_{t}
		\qquad \text{and} \qquad 
		\tau_{t, K-1} = 1 - \overline{\alpha}_{t-1}
	\end{align}
	to be the two endpoints of $[1 - \overline{\alpha}_{t-1}, 1 - \overline{\alpha}_{t}]$, 
	and create equi-spaced points within this interval as
	\begin{align}
		\tau_{t, i} = \tau_{t, 0} - \frac{i}{K - 1}(\tau_{t, 0} - \tau_{t, K-1} ),
		\qquad 0< i < K, 
	\end{align}
\end{subequations}
which clearly satisfy $\tau_{t,K-1}< \tau_{t,K-2} < \cdots < \tau_{t,0}.$ 
It then boils down to how to exploit the score functions at these $K$ points to approximately
compute \eqref{eq:integral-tau1-tau2}.

%solve ODE~\eqref{eq:ODE} within this interval $[1 - \overline{\alpha}_{t-1}, 1 - \overline{\alpha}_{t}]$. 

%. Our goal is to approximate the true score function on this interval, given score estimates at those endpoints of the sub-intervals.
%Let us first define those end points which we call $\{\tau_{t,i}\}_{0\leq i\leq K-1, 1\leq t\leq T}$. 
%For any $1 \leq t \leq T-1$, we set 

%In addition, for any $2 \leq t \leq T$, $0 < i \leq K - 1$, let

% In words,  the sequence $\{\tau_{t,i}\}_{0\leq i\leq K-1}$ takes the value of those endpoints of equal sub-intervals of  $[1 - \overline{\alpha}_t, 1 - \overline{\alpha}_{t+1}]$ 
% with $\tau_{t,K-1}< \tau_{t,K-2} < \cdots < \tau_{t,0}.$
% \yutingcomment{it is probably easier to include a figure.}

%Suppose for each $\tau_{t, i}$, the forward process~\eqref{eq:forward-alt} takes value $x_{\tau_{t, i}}$, which we shall elaborate on momentarily. Obtain score estimates at those values. 

Our strategy is to attempt approximation by means of the Lagrange interpolation. 
To be precise, consider the renowned Lagrange basis polynomials as follows
\begin{align}\label{ineq1}
	\psi_i(\tau)
	% = \Phi^{-1}[(\tau - \tau_{t, 0})^j]_{0 \le j < K}
	\coloneqq \frac{\prod_{i': i' \ne i} (\tau - \tau_{t, i'} )}{\prod_{i': i' \ne i} (\tau_{t, i} - \tau_{t, i'} )}, 
	\qquad 0 \le i <  K.
\end{align}
Given a set of points $(\tau_{t,i}, (1-\tau_{t, i})^{-3/2}s_{\tau_{t, i}}(Y^{\mathsf{ode}}_{\tau_{t, i}}))$ for $0 \le i < K$, 
we would like to approximate $\frac{1}{(1-\tau)^{3/2}}s^{\star}_{\tau} (x_{\tau})$ via
the Lagrange interpolating polynomial passing through the above $K$ points, i.e., 
\begin{align}
	\frac{1}{(1-\tau)^{3/2}}s_{\tau}^{\star} (Y^{\mathsf{ode}}_{\tau})
	%= \sum_{0 \le j < k} \phi_j(\tau - \tau_{t, 0})^j
	\approx \sum\nolimits_{0 \le i < K} \psi_i(\tau)
	\frac{s_{\tau_{t, i}}(Y^{\mathsf{ode}}_{\tau_{t, i}})}{(1-\tau_{t, i})^{3/2}},\quad \forall \tau \in [\tau_{t,K-1}, \tau_{t,0}]. \label{eq:score_update}
\end{align}
Here, we have also replaced the true score $s_t^{\star}$ with the score estimate $s_t$ on the right-hand side of \eqref{eq:score_update}. Note that the Lagrange interpolating polynomial on right-hand side of \eqref{eq:score_update} forms a unique  degree-$(K-1)$ polynomial passing through these $K$ points, and we take advantage of $K$ score function evaluations to approximate $\frac{1}{(1-\tau)^{3/2}}s_{\tau}^{\star} (Y^{\mathsf{ode}}_{\tau})$.

%This motivates the use of a $(K-1)$-th order approximation of the true score function. 

\paragraph{Successive refinement.}

%Since we only get to observe the forward process at those discrete points with $\tau = 1 - \overline{\alpha}_t,$ for $1\leq t\leq T$,

Noteworthily, the polynomial appproximation \eqref{eq:score_update} still cannot be readily used, given that the points $\{Y^{\mathsf{ode}}_{\tau_{t, i}}\}$ are inaccessible in general. To remedy this issue, we propose a successive refinement scheme that alternates between estimating $\{Y^{\mathsf{ode}}_{\tau_{t, i}}\}$ and computing the the associated scores. 
%
% The idea is to approximate this sequence using ODE~\eqref{eq:ODE} while invoking the polynomial approximation of score functions in \eqref{eq:score_update}. 
%More specifically, we propose to alternate between approximating the true score function via \eqref{eq:score_update} and updating the sequence $\{x_{\tau_{t, i}}\}$ using the score function estimates. 
%
More precisely, let us perform an iterative estimation procedure containing $N$ rounds, where in each round $n$, we produce a sequence $\{x_{\tau_{t, i}}^{(n)}\}_{0\leq i < K}$ as an estimate of  $\{Y^{\mathsf{ode}}_{\tau_{t, i}}\}$. When $\{x_{\tau_{t, i}}^{(n)}\}_{0\leq i < K}$ is available, 
we compute, for each $\tau' \in \{\tau_{t,0},\dots,\tau_{t,K-1}\}$, 
%
%the proposed update rule for computing $\{x_{\tau_t, i}^{(n+1)}\}_{0\leq i \leq K-1}$ is given by
%
% By iterating over this process for $N$ times, we obtain a sequence $\{x_{\tau_t, i}^{(n)}\}_{0\leq i \leq K-1, 0 \leq n \leq N}.$ 
%Mathematically, suppose $x_{\tau_{t, 0}}$ and $\{x^{(0)}_{\tau_{t, i}}\}_{0\leq i\leq K-1}$ are given, for every $n = 0, \ldots N-1$, we obtain the sequence $\{x_{\tau_{t, i}}^{(n+1)}\}_{0\leq i \leq K-1}$ via the following rule 
%
%\begin{subequations}
%
\begin{align}
	%\forall i = 0, \ldots, K-1, \qquad
	\frac{x_{\tau'}^{(n+1)}}{\sqrt{1-\tau'}} = 
	%\frac{x_{\tau_{t, 0}}}{\sqrt{1-\tau_{t, 0}}} - \int_{\tau_{t, 0}}^{\tau'} \frac{1}{2(1-\tau)^{3/2}}s_{\tau}^{(n)}\mathrm{d}\tau \notag\\
	%&\coloneqq 
	& \frac{x_{\tau_{t, 0}}}{\sqrt{1-\tau_{t, 0}}} + \sum_{0 \le j \leq K-1} \gamma_{t, j}(\tau') \frac{1}{(1-\tau_{t, j})^{3/2}}s_{\tau_{t, j}}(x_{\tau_{t, j}}^{(n)}). \label{eq:update_rule}\\
	&\text{where}\quad \gamma_{t, i}(\tau') \coloneqq \int_{\tau'}^{\tau_{t, 0}}\psi_i(\tau)\mathrm{d}\tau 
	\quad \text{and}\quad x_{\tau_{t, 0}} = Y_t.
	\label{def:gamma}
\end{align}
In a nutshell, we employ $\{x_{\tau_{t, i}}^{(n)}\}_{0\leq i \leq K-1}$ to update the scores used in the polynomial approximation \eqref{eq:score_update}, which in turn helps us generate $\{x_{\tau_{t, i}}^{(n+1)}\}_{0\leq i \leq K-1}$ as inspired by the ODE \eqref{eq:ODE}. This strategy is expected to yield improved estimates of $\{Y^{\mathsf{ode}}_{\tau_{t, i}}\}$ as the number of rounds increases.

%and $x_{\tau_{t, 0}}$ is taken t

%\end{subequations}
%
%for every $i = 0, \ldots, K-1$. 
%Here, $\tau'$ denotes $\tau_{t, i}$ and
%\begin{align}

%\end{align}

\paragraph{Algorithm description.} 
With the above ideas in mind, we are now ready to present our algorithm. 
%Working backward from $t=T$ to $t=1$, our algorithm maintains a sequence $\{x_{\tau_t, i}^{(n)}\}_{0\leq i <K, 0 \leq n \leq N}$ for each $t$.  

\begin{itemize}
	\item \textbf{Initialization.} 
	Let us initialize our algorithm with $x_{\tau_{T, 0}} = Y_{T} \sim \mathcal{N}(0, I_d)$ and $x_{\tau_{T, i}}^{(0)} = x_{\tau_{T, 0}}$ for each %$0 \leq n \leq N$ and 
	$0 < i < K$. 
	To ease presentation, we also take 
	$x_{\tau_{T+1, K-1}}^{(N)} = x_{\tau_{T, 0}}$. 
	% Note that by definition $x_{\tau_{t, 0}} = x_{1-\overline{\alpha}_{t}} = X_t.$
	%
	% \yutingcomment{where do you have $n$?}

	% Let $\Phi \in \mathbb{R}^{K\times K}$ with $\Phi_{ji} = (\tau_{t, i} - \tau_{t, 0})^j$.
	% Then given $(1-\tau_{t, i})^{-3/2}s_{\tau_{t, i}}(x_{\tau_{t, i}}^{(n)})$ for $0 \le i < K$, we can use the Taylor expansion to approximate $(1-\tau)^{-3/2}s_{\tau}$ as follows:
	% \begin{align}
		% (1-\tau)^{-3/2}s_{\tau}^{(n)} 
		% %= \sum_{0 \le j < k} \phi_j^{(n)}(\tau - \tau_{t, 0})^j
		% := \sum_{0 \le i < K} \psi_i(\tau)(1-\tau_{t, i})^{-3/2}s_{\tau_{t, i}}(x_{\tau_{t, i}}^{(n)}),\quad \forall \tau \in [\tau_{t,K-1}, \tau_{t,0}]\label{eq:score_update}
		% \end{align}
	% where Lagrange interpolating polynomial
	% %\begin{align*}
	% %\big[\phi_j^{(n)}\big]_{0 \le j < k}
	% %= \Phi^{-1}
	% %\big[(1-\tau_{t, i})^{-3/2}s_{\tau_{t, i}}(x_{\tau_{t, i}}^{(n)})\big]_{0 \le i < k},
	% %\end{align*}
	% %which satisfies
	% \begin{align}\label{ineq1}
		% \big[\psi_i(\tau)\big]_{0 \le i < K}
		% = \Phi^{-1}[(\tau - \tau_{t, 0})^j]_{0 \le j < K}
		% = \bigg[\frac{\prod_{i' \ne i} (\tau_{t, i'} - \tau)}{\prod_{i' \ne i} (\tau_{t, i'} - \tau_{t, i})}\bigg]_{0 \le i < K}.
		% \end{align}
	%The above formulation leads to the following update rule.
	%Initialize $x_{\tau_{T, i}}^{(0)} = x_{\tau_{T, 0}} \sim \mathcal{N}(0, I)$.
	%For any \blue{fixed} $2 \le t \le T $
	
	\item \textbf{Iterative update rule.}
	Working backward from $t=T$ to $t=1$, 
	we generate the point $Y_{t-1}$ as follows. Look at the time points $\{\tau_{t,i}\}_{0\leq i < K}$, and initialize both the sequence $\{x_{\tau_{t, i}}^{(0)}\}_{0\leq i \leq K-1}$ and $x_{\tau_{t, 0}}$ using the value $x_{\tau_{t+1, K-1}}^{(N)}$ obtained in the previous iteration.  For each round $n=0,\dots,N-1$, we generate a sequence $\{x_{\tau_{t, i}}^{(n+1)}\}_{0\leq i \leq K-1}$ according to Eq.~\eqref{eq:update_rule}. 
	After $N$ rounds of successive refinement, we output $Y_{t-1} = x_{\tau_{t, K-1}}^{(N)}$. 
	
	%For every $n = 0, \ldots N-1$, we obtain the sequence $\{x_{\tau_{t, i}}^{(n+1)}\}_{0\leq i \leq K-1}$ via Eq.~\eqref{eq:update_rule}.  
	
	% . Let us describe the iterative process for generating $Y_t$.  
	%
	%Suppose that we have already obtained $x_{\tau_{t+1, K-1}}^{(N)}$. Now the goal is to update $\{x_{\tau_{t,i}}^{(n)}\}_{0\leq i \leq K-1, 0\leq n\leq N}$. 
	%Initialize both the sequence $\{x_{\tau_{t, i}}^{(0)}\}_{0\leq i \leq K-1}$ and $x_{\tau_{t, 0}}$ with the value $x_{\tau_{t+1, K-1}}^{(N)}$. %in the deterministic case and $x_{\tau_{t-1, i}}^{(0)} = x_{\tau_{t-1, 0}} = \sqrt{\alpha_{t-1}}x_{\tau_{t, k-1}}^{(N)} + \sqrt{1-\alpha_{t-1}}z_t$ with $z_t \sim \mathcal{N}(0, I)$ in the stochastic case,

	%\item \textbf{Output of the algorithm.} We define $Y_{t} = x_{\tau_t, 0}$ for every $1 \leq t \leq T$.
\end{itemize}
The whole procedure of the proposed algorithm is summarized in Algorithm~\ref{algorithm:high_order}. 

Several remarks are in order. To begin with, once the value of $Y_T$ is given, the sequence $\{x_{\tau_{t, i}}^{(n)}\}$ is purely deterministic. Thus, when accurate score functions are available at the requested time points, the discrepancy between the distributions of $Y_t$ and $X_t$ mainly stems from the approximation error when interpolating a complicated function using polynomials of degree $K-1$. 
In addition, our algorithm requires computing the score functions at the sequence $\{x_{\tau_{t, i}}^{(n)}\}_{0\leq i \leq K-1, 0\leq n \leq N, 1\leq t\leq T}$, and hence the total number of score function evaluations is  $TKN$. As we shall see momentarily in our theoretical development,  $K$ is taken to be a constant, whereas $N$ is chosen to be logarithmically large. Consequently, the total number of score function evaluations is on the order of $\widetilde{O}(T)$. 

 \begin{remark}[The time indices $t$ vs.~$\tau$]
	In Algorithm~\ref{algorithm:high_order}, we employ both the discrete-time index $t$ and the continuous variable $\tau$, both of which are instrumental in describing our algorithm. Recall that $t$ is a discrete step-index for both the forward and backward processes in discrete time, while $\tau$ is the continuous time variable (cf.~\eqref{eq:ODE}) taking value in $[0,1]$. At Step $t$, we iteratively refine our estimate of $Y_{t-1}$, or equivalently, $Y_{1-\overline{\alpha}_{t-1}}^{\sf ode}$, at $K$ time points within the interval $[1-\overline{\alpha}_{t-1}, 1-\overline{\alpha}_t]$, where these intermediate points do not need to coincide with the time indices of the actual discrete-time process.
	\end{remark}

\section{Main theory}
\label{sec:main-theory}

Encouragingly, the proposed algorithm comes with intriguing convergence guarantees, as we present in this section.  
%In what follows, we begin by introducing a couple of key assumptions, followed by the main theoretical guarantees. 
Throughout this paper, we let $q_t$ (resp.~$p_t$) denote the distribution of $X_t$ (resp.~$Y_t$).

\subsection{Assumptions}\label{sec:assumptions}

Let us gather the key assumptions needed to state our main theorem. 
First of all, the sequence $\{\beta_t\}$ that determines the variance of the injective noise (cf.~\eqref{eq:forward-process}) is taken to be
\begin{subequations}
	\begin{align}
		\beta_1 &= 1 - \alpha_1 = {T^{-c_0}}\label{eq:learning_rate_1}\\
		\beta_t &= 1 - \alpha_t = \frac{c_1\log T}{T}\min\bigg\{\beta_1\left(1 + \frac{c_1\log T}{T}\right)^t, 1\bigg\},
		\qquad t\geq 2
		\label{eq:learning_rate_2}
	\end{align}
\end{subequations}
for some large enough numerical constants $c_0,c_1>0$. In a nutshell, $\beta_t$ undergoes two phases, growing exponentially fast at the beginning and then staying flat in the remaining steps. Such a two-phase schedule is consistent with what is used in the state-of-the-art diffusion theory work (e.g., \citet{benton2024nearly,li2024sharp,li2024d,huang2024denoising}).  

Next, we turn to assumptions for the target data distribution. Our goal is to accommodate a very broad family of data distributions without imposing restrictions like smoothness or log-concavity. In fact, the only assumption we make about $\Pdata$ is the following boundedness condition, where the radius is allowed to be polynomially large, with an arbitrarily large constant degree. A polynomially large support size can readily accommodate a very broad range of applications. 
\begin{assumption}[Target data distribution]\label{assump:data_distribution}
	The target distribution $\Pdata$ satisfies
	\begin{align*}
		\mprob\left(\|X_0\|_2 \leq T^{c_R}\,| \,X_0 \sim p_{\Pdata}\right) = 1
	\end{align*}
	for some arbitrarily large constant $c_R > 0$.
\end{assumption}
\begin{remark}
		By leveraging the techniques in \citet{li2024provable}, one might be able to further relax Assumption \ref{assump:data_distribution} to a moment condition of the form $$\mathbb{E}[\|X_0\|_2^2] \leq T^{2c_R}$$ for an arbitrarily large constant ${c_R}$ without affecting our convergence rates. We do not pursue this refinement in the present work for simplicity.
\end{remark}

In the presence of imperfect score estimation, it is natural to anticipate that the sampling quality is largely affected by the goodness of the score function estimates. In light of this, we introduce another set of assumptions to capture the quality of score function estimates. The first one below is concerned with the mean squared score estimation error, averaged over all time steps and iterations. 
\begin{assumption}\label{assump:score_estimation_1}
	Suppose that the score estimates at hand satisfy
	%\begin{align*}
	%\frac{1}{TNk}\sum_{t, i, n} \mathbb{E}_{Y_t \sim q_t}\Big[\big\|s_{\tau_{t, i}}(x_{\tau_{t, i}}^{(n)}) - s_{\tau_{t, i}}^{\star}(x_{\tau_{t, i}}^{(n)})\big\|_2^2\Big] \le \varepsilon_{\mathsf{score}}^2,
	%\end{align*}
	\begin{align*}
		\frac{1}{TNK}\sum\nolimits_{t=1}^T \mathbb{E}_{Y_t \sim q_t}\left[\varepsilon_{\mathsf{score}, t}^2(Y_t)\right] \le \varepsilon_{\mathsf{score}}^2,
	\end{align*}
	where $\varepsilon_{\mathsf{score}, t}^2(x_{\tau_{t}, 0}) = \sum_{i=0}^{K-1}\sum_{n=0}^{N-1}\big\|s_{\tau_{t, i}}(x_{\tau_{t, i}}^{(n)}) - s_{\tau_{t, i}}^{\star}(x_{\tau_{t, i}}^{(n)})\big\|_2^2$.
\end{assumption}
% \yutingcomment{isn't $x_{\tau_{t, i}}^{(n)}$ an algorithm dependent quantity? a bit strange...}

In stark contrast to the SDE-based samplers (e.g., DDPM) for which the $\ell_2$ estimation error alone suffices for the stability analysis (e.g., \citet{chen2022sampling,benton2024nearly}),  the ODE-based samplers require additional assumptions in order to preclude some unfavorable problem instances.  See, e.g., \cite{li2024sharp} for a lower bound that justifies why $\ell_2$ score accuracy assumption alone is insufficient. In this work, we impose the following assumption concerning the goodness of the associated Jacobian errors. 
\begin{assumption}\label{assump:score_estimation_2}
	Suppose that $s_t(\cdot)$ is continuously differentiable for each $1 \leq t \leq T$, and satisfies
	%\begin{align*}
	%\frac{1}{TNk}\sum_{t,i,n} \mathbb{E}_{Y_t \sim q_t}\bigg[\bigg\|\frac{\partial s_{\tau_{t, i}}(x_{\tau_{t, i}}^{(n)})}{\partial x} - \frac{\partial s_{\tau_{t, i}}^{\star}(x_{\tau_{t, i}}^{(n)})}{\partial x}\bigg\|^2\bigg] \le \varepsilon_{\mathsf{Jacobi}}^2,
	%\end{align*}
	\begin{align*}
		\frac{1}{TNK}\sum\nolimits_{t=1}^T  \mathbb{E}_{Y_t \sim q_t}\left[\varepsilon_{\mathsf{Jacobi}, t}^2(Y_t)\right] \le \varepsilon_{\mathsf{Jacobi}}^2,
	\end{align*}
	where $\varepsilon_{\mathsf{Jacobi}, t}^2(x_{\tau_{t}, 0}) = \sum_{i=0}^{K-1}\sum_{n=0}^{N-1} \Big\|\frac{\partial s_{\tau_{t, i}}(x_{\tau_{t, i}}^{(n)})}{\partial x} - \frac{\partial s_{\tau_{t, i}}^{\star}(x_{\tau_{t, i}}^{(n)})}{\partial x}\Big\|^2$.
\end{assumption}
\begin{remark}
	When $N = K = 1$, these are equivalent to  \citet[Assumptions~1 and 2]{li2024sharp}. 
\end{remark}

\begin{comment}
	\paragraph{Convergence rate.}
	\begin{theorem}\label{theorem:main-stochastic}
		For the stochastic sampler,
		\begin{align}
			\mathsf{TV}\big(q_1, p_1\big) \leq C_1 \sqrt{d\log T}\bigg(\frac{d\log^2 T}{T}\bigg)^{k}
			+C_1\varepsilon_{\mathsf{score}}\log T
			\label{eq:ratio_stochastic}
		\end{align}
	\end{theorem}
\end{comment}

\subsection{Convergence guarantees}

We are now in a position to present our main theoretical guarantees for the proposed algorithm. 

\begin{theorem}
	% [Theorem 1]
	\label{theorem:main}
	Suppose that Assumptions \ref{assump:data_distribution}, \ref{assump:score_estimation_1} and \ref{assump:score_estimation_2} hold. Let $K>0$ be an arbitrary fixed integer.  If $T \geq C_2d^2\log^3T$ and $N = \lceil C_3\log T\rceil$ for some large enough constants $C_2, C_3 > 0$, then Algorithm~\ref{algorithm:high_order} achieves
	%
	%For the deterministic sampler,
	\begin{align}
		\mathsf{TV}\big(p_{X_1}, p_{Y_1}\big) \leq C_1 d^2\bigg(\frac{d\log^2 T}{T}\bigg)^{K}\log^2 T
		+C_1\sqrt{d\log^{4}T}\,\varepsilon_{\mathsf{score}}+ C_1d(\log^{3/2} T)\varepsilon_{\mathsf{Jacobi}}.
		\label{eq:ratio}
	\end{align}
	% where we recall that $q_1$ and $p_1$ represent the distributions of $X_1$ and $Y_1$, respectively. 
\end{theorem}

When exact score functions are available (so that $\varepsilon_{\mathsf{score}}=\varepsilon_{\mathsf{Jacobi}}=0$), Theorem~\ref{theorem:main} reveals that the number of iterations needed to approximate $q_1$ to within $\varepsilon$ TV distance is no greater than 
\begin{align}
	\text{(iteration complexity)}\qquad \widetilde{O}\bigg( \frac{ d^{1+2/K} }{  \varepsilon^{1/K} }  \bigg). 
	\label{eq:iteration-complexity-ours}
\end{align}
Given that $K$ is a fixed constant and $N\asymp \log T$, the total number of score function evaluations of the proposed algorithm is given by $O(NKT)=O(T\log T)$, which can be as small as  $\widetilde{O}\big( \frac{ d^{1+2/K} }{  \varepsilon^{1/K} } \big)$ (cf.~\eqref{eq:iteration-complexity-ours}). 
In particular, when $K$ grows, one can approximately interpret this number of function evaluations as 
$$\widetilde{O}\bigg(\frac{d^{1+o(1)}}{\varepsilon^{1/K}}\bigg),$$ 
resulting in substantial speed-ups compared to prior theory in \cite{li2024sharp,li2024accelerating,li2024provable,wu2024stochastic}. In particular, our algorithm achieves significant acceleration comparing to the standard discretized probability flow ODE sampler (without the successive-refinement step), which attains $\varepsilon$-accuracy with $$\widetilde{O}(d/\varepsilon)$$ score function evaluations \citep{li2024sharp}.
In the noisy case when we only have access to inexact scores, our result guarantees that the TV distance between our output distribution and the target distribution scales proportionally to the two error metrics  $\varepsilon_{\mathsf{score}}$ and $\varepsilon_{\mathsf{Jacobi}}.$ It is worth noting that: without assumptions on the smoothness of the score estimates, additional assumptions like the one concerning Jacobian errors cannot be avoided in general, as already illustrated by a counterexample in \cite{li2023towards} for the probability flow ODE.

%As mentioned, the total number of score estimations of our proposed algorithm is equal to $NKT$, where $K$ is a constant, $N$ is of order $\log T$. Therefore, compared to DDIM and DDPM, which only require $T$ score estimations, our proposed algorithm needs additional score estimations, but only by a factor of $\log T.$  

Notably, our approach bears some similarity with other higher-order ODE solvers recently developed in the literature, such as the DPM solver \citep{lu2022dpm}, DPM-Solver++ \citep{lu2025dpm}, and UNiPC \citep{zhao2023unipc}. While remarkable empirical success of such methods have been reported, their theoretical footings remain significantly underdeveloped.  Recently, \cite{huang2024convergence} analyzed the theoretical performance of the $K$-th order Runge-Kutta method for solving the probability flow ODE, for any constant order $K$. While their theory is comparable to ours, it requires higher-order smoothness on the score estimates (more precisely, they required the score estimate's first $(K+1)$-th derivatives to be bounded). In addition, their iteration complexity also scales linearly in the support size of $\Pdata$; in contrast, our theory is nearly independent from such a support size as long as it is polynomially bounded. 
Finally, while we were finalizing our paper, we became aware of a concurrent work \cite{huang2025fast}, which --- under a different set of assumptions --- also relaxes the high-order smoothness condition previously imposed in \cite{huang2024convergence}. Compared with our results, their analysis requires a stronger assumption on the support size, and their total-variation bound depends heavily upon the early-stopping parameter (denoted by $\tau$ therein).

\section{Main analysis}
\label{sec:analysis}

This section outlines the proof steps for establishing Theorem~\ref{theorem:main}. 
Here and throughout, we assume 
\begin{subequations}
	\begin{align}
		\varepsilon_{\mathsf{score}} \leq ({C_1\sqrt{d}\log^2 T})^{-1},\label{bound:score_estimation_1}\\
		\varepsilon_{\mathsf{Jacobi}} \leq ( C_1d\log^{3/2} T )^{-1}; \label{bound:score_estimation_2}
	\end{align}
\end{subequations}
otherwise,  \eqref{eq:ratio} is trivially satisfied given TV distance is always bounded by 1. 
%\paragraph{Additional notation.} 
\subsection{Additional notation}
%\paragraph{Additional notation.} 
%
Before proceeding to the main proof, 
we find it helpful to introduce an additional set of notation. 
Fix $x_T \in\mathbb{R}^d$, we slightly abuse notation by letting $\{x_{\tau_{t, i}}^{(n)}\}$ denote the deterministic sequence defined by the update rule \eqref{eq:update_rule} with $x_{\tau_{T, 0}} = x_T$. 
%For any point $x_T \in\mathbb{R}^d$, we let $\{x_{\tau_{t, i}}^{(n)}\}$ denote the deterministic sequence satisfying the update rule \eqref{eq:update_rule} with $x_{\tau_{T, 0}} = x_T$. 
We also define $x_{t} := x_{\tau_t, 0}$ for all $1 \leq t \leq T-1$. %Then according to the update rule, $x_{\tau_{t,i}}^{(n)} = \phi_{t,i}^{(n)}(x_{\tau_{t,i}}^{(0)})$ is a function of $x_{\tau_{t,i}}^{(0)}$, where $$\phi_{t,i}^{(n+1)}(x) = \sqrt{\frac{1-\tau_{t, i}}{1-\tau_{t, 0}}}x + \sqrt{1-\tau_{t, i}}\sum_{0 \leq j < K}\gamma_{t, j}(\tau_{t, i})(1 - \tau_{t, j})^{-3/2}s_{\tau_{t, j}}(\phi_{t,j}^{(n)}(x))$$ and $\phi_{t,i}^{(0)}(x) = x$. 
We let $x_t^\star$ (resp.~$Y_t^\star$) denote the solution of ODE \eqref{eq:ODE} at $\tau = \tau_{t,0} = 1 - \overline{\alpha}_t$ with the initial condition $x_{\tau_{t+1,0}}^\star = x_{t+1}$ (resp.~$x_{\tau_{t+1,0}}^\star = Y_{t+1}$). For any $\tau \in [\tau_{t,K-1}, \tau_{t,0}] = [1-\overline{\alpha}_{t-1}, 1-\overline{\alpha}_t]$, we denote by $x_{\tau}^\star$ the solution of ODE \eqref{eq:ODE} at $\tau$ with the initial condition $x_{\tau_{t,0}}^\star = x_{t}$. For any $0 \leq i \leq K-1$, $1 \leq t \leq T$, $1 \leq n \leq N$, we define 
\begin{align*}
	\varepsilon_{\mathsf{score}, t, i}^{(n)}(x_{\tau_{t, i}}^{(n)}) \coloneqq \big\|s_{\tau_{t, i}}(x_{\tau_{t, i}}^{(n)}) - s_{\tau_{t, i}}^{\star}(x_{\tau_{t, i}}^{(n)})\big\|_2;~~\varepsilon_{\mathsf{Jacobi}, t, i}^{(n)}(x_{\tau_{t, i}}^{(n)}) \coloneqq \bigg\|\frac{\partial s_{\tau_{t, i}}(x_{\tau_{t, i}}^{(n)})}{\partial x} - \frac{\partial s_{\tau_{t, i}}^{\star}(x_{\tau_{t, i}}^{(n)})}{\partial x}\bigg\|.
\end{align*}
\begin{comment}
	Moreover, we let
	\begin{align*}
		\big(\varepsilon_{\mathsf{score}, t, i}^{(n)}\big)^2 := \mathbb{E}\left[\big(\varepsilon_{\mathsf{score}, t, i}^{(n)}(x_{\tau_{t, i}}^{(n)})\big)^2\right]~\quad~\text{and}~\quad~\big(\varepsilon_{\mathsf{Jacobi}, t, i}^{(n)}\big)^2 := \mathbb{E}\left[\big(\varepsilon_{\mathsf{Jacobi}, t, i}^{(n)}(x_{\tau_{t, i}}^{(n)})\big)^2\right].
	\end{align*}
	denote the population versions.
\end{comment}
%
%For any $x_{T} = x_{\tau_{T},0} \in \mathbb{R}^d$, we let $\{x_{\tau_{t,i}}^{(n)}\}$ denote the deterministic sequence generated by \eqref{eq:update_rule}, and define $x_t = x_{\tau_t, 0} = x_{\tau_{t-1},K-1}^{(N)}$.
It is also defined that
\begin{subequations}
	\begin{align}
		\xi_t(x_{t}) &:= %\frac{d^{5/2}\log^{7/2}T}{T^{2}}\sqrt{\sum_{i,n}\big(\varepsilon_{\mathsf{score}, t, i}^{(n)}(x_{\tau_{t, i}}^{(n)})\big)^2} + \frac{d^{3/2}\log^{3/2} T}{T}\sqrt{\sum_{i,n}\big(\varepsilon_{\mathsf{Jacobi}, t, i}^{(n)}(x_{\tau_{t, i}}^{(n)})\big)^2}
		\frac{\log T}{T}\left(d\sqrt{\sum_{i,n}\big(\varepsilon_{\mathsf{Jacobi}, t, i}^{(n)}(x_{\tau_{t, i}}^{(n)})\big)^2} + \sqrt{d\log T}\sqrt{\sum_{i,n}\big(\varepsilon_{\mathsf{score}, t, i}^{(n)}(x_{\tau_{t, i}}^{(n)})\big)^2}\right),\label{eq9a}\\
		S_t(x_T) &:= \sum_{k=2}^t\xi_t(x_t)~\quad~\text{for}~t\geq 2,~\quad~\text{and}~S_1(x_T) = 0.\label{eq9b}
	\end{align}
\end{subequations}
\begin{comment}
	In addition, we let
	\begin{align}\label{eq:score_t}
		\varepsilon_{\mathsf{score}, t}(x_t) := \sqrt{\sum_{i,n}\big(\varepsilon_{\mathsf{score}, t, i}^{(n)}(x_{\tau_{t, i}}^{(n)})\big)^2},~\quad~\text{and}~\quad~\varepsilon_{\mathsf{Jacobi}, t}(x_t) := \sqrt{\sum_{i,n}\big(\varepsilon_{\mathsf{Jacobi}, t, i}^{(n)}(x_{\tau_{t, i}}^{(n)})\big)^2}.
	\end{align}
\end{comment}
%
From the definitions of $\varepsilon_{\mathsf{score}, t}(\cdot)$ and $\varepsilon_{\mathsf{Jacobi}, t}(\cdot)$, it follows that
{\small 
	\begin{align}
		\frac{1}{T}\sum_{t}\mathbb{E}_{Y_t \sim q_t}\left[\varepsilon_{\mathsf{score}, t}(Y_t)\right] &\leq \sqrt{\frac{1}{T}\sum_{t}\mathbb{E}_{Y_t \sim q_t}\left[\varepsilon_{\mathsf{score}, t}^2(Y_t)\right]} = \sqrt{NK}\varepsilon_{\mathsf{score}} \asymp \varepsilon_{\mathsf{score}}\sqrt{\log T},\label{ineq47a}\\
		\frac{1}{T}\sum_{t}\mathbb{E}_{Y_t \sim q_t}\left[\varepsilon_{\mathsf{Jacobi}, t}(Y_t)\right] &\leq \sqrt{\frac{1}{T}\sum_{t}\mathbb{E}_{Y_t \sim q_t}\left[\varepsilon_{\mathsf{Jacobi}, t}^2(Y_t)\right]} = \sqrt{NK}\varepsilon_{\mathsf{Jacobi}} \asymp \varepsilon_{\mathsf{Jacobi}}\sqrt{\log T}.\label{ineq47b}
	\end{align}
}
%We mainly focus on the case that $\max\{-\log p_{\tau}(x_{\tau}^\star), -\log p_{X_{\tau_{t,i}}}(x_{t,i}^{(n)})\} \lesssim d\log T$, where the remaining case can be bounded easily and is omitted here due to the overwhelming probability.

\subsection{Preliminary facts}
%\paragraph{Important facts.}
The following lemmas will be frequently used throughout the proof. The first lemma introduces some helpful properties concerning the learning rates.
\begin{lemma}\label{lemma:learning_rate}
	For large enough $T$, one has
	\begin{subequations}
		\begin{align}
			\alpha_{t} &\geq 1 - \frac{c_1\log T}{T} \geq \frac{1}{2},\quad~1 \leq t \leq T\label{ineq:learning_rate_a}\\
			\frac{1}{2}\frac{1-\alpha_{t}}{1 - \overline{\alpha}_t} \leq \frac{1}{2}\frac{1-\alpha_{t}}{\alpha_{t} - \overline{\alpha}_t} &\leq \frac{1-\alpha_{t}}{1 - \overline{\alpha}_{t-1}} \leq \frac{4c_1\log T}{T},\quad~2 \leq t \leq T\label{ineq:learning_rate_b}\\
			1 &\leq \frac{1 - \overline{\alpha}_{t}}{1 - \overline{\alpha}_{t-1}} \leq 1 + \frac{4c_1\log T}{T},\quad~2 \leq t \leq T\label{ineq:learning_rate_c}\\
			\overline{\alpha}_T &\leq \frac{1}{T^{c_2}},\label{ineq:learning_rate_d}\\
			\frac{\overline{\alpha}_{t+1}}{1-\overline{\alpha}_{t+1}} &\leq \frac{\overline{\alpha}_{t}}{1-\overline{\alpha}_{t}} \leq \frac{4\overline{\alpha}_{t+1}}{1-\overline{\alpha}_{t+1}},\quad~1 \leq t < T\label{ineq:learning_rate_e}\\
				\Big|\frac{\tau_{t, i_1} - \tau_{t, i_2}}{\tau_{t, i_3}(1-\tau_{t, i_4})}\Big| &\leq 8c_1\frac{\log T}{T},\quad~2 \leq t \leq T, 0 \leq i_1, i_2, i_3, i_4 \leq K-1,\label{ineq:learning_rate_f}\\
				|\gamma_{t, i}(\tau_{t,j})| &\le 2^K (\tau_{t, 0} - \tau_{t,j}),\quad~\forall 0 \leq i, j \leq K-1\label{ineq:learning_rate_g}\\
	        1 - \tau_{t,i}& \asymp 1 - \tau_{t,j},~\quad~\tau_{t, i} \asymp \tau_{t, j},~\quad~\forall 0 \leq i, j \leq K-1, 2 \leq t \leq T.\label{ineq:learning_rate_h}
		\end{align}
	Here, $c_2$ is a sufficiently large constant.
	\end{subequations}
\end{lemma}
For notational convenience, we define
\begin{align}\label{def:theta}
	\theta_{\tau}(x) := \max\left\{-\frac{\log p_{\overline{X}_\tau}(x)}{d\log T}, c_6\right\} 
\end{align}
for some large enough constant $c_6 > 0$. The second lemma establishes the tail bounds of $X_0$ conditioned on the continuous-time forward process $\overline{X}_{\tau}$ (cf.~\eqref{defn:X-tau-process}).
\begin{lemma}[Lemma 1 in \cite{li2024sharp}]\label{lemma:tail_bound}
	Suppose that Assumption \ref{assump:data_distribution} holds and $\tau \leq 1 - \frac{1}{T^{c_0}}$. Then for any $c_5 \geq 2$, conditioned on $\overline{X}_{\tau} = y$, we have
	\begin{align}\label{ineq6}
		\left\|\sqrt{1 - \tau}X_0 - y\right\|_2 \leq 5c_5\sqrt{\theta_{\tau}(y)d\tau\log T}
	\end{align}
with probability exceeding $1 - \exp(-c_5^2\theta_{\tau}(y)d\log T)$. Furthermore, one has
\begin{subequations}
	\begin{align}
		\mathbb{E}\left[\left\|\sqrt{1 - \tau}X_0 - y\right\|_2\ |\ \overline{X}_{\tau} = y\right] &\leq 12\sqrt{\theta_{\tau}(y)d\tau\log T},\label{ineq7a}\\
		\mathbb{E}\left[\left\|\sqrt{1 - \tau}X_0 - y\right\|_2^2\ |\ \overline{X}_{\tau} = y\right] &\leq 120\theta_{\tau}(y)d\tau\log T,\label{ineq7b}\\
		\mathbb{E}\left[\left\|\sqrt{1 - \tau}X_0 - y\right\|_2^3\ |\ \overline{X}_{\tau} = y\right] &\leq 1040\left(\theta_{\tau}(y)d\tau\log T\right)^{3/2},\label{ineq7c}\\
		\mathbb{E}\left[\left\|\sqrt{1 - \tau}X_0 - y\right\|_2^4\ |\ \overline{X}_{\tau} = y\right] &\leq 10080\left(\theta_{\tau}(y)d\tau\log T\right)^{2},\label{ineq7d}.
	\end{align}
\end{subequations}
\end{lemma}
The next lemma provides key properties of the score function $s_{\tau}^\star(\cdot)$.
\begin{lemma}\label{lemma:Hessian}
	One has
	\begin{subequations}
		\begin{align}
			\left\|s_{\tau}^\star(x)\right\|_2 &\lesssim \sqrt{\frac{d\theta_{\tau}(x)\log T}{\tau}},\label{score_1}\\
			\left\|\frac{\partial s_{\tau}^\star(x)}{\partial x}\right\| &\lesssim \frac{d\theta_{\tau}(x)\log T}{\tau}.\label{score_2}
		\end{align}
	\end{subequations}
Furthermore, if $-\log p_{\overline{X}_\tau}(\lambda x_1 + (1 - \lambda)x_2) \lesssim d\log T$ for all $\lambda \in [0,1]$, then
\begin{align}\label{score_derivative_difference}
	\left\|\frac{\partial s_{\tau}^\star(x_1)}{\partial x} - \frac{\partial s_{\tau}^\star(x_2)}{\partial x}\right\| &\lesssim \sqrt{\frac{d^3\log^3 T}{\tau^3}}\|x_1 - x_2\|_2.
\end{align}
\end{lemma}
The proofs of Lemma \ref{lemma:learning_rate} and Lemma \ref{lemma:Hessian} are deferred to Sections \ref{sec:proof_lemma_learning_rate} and \ref{sec:proof_lemma_Hessian}, respectively.

The following lemma shows that if $-\log p_{\overline{X}_{\tau}}(x_{\tau}^\star)$ is not too large, then $-\log p_{\overline{X}_{\tau'}}(x_{\tau'}^\star)$ can be well-controlled for all $\tau'$ sufficiently close to $\tau$.
\begin{lemma}\label{lemma:log_density}
	Suppose that $-\log p_{\overline{X}_{\widetilde\tau}}(x_{\widetilde\tau}^\star) \leq \theta d\log T$ for some $\theta > 1$. Then for all $|\tau' - \widetilde\tau| \leq c_0\widetilde\tau(1 - \widetilde\tau)$ for some sufficiently small constant $c_0 > 0$, one has
	\begin{align}
		-\log p_{\overline{X}_{\tau'}}(x_{\tau'}^\star) \leq 2\theta d\log T.
	\end{align}
\end{lemma}
The proof of Lemma \ref{lemma:log_density} can be found in Section \ref{proof:Lemma_log_density}. The last lemma reveals that the the distributions of $X_T$ and $Y_T$ are close.
\begin{lemma}[Lemma 3 in \cite{li2024sharp}]\label{lemma:distance_y_T}
	Suppose that $T$ is large enough. Then
	\begin{align}
		\big(\mathsf{TV}(p_{X_T}~||~p_{Y_T})\big)^2 \leq \frac{1}{2}\mathsf{KL}(p_{X_T}~||~p_{Y_T}) \lesssim \frac{1}{T^{200}}.\label{ineq:distance_y_T}
	\end{align}
\end{lemma}

%For any $x_{T} = x_{\tau_{T},0} \in \mathbb{R}^d$, we let $\{x_{\tau_{t,i}}^{(n)}\}$ denote the deterministic sequence generated by \eqref{eq:update_rule}, and define $x_t = x_{\tau_t, 0} = x_{\tau_{t-1},K-1}^{(N)}$. 

%We mainly focus on the case that $\max\{-\log p_{\tau}(x_{\tau}^\star), -\log p_{X_{\tau_{t,i}}}(x_{t,i}^{(n)})\} \lesssim d\log T$, where the remaining case can be bounded easily and is omitted here due to the overwhelming probability.
\subsection{Main steps for proving Theorem \ref{theorem:main}}
We now proceed to present the proof ideas, which comprises multiple steps as detailed below. 
\paragraph{Step 1: controlling the density ratio.} To establish the convergence rate \eqref{eq:ratio}, a pivotal step is to control the density ratio $\frac{p_{Y_{t-1}}(x_{t-1})}{p_{X_{t-1}}(x_{t-1})}/\frac{p_{Y_{t}}(x_{t})}{p_{X_{t}}(x_{t})}$. Our starting point is the following observation: 
	\begin{align}\label{eq5}
		&\frac{p_{Y_{t-1}}(x_{t-1})}{p_{X_{t-1}}(x_{t-1})}= \frac{p_{\sqrt{\alpha_t}Y_{t-1}}(\sqrt{\alpha_t}x_{t-1})}{p_{\sqrt{\alpha_t}X_{t-1}}(\sqrt{\alpha_t}x_{t-1})}=\frac{p_{\sqrt{\alpha_t}Y_{t-1}}(\sqrt{\alpha_t}x_{t-1})}{p_{Y_{t}}(x_{t})}\left(\frac{p_{\sqrt{\alpha_t}X_{t-1}}(\sqrt{\alpha_t}x_{t-1})}{p_{X_{t}}(x_{t})}\right)^{-1}\frac{p_{Y_{t}}(x_{t})}{p_{X_{t}}(x_{t})}\notag\\
		&= \frac{p_{\sqrt{\alpha_t}Y_{t-1}}(\sqrt{\alpha_t}x_{t-1})}{p_{\sqrt{\alpha_t}Y_{t-1}^{\star}}(\sqrt{\alpha_t}x_{t-1}^{\star})}\cdot\frac{p_{\sqrt{\alpha_t}Y_{t-1}^{\star}}(\sqrt{\alpha_t}x_{t-1}^{\star})}{p_{Y_{t}}(x_{t})}\left(\frac{p_{\sqrt{\alpha_t}X_{t-1}}(\sqrt{\alpha_t}x_{t-1})}{p_{\sqrt{\alpha_t}X_{t-1}}(\sqrt{\alpha_t}x_{t-1}^{\star})}
		\cdot\frac{p_{\sqrt{\alpha_t}X_{t-1}}(\sqrt{\alpha_t}x_{t-1}^{\star})}{p_{X_{t}}(x_{t})}\right)^{-1}\frac{p_{Y_{t}}(x_{t})}{p_{X_{t}}(x_{t})}. 
	\end{align}
By virtue of the defition of $Y_{t-1}^\star$ and the fact that we can transport the distribution of $X_{t}$ to the one of $X_{t-1}$ via \eqref{eq:ODE}, one has
\begin{align}\label{eq6}
	\frac{p_{\sqrt{\alpha_t}Y_{t-1}^{\star}}(\sqrt{\alpha_t}x_{t-1}^{\star})}{p_{Y_{t}}(x_{t})} = \frac{p_{\sqrt{\alpha_t}X_{t-1}}(\sqrt{\alpha_t}x_{t-1}^{\star})}{p_{X_{t}}(x_{t})}.
\end{align}
In addition, the following lemma provides upper bounds for $\frac{p_{\sqrt{\alpha_t}X_{t-1}}(\sqrt{\alpha_t}x_{t-1})}{p_{\sqrt{\alpha_t}X_{t-1}}(\sqrt{\alpha_t}x_{t-1}^{\star})}$ and $\frac{p_{\sqrt{\alpha_t}Y_{t-1}}(\sqrt{\alpha_t}x_{t-1})}{p_{\sqrt{\alpha_t}Y_{t-1}^{\star}}(\sqrt{\alpha_t}x_{t-1}^{\star})}$, respectively.
\begin{lemma}\label{lemma:density_ratio_1}
    We have
    \begin{align}\label{ineq18}
        \frac{p_{\sqrt{\alpha_t}X_{t-1}}(\sqrt{\alpha_t}x_{t-1})}{p_{\sqrt{\alpha_t}X_{t-1}}(\sqrt{\alpha_t}x_{t-1}^{\star})} = \exp\bigg(O\bigg(\frac{\|x_{t-1} - x_{t-1}^{\star}\|_2^2}{1 - \overline{\alpha}_{t-1}} + \sqrt{\frac{d\|x_{t-1} - x_{t-1}^{\star}\|_2^2\log T}{1 - \overline{\alpha}_{t-1}}}\bigg)\bigg).
    \end{align}
    If, furthermore, $\bigg\|\frac{\partial x_{\tau_{t, K-1}}^{(N)}/\sqrt{1-\tau_{t, K-1}}}{\partial x_{\tau_{t, 0}}/\sqrt{1-\tau_{t, 0}}}\bigg\|^{-1} \lesssim 1$, then
    \begin{align}
	\frac{p_{\sqrt{\alpha_t}Y_{t-1}}(\sqrt{\alpha_t}x_{t-1})}{p_{\sqrt{\alpha_t}Y_{t-1}^{\star}}(\sqrt{\alpha_t}x_{t-1}^{\star})}
	&= \exp\bigg(O\bigg(d\bigg\|\frac{\partial x_{\tau_{t, K-1}}^{\star}/\sqrt{1-\tau_{t, K-1}}}{\partial x_{\tau_{t, 0}}^{\star}/\sqrt{1-\tau_{t, 0}}} - \frac{\partial x_{\tau_{t, K-1}}^{(N)}/\sqrt{1-\tau_{t, K-1}}}{\partial x_{\tau_{t, 0}}/\sqrt{1-\tau_{t, 0}}}\bigg\|\bigg)\bigg).\label{ineq39}
\end{align}
\end{lemma}

The proof of Lemma \ref{lemma:density_ratio_1} is postponed to Section \ref{sec:proof_lemma_density_ratio_1}. Combining \eqref{eq5}, \eqref{eq6}, \eqref{ineq18} and \eqref{ineq39}, we know that if $\bigg\|\frac{\partial x_{\tau_{t, K-1}}^{(N)}/\sqrt{1-\tau_{t, K-1}}}{\partial x_{\tau_{t, 0}}/\sqrt{1-\tau_{t, 0}}}\bigg\|^{-1} \lesssim 1$, then
we have
\begin{align}
	\frac{p_{Y_{t-1}}(x_{t-1})}{p_{X_{t-1}}(x_{t-1})}
	= \frac{p_{\sqrt{\alpha_t}Y_{t-1}}(\sqrt{\alpha_t}x_{t-1})}{p_{\sqrt{\alpha_t}X_{t-1}}(\sqrt{\alpha_t}x_{t-1})}
	= \exp(O(\zeta_t(x_t)))\frac{p_{Y_{t}}(x_{t})}{p_{X_{t}}(x_{t})},
\end{align}
where
\begin{align}\label{ineq41}
	\zeta_t(x_t) := d\bigg\|\frac{\partial x_{\tau_{t, K-1}}^{\star}/\sqrt{1-\tau_{t, K-1}}}{\partial x_{\tau_{t, 0}}^{\star}/\sqrt{1-\tau_{t, 0}}} - \frac{\partial x_{\tau_{t, K-1}}^{(N)}/\sqrt{1-\tau_{t, K-1}}}{\partial x_{\tau_{t, 0}}/\sqrt{1-\tau_{t, 0}}}\bigg\|
	+ \frac{\|x_{\tau_{t, K-1}}^{\star} - x_{\tau_{t, K-1}}^{(N)}\|_2^2}{1 - \overline{\alpha}_{t-1}} + \sqrt{\frac{d\|x_{\tau_{t, K-1}}^{\star} - x_{\tau_{t, K-1}}^{(N)}\|_2^2\log T}{1 - \overline{\alpha}_{t-1}}}.
\end{align}
The next lemma reveals that $\zeta_t(x_t)$ is small for ``typical'' points.
\begin{lemma}\label{lemma:main}
	We define the events
	\begin{align*}
		\mathcal{E}_t' := \big\{\varepsilon_{\mathsf{Jacob}, t, i}^{(n)}\big(x_{\tau_{t, i}}^{(n)}\big)\log T \le T, \text{ for all }i, n\big\}
	\end{align*}
	and
	\begin{align}
		\mathcal{E}_t &:= \mathcal{E}_t' \cap \{-\log p_{X_{\tau_{t, i}}}\big(\lambda x_{\tau_{t, i}}^{(n)} + (1 - \lambda)x_{\tau_{t, i}}^\star\big) \leq C_4d\log T, \text{ for all }i, n\}\notag\\ &\qquad \cap \{-\log p_{X_{\tau}}(x_{\tau}^{\star}) \leq C_4d\log T\text{ for all }\tau_{t,K-1} \leq \tau \leq \tau_{t, 0}\},\label{def:event_E}
	\end{align}
	where $C_4 > 0$ is a sufficiently large constant.
	Then on the event $\mathcal{E}_t$, for any $0 \leq i \leq K-1$, one has
    \begin{subequations}
    	\begin{align}
    		&\bigg\|\frac{\partial x_{\tau_{t, K-1}}^{(N)}/\sqrt{1-\tau_{t, K-1}}}{\partial x_{\tau_{t, 0}}/\sqrt{1-\tau_{t, 0}}}\bigg\|^{-1} \lesssim 1,\label{ineq40a}\\
    		&\big\|x_{\tau_{t,i}}^{(N)} - x_{\tau_{t,i}}^{\star}\big\|_2^2 \leq C\tau'^{2}\frac{\log^2 T}{T^2}\sum_{0 \leq n \leq N}\sum_{0 \le i < k}\big(\varepsilon_{\mathsf{score}, t, i}^{(n)}(x_{\tau_{t, i}}^{(n)})\big)^2 + 3C\frac{d\tau'\log^3 T}{T^2}\bigg(\frac{d\log^2 T}{T}\bigg)^{2K},\label{ineq40b}\\
    		&\bigg\|\frac{\partial x_{\tau_{t,i}}^{(N)}/\sqrt{1-\tau_{t,i}}}{\partial x_{\tau_{t, 0}}/\sqrt{1-\tau_{t, 0}}} - \frac{\partial x_{\tau_{t,i}}^{\star}/\sqrt{1-\tau_{t,i}}}{\partial x_{\tau_{t, 0}}^{\star}/\sqrt{1-\tau_{t, 0}}}\bigg\|^2\notag\\ &\quad\lesssim \frac{d^3\log^7 T}{T^4}\sum_{i,n}\big(\varepsilon_{\mathsf{score}, t, i}^{(n)}(x_{\tau_{t, i}}^{(n)})\big)^2 + \frac{\log^2 T}{T^2}\tau'^{2}\sum_{i,n}\big(\varepsilon_{\mathsf{Jacobi}, t, i}^{(n)}(x_{\tau_{t, i}}^{(n)})\big)^2 + \Big(\frac{d\log^2 T}{T}\Big)^{2K+2}.\label{ineq40c}
    	\end{align}
    \end{subequations}
\end{lemma}
The proof of Lemma \ref{lemma:main} is deferred to Section \ref{sec:proof_lemma_main}. By virtue of Lemma \ref{lemma:main} and \eqref{ineq41}, one can show that on the event $\mathcal{E}_t \cap \{\xi_t(x_t) \leq c\}$, one has
\begin{align}\label{ineq42}
		&\frac{p_{Y_{t-1}}(x_{t-1})}{p_{X_{t-1}}(x_{t-1})}\notag\\
	&\quad= \exp\left(O\left(\frac{d\log T}{T}\sqrt{\sum_{i,n}\big(\varepsilon_{\mathsf{Jacobi}, t, i}^{(n)}(x_{\tau_{t, i}}^{(n)})\big)^2} + \frac{\sqrt{d\log^3 T}}{T}\sqrt{\sum_{i,n}\big(\varepsilon_{\mathsf{score}, t, i}^{(n)}(x_{\tau_{t, i}}^{(n)})\big)^2} + d\bigg(\frac{d\log^2 T}{T}\bigg)^{K+1}\right)\right)\frac{p_{Y_{t}}(x_{t})}{p_{X_{t}}(x_{t})}\notag\\
	&\quad= \exp\left(O\left(\xi_t(x_t) + d\bigg(\frac{d\log^2 T}{T}\bigg)^{K+1}\right)\right)\frac{p_{Y_{t}}(x_{t})}{p_{X_{t}}(x_{t})},
\end{align}
provided that $T \gtrsim d^2\log^2 T$. Here, $\xi_t(\cdot)$ is defined in \eqref{eq9a}. 
\begin{remark}
	It is worth noting that Lemma \ref{lemma:main} and \eqref{ineq42} play a crucial role in deriving the desired convergence rate, and establishing them forms one of the most important components of our proof. To control the approximation errors $\|x_{\tau_{t, K-1}}^{\star} - x_{\tau_{t, K-1}}^{(N)}\|_2^2$ and $\bigg\|\frac{\partial x_{\tau_{t, K-1}}^{\star}/\sqrt{1-\tau_{t, K-1}}}{\partial x_{\tau_{t, 0}}^{\star}/\sqrt{1-\tau_{t, 0}}} - \frac{\partial x_{\tau_{t, K-1}}^{(N)}/\sqrt{1-\tau_{t, K-1}}}{\partial x_{\tau_{t, 0}}/\sqrt{1-\tau_{t, 0}}}\bigg\|$, we prove sharp bounds on high-order derivatives $\bigg\|\frac{\partial^k}{\partial \tau^k} \frac{s_{\tau}^{\star}(x_{\tau}^{\star})}{(1-\tau)^{3/2}}\bigg\|_2$ for any fixed $k$, and show that the successive refinements $\{x_{\tau_{t, K-1}}^{(n)}\}_{n \geq 1}$ contract at the desired rates for $n \lesssim \log T$. Detailed steps are deferred to the supplementary material due to space limitations.
\end{remark}
Furthermore, the following lemma shows that the log densities $\{-\log p_{X_{\tau_{t,i}}}\big(\lambda x_{\tau_{t,i}}^{(n+1)} + (1-\lambda)x_{\tau_{t,i}}^\star\big)\}_{i,n, \lambda \in [0,1]}$ can be well-controlled as long as $x_{\tau_{t,0}}$ is ``typical" and the score estimation errors are not extremely large.
\begin{lemma}\label{lemma:log_density_ratio_1}
	If $C_{10}\frac{\theta_{\tau_{t,0}}(x_{\tau_{t,0}})d\log^2 T + \sqrt{\theta_{\tau_{t,0}}(x_{\tau_{t,0}})\sum_{i,n}\big(\varepsilon_{\mathsf{score}, t, i}^{(n)}(x_{\tau_{t, i}}^{(n)})\big)^2d\log^3 T}}{T} \leq 1$ for some large enough constant $C_{10} > 0$, then for all $0 \leq i \leq K - 1, 0 \leq n \leq N - 1$, one has
	\begin{subequations}
		\begin{align}
			-\log p_{X_{\tau_{t,i}}}\big(\lambda x_{\tau_{t,i}}^{(n+1)} + (1-\lambda)x_{\tau_{t,i}}^\star\big) &\leq 2.1d\theta_t\log T,\label{ineq38a}\\
			\log\frac{p_{\sqrt{\frac{1-\tau_{t,0}}{1-\tau_{t,i}}}X_{\tau_{t,i}}}\left(\sqrt{\frac{1-\tau_{t,0}}{1-\tau_{t,i}}}x_{\tau_{t,i}}^{(n+1)}\right)}{p_{X_{\tau_{t,0}}}(x_{\tau_{t,0}})} &\leq \frac{4c_1d\log T}{T} + C_{10}\left\{\frac{d^2\theta_t^2\log^4 T}{T^2} + \frac{\sqrt{d\theta_t\log^3 T}}{T}\sqrt{\sum_{i}\big(\varepsilon_{\mathsf{score}, t, i}^{(n)}(x_{\tau_{t, i}}^{(n)})\big)^2}\right\}.\label{ineq38b}
		\end{align}
	\end{subequations}
\end{lemma}
\paragraph{Step 2: decomposing the TV distance of interest.}
Next, let us  define the sets
\begin{align}\label{event:typical}
	\mathcal{E} = \{x: q_1(x) > \max\{p_1(x), \exp(-c_6d\log T)\}\},
\end{align}
and
\begin{align}\label{set:typical_1}
	\mathcal{I}_1 := \{x_T\ | \ S_T(x_T) \leq c_3\}
\end{align}
for some small enough constant $c_3 > 0$.
Then we learn from \citet[Eqn. (51) and (53)]{li2024sharp} that 
\begin{align}\label{ineq43}
	\mathsf{TV}(q_1, p_1) &\leq \mathbb{E}_{Y_1 \in p_1}\left[\Big(\frac{q_1(Y_1)}{p_1(Y_1)} - 1\Big)\ind\{Y_1 \in \mathcal{E}\}\right] + \exp(-c_6d\log T)\notag\\
	&= \underbrace{\mathbb{E}_{Y_T \in p_T}\left[\Big(\frac{q_1(Y_1)}{p_1(Y_1)} - 1\Big)\ind\{Y_1 \in \mathcal{E}, Y_T \in \mathcal{I}_1\}\right]}_{=:\alpha_1} + \underbrace{\mathbb{E}_{Y_T \in p_T}\left[\Big(\frac{q_1(Y_1)}{p_1(Y_1)} - 1\Big)\ind\{Y_1 \in \mathcal{E}, Y_T \notin \mathcal{I}_1\}\right]}_{=:\alpha_2}\notag\\
	&\quad + \exp(-c_6d\log T).
\end{align}
provided that $c_6 \geq 4(c_R + 2)$. This motivates us to bound $\alpha_1$ and $\alpha_2$ separately.

\paragraph{Step 3: bounding $\alpha_1$.}
We let $\tau(x_T)$ denote the following quantity:
\begin{align}\label{def:tau}
	\tau(x_T) := \max\{2 \leq t \leq T + 1: S_{t-1}(x_T) \leq c_{3}\}.
\end{align}
Then for any $x_T \in \mathcal{I}_1$, one has
\begin{align}\label{ineq44}
	\tau(x_T) = T + 1.
\end{align}
In addition, we have the following two important properties regarding $\tau(x_T)$:
\begin{lemma}\label{lemma:log_density_bound}
	If $-\log q_1(x_1) \leq c_6d\log T$, then for all $1 \leq \ell < \tau(x_T)$, one has
	\begin{align}\label{ineq:log_density_bound}
		-\log q_{\ell}(x_{\ell}) \leq 2c_6d\log T
	\end{align}
	as long as $c_6 > 5c_1$.
\end{lemma}
\begin{lemma}\label{lemma:log_density_ratio_2}
    For $\tau = \tau(x_{T})$, We have
    \begin{align}
    	\frac{q_1(x_{1})}{p_1(x_{1})}
    	&= \left(1 + O\left(\sum_{t < \tau}\xi_t(x_t) + d^2\log^2 T\bigg(\frac{d\log^2 T}{T}\bigg)^{K}\right)\right)\frac{q_{\tau-1}(x_{\tau-1})}{p_{\tau-1}(x_{\tau-1})},\label{ineq45a}\\
    	&\qquad\frac{q_{\ell}(x_{\ell})}{2p_{\ell}(x_{\ell})} \leq \frac{q_1(x_{1})}{p_1(x_{1})} \leq \frac{2q_{\ell}(x_{\ell})}{p_{\ell}(x_{\ell})}, \quad \forall \ell < \tau.\label{ineq45b}
    \end{align}
\end{lemma}
The proofs of Lemma \ref{lemma:log_density_bound} and Lemma \ref{lemma:log_density_ratio_2} are postponed to Sections \ref{sec:proof_lemma_log_density_bound} and \ref{sec:proof_lemma_log_density_ratio_2}, respectively.
By virtue of \eqref{ineq44} and Lemma \ref{lemma:log_density_ratio_2}, one can show that
\begin{align}
	&\mathbb{E}_{Y_T \in p_T}\left[\Big(\frac{q_1(Y_1)}{p_1(Y_1)} - 1\Big)\ind\{Y_1 \in \mathcal{E}, Y_T \in \mathcal{I}_1\}\right]\notag\\
	&\quad = \mathbb{E}_{Y_T \in p_T}\left[\left(\left(1 + \sum_{t}\xi_t(Y_t) + d^2\log^2 T\bigg(\frac{d\log^2 T}{T}\bigg)^{K}\right)\frac{q_T(Y_T)}{p_T(Y_T)} - 1\right)\ind\{Y_1 \in \mathcal{E}, Y_T \in \mathcal{I}_1\}\right]\notag\\
	&\quad = \int\left\{\left(\left(1 + \sum_{t}\xi_t(x_t) + d^2\log^2 T\bigg(\frac{d\log^2 T}{T}\bigg)^{K}\right)q_T(x_T) - p_T(x_T)\right)\ind\{x_1 \in \mathcal{E}, x_T \in \mathcal{I}_1\}\right\}{\rm d}x_T\notag\\
	&\quad \leq \int|q_T(x_T) - p_T(x_T)|{\rm d}x_T + d^2\log^2 T\bigg(\frac{d\log^2 T}{T}\bigg)^{K} +  \int S_T(x_T)q_T(x_T)\ind\{x_1 \in \mathcal{E}, x_T \in \mathcal{I}_1\}{\rm d}x_T.\label{ineq46}
\end{align}
Additionally, we make the observation that 
\begin{align}
	&\int S_T(x_T)q_T(x_T)\ind\{x_1 \in \mathcal{E}, x_T \in \mathcal{I}_1\}{\rm d}x_T\notag\\
	&\quad = \sum_{t=2}^{T}\int \frac{\log T}{T}\left(d\sqrt{\sum_{i,n}\big(\varepsilon_{\mathsf{Jacobi}, t, i}^{(n)}(x_{\tau_{t, i}}^{(n)})\big)^2} + \sqrt{d\log T}\sqrt{\sum_{i,n}\big(\varepsilon_{\mathsf{score}, t, i}^{(n)}(x_{\tau_{t, i}}^{(n)})\big)^2}\right)\ind\{x_1 \in \mathcal{E}, x_T \in \mathcal{I}_1\}{\rm d}x_T\notag\\
	&\quad = \sum_{t=2}^{T}\mathbb{E}_{Y_T \sim p_T}\left[\frac{\log T}{T}\left(d\varepsilon_{\mathsf{Jacobi}, t}(Y_t) + \sqrt{d\log T}\varepsilon_{\mathsf{score}, t}(Y_t)\right)\frac{q_t(Y_T)}{p_t(Y_T)}\ind\{x_1 \in \mathcal{E}, x_T \in \mathcal{I}_1\}\right]\notag\\
	&\quad \stackrel{\eqref{ineq45b}}{\leq} 4\sum_{t=2}^{T}\mathbb{E}_{Y_T \sim p_T}\left[\frac{\log T}{T}\left(d\varepsilon_{\mathsf{Jacobi}, t}(Y_t) + \sqrt{d\log T}\varepsilon_{\mathsf{score}, t}(Y_t)\right)\frac{q_t(Y_t)}{p_t(Y_t)}\ind\{x_1 \in \mathcal{E}, x_T \in \mathcal{I}_1\}\right]\notag\\
	&\quad \leq 4\sum_{t=2}^{T}\mathbb{E}_{Y_t \sim p_t}\left[\frac{\log T}{T}\left(d\varepsilon_{\mathsf{Jacobi}, t}(Y_t) + \sqrt{d\log T}\varepsilon_{\mathsf{score}, t}(Y_t)\right)\frac{q_t(Y_t)}{p_t(Y_t)}\right]\notag\\
	&\quad = 4\sum_{t=2}^{T}\mathbb{E}_{Y_t \sim q_t}\left[\frac{\log T}{T}\left(d\varepsilon_{\mathsf{Jacobi}, t}(Y_t) + \sqrt{d\log T}\varepsilon_{\mathsf{score}, t}(Y_t)\right)\right]\notag\\
	&\quad \stackrel{\eqref{ineq47a}~\text{and}~\eqref{ineq47b}}{\lesssim} d\log^{3/2}T\varepsilon_{\mathsf{Jacobi}} + \sqrt{d\log^4 T}\varepsilon_{\mathsf{score}}.
\end{align}
Putting the previous two inequalities together, one arrives at
\begin{align}
	\alpha_1 &= \mathbb{E}_{Y_T \in p_T}\left[\Big(\frac{q_1(Y_1)}{p_1(Y_1)} - 1\Big)\ind\{Y_1 \in \mathcal{E}, Y_T \in \mathcal{I}_1\}\right]\notag\\ &\lesssim d^2\log^2 T\bigg(\frac{d\log^2 T}{T}\bigg)^{K} + d\log^{3/2}T\varepsilon_{\mathsf{Jacobi}} + \sqrt{d\log^4 T}\varepsilon_{\mathsf{score}}.\label{ineq:alpha_1}
\end{align}

\paragraph{Step 4: bounding $\alpha_2$.} Now, we turn to bounding $\alpha_2$. We let $\tau = \tau(x_T)$ define the following sets:
\begin{align}
    \mathcal{I}_2 &:= \{x_T~|~c_3 \leq S_{\tau}(x_T) \leq 2c_3\},\label{set_I_2}\\
    \mathcal{I}_3 &:= \left\{x_T~|~S_{\tau-1}(x_T) \leq c_3, \xi_{\tau}(x_T) \geq c_3, \frac{q_{\tau-1}(x_{\tau-1})}{p_{\tau-1}(x_{\tau-1})} \leq 8\frac{q_{\tau}(x_{\tau})}{p_{\tau}(x_{\tau})}\right\},\label{set_I_3}\\
    \mathcal{I}_4 &:= \left\{x_T~|~S_{\tau-1}(x_T) \leq c_3, \xi_{\tau}(x_T) \geq c_3, \frac{q_{\tau-1}(x_{\tau-1})}{p_{\tau-1}(x_{\tau-1})} > 8\frac{q_{\tau}(x_{\tau})}{p_{\tau}(x_{\tau})}\right\},\label{set_I_4}
\end{align}
It is straightforward to verify that $\mathcal{I}_1 \cup \mathcal{I}_2 \cup \mathcal{I}_3 \cup \mathcal{I}_4 = \mathbb{R}^d$. The following lemma indicates that $\alpha_2$ can be well-controlled.
\begin{lemma}\label{lemma:other_set}
    The following inequality holds:
    \begin{align}
        \mathbb{E}_{Y_T \sim p_T}\left[\Big(\frac{q_1(Y_1)}{p_1(Y_1)} - 1\Big)\ind\{Y_1 \in \mathcal{E}, Y_T \in \mathcal{I}_2 \cup \mathcal{I}_3 \cup \mathcal{I}_4\}\right] \lesssim d^2\log^2 T\bigg(\frac{d\log^2 T}{T}\bigg)^{K} + d\log^{3/2}T\varepsilon_{\mathsf{Jacobi}} + \sqrt{d\log^4 T}\varepsilon_{\mathsf{score}}.\label{ineq48}
    \end{align}
\end{lemma}
An immediate consequence of Lemma \ref{lemma:other_set} is
\begin{align}
    \alpha_2 &= \mathbb{E}_{Y_T \in p_T}\left[\Big(\frac{q_1(Y_1)}{p_1(Y_1)} - 1\Big)\ind\{Y_1 \in \mathcal{E}, Y_T \notin \mathcal{I}_1\}\right]\notag\\
    &\leq \mathbb{E}_{Y_T \in p_T}\left[\Big(\frac{q_1(Y_1)}{p_1(Y_1)} - 1\Big)\ind\{Y_1 \in \mathcal{E}, Y_T \in \mathcal{I}_2 \cup \mathcal{I}_3 \cup \mathcal{I}_4\}\right]\notag\\
    &\lesssim d^2\log^2 T\bigg(\frac{d\log^2 T}{T}\bigg)^{K} + d\log^{3/2}T\varepsilon_{\mathsf{Jacobi}} + \sqrt{d\log^4 T}\varepsilon_{\mathsf{score}}.\label{ineq:alpha_2}
\end{align}
Here, the second line comes from the facts $q_1(x) \geq p_1(x)$ for any $x \in \mathcal{E}$ and $\mathcal{I}_1^c \subseteq \mathcal{I}_2 \cup \mathcal{I}_3 \cup \mathcal{I}_4$.

\paragraph{Step 5: bounding ${\sf TV}(p_1, q_1)$ } Combining \eqref{ineq43}, \eqref{ineq:alpha_1} and \eqref{ineq:alpha_2}, one arrives at the advertised result: 
\begin{align}
    {\sf TV}(p_1, q_1) &\lesssim d^2\log^2 T\bigg(\frac{d\log^2 T}{T}\bigg)^{K} + d\log^{3/2}T\varepsilon_{\mathsf{Jacobi}} + \sqrt{d\log^4 T}\varepsilon_{\mathsf{score}} + \exp(-c_6d\log T)\notag\\
    &\asymp d^2\log^2 T\bigg(\frac{d\log^2 T}{T}\bigg)^{K} + d\log^{3/2}T\varepsilon_{\mathsf{Jacobi}} + \sqrt{d\log^4 T}\varepsilon_{\mathsf{score}}.\label{ineq49}
\end{align}

\section{Discussion and future directions}

In this work, we have proposed an algorithm to speed up data generation in diffusion models, and have contributed towards a new theoretical framework for understanding the efficacy of higher-order ODE-based sampling methods. The proposed algorithm exploits $K$-th order ODE approximation combined with a successive refining scheme, which provably yields  $\varepsilon$-precision (measured by the TV distance between the output distribution and the target distribution) within $\widetilde{O}\big(d^{1+2/K}/\varepsilon^{1/K}\big)$ iterations. Our theoretical guarantees hold under fairly mild assumptions on the target distribution, with no restrictions on smoothness and log-concavity, and are capable of accommodating imperfect score estimation.

Our findings leave open several interesting questions for future investigation.  
For instance, while our theory allows the order $K$ to be an arbitrary large constant,  it is worth exploring whether $K$ can be allowed to scale with $T$ (e.g., whether $K$ can reach the order of $\log T$), as accomplishing this might potentially lead to further acceleration. 
Additionally, our method is a deterministic sampler designed to speed up diffusion ODE, 
and it remains unclear whether a similar iteration complexity can be achieved with an accelerated stochastic sampler (i.e., the SDE counterpart). Furthermore, 
while this work has focused on the iteration complexity of the sampling phase,  how to develop an end-to-end theory that also incorporates the score matching phase forms another direction for future study.  
Finally, when the target data distribution exhibits intrinsic low dimensionality, it would be interesting to examine whether the sampling process is further expedited.

\section*{Acknowledgments}

G.~Li is supported in part by the Chinese University of Hong Kong Direct Grant for Research and the Hong Kong Research Grants Council ECS 2191363.
Y.~Wei is supported in part by the NSF grants CCF-2106778, CCF-2418156 and 
CAREER award DMS-2143215. 
Y.~Chen is supported in part by the Alfred P.~Sloan Research Fellowship,  the ONR grants N00014-22-1-2354 and N00014-25-1-2344,  the NSF grants 2221009 and 2218773, 
the Wharton AI \& Analytics Initiative's AI Research Fund, 
and the Amazon Research Award.

%\newpage
\appendix

\section{Proof of Lemma \ref{lemma:main}}\label{sec:proof_lemma_main}
For $\tau_{t-1,0} \leq \tau < \tau_{t, 0}$, we let $x_{\tau}^\star$ denote the solution of \eqref{eq:ODE} with the initial condition $x_{\tau_{t, 0}}^\star = x_{\tau_{t, 0}}$. We let
\begin{align*} 
	\xi_{\mathsf{score}, \tau'}^{(n)} &:= \frac{x_{\tau'}^{(n+1)}}{\sqrt{1-\tau'}} - \frac{x_{\tau_{t, 0}}}{\sqrt{1-\tau_{t, 0}}} - \sum_{0 \le i < K} \gamma_{t, i}(\tau')(1-\tau_{t, i})^{-3/2}s_{\tau_{t, i}}^{\star}(x_{\tau_{t, i}}^{(n)}) \\
	&= \sum_{0 \le i < K} \gamma_{t, i}(\tau')(1-\tau_{t, i})^{-3/2}\big[s_{\tau_{t, i}}(x_{\tau_{t, i}}^{(n)}) - s_{\tau_{t, i}}^{\star}(x_{\tau_{t, i}}^{(n)})\big].
\end{align*}
Then one can derive
\begin{align}
	&\frac{x_{\tau'}^{(n+1)}}{\sqrt{1-\tau'}} - \frac{x_{\tau'}^{\star}}{\sqrt{1-\tau'}}\notag\\ &\quad= \Big(\frac{x_{\tau'}^{(n+1)}}{\sqrt{1-\tau'}} - \frac{x_{\tau_{t, 0}}}{\sqrt{1-\tau_{t, 0}}}\Big) - \Big(\frac{x_{\tau'}^{\star}}{\sqrt{1-\tau'}}-\frac{x_{\tau_{t, 0}}^{\star}}{\sqrt{1-\tau_{t, 0}}}\Big)\notag\\
	&\quad=\sum_{0 \le i < K} \gamma_{t, i}(\tau')(1-\tau_{t, i})^{-3/2}s_{\tau_{t, i}}(x_{\tau_{t, i}}^{(n)}) + \int_{\tau_{t, 0}}^{\tau'} \frac{1}{2(1-\tau)^{3/2}}s_{\tau}^{\star}\mathrm{d}\tau\notag\\
	&\quad= \sum_{0 \le i < K} \gamma_{t, i}(\tau')(1-\tau_{t, i})^{-3/2}\big[s_{\tau_{t, i}}(x_{\tau_{t, i}}^{(n)}) - s_{\tau_{t, i}}^{\star}(x_{\tau_{t, i}}^{(n)})\big]\notag\\&\qquad + \left[\sum_{0 \le i < K} \gamma_{t, i}(\tau')(1-\tau_{t, i})^{-3/2} s_{\tau_{t, i}}^{\star}(x_{\tau_{t, i}}^{(n)}) - \left(-\int_{\tau_{t, 0}}^{\tau'} \frac{1}{2(1-\tau)^{3/2}}s_{\tau}^{\star}(x_{\tau}^\star)\mathrm{d}\tau\right)\right]\notag\\
	&\quad= \xi_{\mathsf{score}, \tau'}^{(n)} + \sum_{0 \le i < K} \gamma_{t, i}(\tau')(1-\tau_{t, i})^{-3/2}\big(s_{\tau_{t, i}}^{\star}(x_{\tau_{t, i}}^{\star}) - s_{\tau_{t, i}}^{\star}(x_{\tau_{t, i}}^{(n)})\big)\notag\\&\qquad + \left[\sum_{0 \le i < K} \gamma_{t, i}(\tau')(1-\tau_{t, i})^{-3/2} s_{\tau_{t, i}}^{\star}(x_{\tau_{t, i}}^{\star}) - \left(-\int_{\tau_{t, 0}}^{\tau'} \frac{1}{2(1-\tau)^{3/2}}s_{\tau}^{\star}(x_{\tau}^\star)\mathrm{d}\tau\right)\right].\label{eq:decomposition_1}
\end{align}
Furthermore, we make the observation that
\begin{align}
	\frac{\partial x_{\tau'}^{(n+1)}/\sqrt{1-\tau'}}{\partial x_{\tau_{t, 0}}/\sqrt{1-\tau_{t, 0}}}
	&= I + \sum_{0 \le i < K} \gamma_{t, i}(\tau')(1-\tau_{t, i})^{-3/2}\frac{\partial s_{\tau_{t, i}}(x_{\tau_{t, i}}^{(n)})}{\partial x_{\tau_{t, i}}^{(n)}/\sqrt{1-\tau_{t, i}}}\frac{\partial x_{\tau_{t, i}}^{(n)}/\sqrt{1-\tau_{t, i}}}{\partial x_{\tau_{t, 0}}/\sqrt{1-\tau_{t, 0}}} \notag\\
	&=: I + \sum_{0 \le i < K} \gamma_{t, i}(\tau')(1-\tau_{t, i})^{-3/2}\frac{\partial s_{\tau_{t, i}}^{\star}(x_{\tau_{t, i}}^{(n)})}{\partial x_{\tau_{t, i}}^{(n)}/\sqrt{1-\tau_{t, i}}}\frac{\partial x_{\tau_{t, i}}^{(n)}/\sqrt{1-\tau_{t, i}}}{\partial x_{\tau_{t, 0}}/\sqrt{1-\tau_{t, 0}}} + \xi_{\mathsf{Jacob}, \tau'}^{(n)}.\label{ineq5}
\end{align} 
and
\begin{align*}
	\frac{\partial x_{\tau'}^{\star}/\sqrt{1-\tau'}}{\partial x_{\tau_{t, 0}}^{\star}/\sqrt{1-\tau_{t, 0}}} = I - \int_{\tau_{t, 0}}^{\tau'} \frac{1}{2(1-\tau)^{3/2}}\frac{\partial s_{\tau}^{\star}(x_{\tau}^{\star})}{\partial x_{\tau}^{\star}/\sqrt{1-\tau}}\frac{\partial x_{\tau}^{\star}/\sqrt{1-\tau}}{\partial x_{\tau_{t, 0}}^{\star}/\sqrt{1-\tau_{t, 0}}}\mathrm{d}\tau.
\end{align*}
Combining the previous two equations, we have
\begin{align}
	&\frac{\partial x_{\tau'}^{(n+1)}/\sqrt{1-\tau'}}{\partial x_{\tau_{t, 0}}/\sqrt{1-\tau_{t, 0}}}-\frac{\partial x_{\tau'}^{\star}/\sqrt{1-\tau'}}{\partial x_{\tau_{t, 0}}^{\star}/\sqrt{1-\tau_{t, 0}}}\notag\\
	&\quad= \xi_{\mathsf{Jacob}, \tau'}^{(n)} + \sum_{0 \le i < K} \gamma_{t, i}(\tau')(1-\tau_{t, i})^{-3/2}\frac{\partial s_{\tau_{t, i}}^{\star}(x_{\tau_{t, i}}^{(n)})}{\partial x_{\tau_{t, i}}^{(n)}/\sqrt{1-\tau_{t, i}}}\frac{\partial x_{\tau_{t, i}}^{(n)}/\sqrt{1-\tau_{t, i}}}{\partial x_{\tau_{t, 0}}/\sqrt{1-\tau_{t, 0}}}\notag\\&\qquad+\int_{\tau_{t, 0}}^{\tau'} \frac{1}{2(1-\tau)^{3/2}}\frac{\partial s_{\tau}^{\star}(x_{\tau}^{\star})}{\partial x_{\tau}^{\star}/\sqrt{1-\tau}}\frac{\partial x_{\tau}^{\star}/\sqrt{1-\tau}}{\partial x_{\tau_{t, 0}}^{\star}/\sqrt{1-\tau_{t, 0}}}\mathrm{d}\tau.\label{eq:decomposition_2}
\end{align}

\paragraph{Controlling estimation error caused by score estimation.}
Cauchy-Schwarz inequality tells us that
\begin{align}
	\|\xi_{\mathsf{score}, \tau'}^{(n)}\|_2^2 &\leq \bigg(\sum_{0 \le i < K} \gamma_{t, i}(\tau')(1-\tau_{t, i})^{-3/2}\varepsilon_{\mathsf{score}, t, i}^{(n)}(x_{\tau_{t, i}}^{(n)})\bigg)^2\notag\\
	&\leq K\sum_{0 \le i < K} \gamma_{t, i}^2(\tau')(1-\tau_{t, i})^{-3}\big\|s_{\tau_{t, i}}(x_{\tau_{t, i}}^{(n)}) - s_{\tau_{t, i}}^{\star}(x_{\tau_{t, i}}^{(n)})\big\|_2^2\notag\\
	&\leq K\sum_{0 \le i < K}2^{2K} (\tau'-\tau_{t, 0})^2(1-\tau_{t, i})^{-3}\big\|s_{\tau_{t, i}}(x_{\tau_{t, i}}^{(n)}) - s_{\tau_{t, i}}^{\star}(x_{\tau_{t, i}}^{(n)})\big\|_2^2\notag\\
	& \asymp \sum_{0 \le i < K}(\tau'-\tau_{t, 0})^2(1-\tau_{t, i})^{-3}\big\|s_{\tau_{t, i}}(x_{\tau_{t, i}}^{(n)}) - s_{\tau_{t, i}}^{\star}(x_{\tau_{t, i}}^{(n)})\big\|_2^2\label{ineq:score_estimation_1}.
\end{align}
here, the penultimate line makes use of \eqref{ineq:learning_rate_g} and the last line holds since $K$ is a constant.

We claim that on the event $\mathcal{E}_t$, for all $0 \leq n \leq N$, we have
\begin{subequations}
	\begin{align}
		\bigg\|\frac{\partial x_{\tau'}^{(n)}/\sqrt{1-\tau'}}{\partial x_{\tau_{t, 0}}/\sqrt{1-\tau_{t, 0}}}\bigg\| &\asymp 1,\label{ineq3a}\\
		\|\xi_{\mathsf{Jacob}, \tau'}^{(n)}\| &\le \frac{1}{2}.\label{ineq3b}
	\end{align}
\end{subequations}
If \eqref{ineq3a} and \eqref{ineq3b} hold, then on the event $\mathcal{E}_t$, we have
\begin{align}
	\|\xi_{\mathsf{Jacob}, \tau'}^{(n)}\|^2
	&= \bigg\|\sum_{0 \le i < K} \gamma_{t, i}(\tau')(1-\tau_{t, i})^{-3/2}\bigg(\frac{\partial s_{\tau_{t, i}}(x_{\tau_{t, i}}^{(n)})}{\partial x_{\tau_{t, i}}^{(n)}/\sqrt{1-\tau_{t, i}}} - \frac{\partial s_{\tau_{t, i}}^{\star}(x_{\tau_{t, i}}^{(n)})}{\partial x_{\tau_{t, i}}^{(n)}/\sqrt{1-\tau_{t, i}}}\bigg)\frac{\partial x_{\tau_{t, i}}^{(n)}/\sqrt{1-\tau_{t, i}}}{\partial x_{\tau_{t, 0}}/\sqrt{1-\tau_{t, 0}}}\bigg\|^2\notag\\
	&\le \bigg(\sum_{0 \le i < K} \gamma_{t, i}(\tau')(1-\tau_{t, i})^{-3/2}\bigg\|\frac{\partial s_{\tau_{t, i}}(x_{\tau_{t, i}}^{(n)})}{\partial x_{\tau_{t, i}}^{(n)}/\sqrt{1-\tau_{t, i}}} - \frac{\partial s_{\tau_{t, i}}^{\star}(x_{\tau_{t, i}}^{(n)})}{\partial x_{\tau_{t, i}}^{(n)}/\sqrt{1-\tau_{t, i}}}\bigg\|\bigg\|\frac{\partial x_{\tau_{t, i}}^{(n)}/\sqrt{1-\tau_{t, i}}}{\partial x_{\tau_{t, 0}}/\sqrt{1-\tau_{t, 0}}}\bigg\|\bigg)^2\notag\\
	&\lesssim \bigg(\sum_{0 \le i < K} \gamma_{t, i}(\tau')(1-\tau_{t, i})^{-3/2}\bigg\|\frac{\partial s_{\tau_{t, i}}(x_{\tau_{t, i}}^{(n)})}{\partial x_{\tau_{t, i}}^{(n)}/\sqrt{1-\tau_{t, i}}} - \frac{\partial s_{\tau_{t, i}}^{\star}(x_{\tau_{t, i}}^{(n)})}{\partial x_{\tau_{t, i}}^{(n)}/\sqrt{1-\tau_{t, i}}}\bigg\|\bigg)^2\notag\\
	&= \bigg(\sum_{0 \le i < K} \gamma_{t, i}(\tau')(1-\tau_{t, i})^{-1}\varepsilon_{\mathsf{Jacobi}, t, i}^{(n)}(x_{\tau_{t, i}}^{(n)})\bigg)^2\notag\\
	&\stackrel{\rm{Cauchy-Schwarz}}{\leq} K\sum_{0 \le i < K} \gamma_{t, i}^2(\tau')(1-\tau_{t, i})^{-2}\big(\varepsilon_{\mathsf{Jacobi}, t, i}^{(n)}(x_{\tau_{t, i}}^{(n)})\big)^2\notag\\
	&\stackrel{\eqref{ineq:learning_rate_g}}{\lesssim} \sum_{0 \le i < K}(\tau'-\tau_{t, 0})^2(1-\tau_{t, i})^{-2}\big(\varepsilon_{\mathsf{Jacobi}, t, i}^{(n)}(x_{\tau_{t, i}}^{(n)})\big)^2\notag\\
	&\stackrel{\text{Lemma }\ref{lemma:learning_rate}}{\lesssim} \frac{\log^2 T}{T^2}\tau'^{2}\sum_{0 \le i < K}\big(\varepsilon_{\mathsf{Jacobi}, t, i}^{(n)}(x_{\tau_{t, i}}^{(n)})\big)^2,\label{ineq:score_estimation_2}
\end{align}
\begin{comment}
	and consequently we have
	\begin{align*}
		\mathbb{E}\big[\|\xi_{\mathsf{Jacob}, \tau'}^{(n)}\|^2\ind_{\mathcal{E}_t}\big]
		\lesssim \sum_{0 \le i < K} (\tau'-\tau_{t, 0})^2(1-\tau_{t, i})^{-2}(\varepsilon_{\mathsf{Jacob}, t, i}^{(n)})^2 \lesssim \frac{\log^2 T}{T^2}\tau'^{2}\sum_{0 \le i < K}(\varepsilon_{\mathsf{Jacob}, t, i}^{(n)})^2.
	\end{align*}
\end{comment}
\paragraph{Controlling discretization error.}
We claim that if $-p_{\overline{X}_{\tau}}(x_{\tau}^\star) \lesssim d\theta\log T$ for $\theta \geq c_6$, then for any interger $k \geq 0$, there exists a constant $c_k > 0$ depending only on $k$ such that
\begin{subequations}
	
	\begin{align}
		\bigg\|\frac{\partial^k}{\partial \tau^k} \frac{s_{\tau}^{\star}(x_{\tau}^{\star})}{(1-\tau)^{3/2}}\bigg\|_2
		\le c_k \sqrt{\frac{d\theta\log T}{\tau(1-\tau)^3}}\bigg(\frac{d\theta\log T}{\tau(1-\tau)}\bigg)^k,\label{eq:derivative_1}
	\end{align}
	and
	\begin{align}
		\bigg\|\frac{\partial^k}{\partial \tau^k} \frac{1}{(1-\tau)^{3/2}}\frac{\partial s_{\tau}^{\star}(x_{\tau}^{\star})}{\partial x_{\tau}^{\star}/\sqrt{1-\tau}}\frac{\partial x_{\tau}^{\star}/\sqrt{1-\tau}}{\partial x_{\tau_{t, 0}}^{\star}/\sqrt{1-\tau_{t, 0}}}\bigg\|
		\le c_k \bigg(\frac{d\theta\log T}{\tau(1-\tau)}\bigg)^{k+1},\label{eq:derivative_2}
	\end{align}
\end{subequations}
which we prove by induction at the end of this proof.
Then Taylor's Theorem together with the fact that $\tau \asymp \tau_{t,0}$ and $1-\tau \asymp 1-\tau_{t,0}$ for all $\tau' \leq \tau \leq \tau_{t, 0}$ tells us that
\begin{align}
	&\frac{x_{\tau'}^{\star}}{\sqrt{1-\tau'}}\notag\\&\quad= \frac{x_{\tau_{t, 0}}^{\star}}{\sqrt{1-\tau_{t, 0}}} - \int_{\tau_{t, 0}}^{\tau'} \frac{1}{2(1-\tau)^{3/2}}s_{\tau}^{\star}\mathrm{d}\tau\notag\\
	&\quad= \frac{x_{\tau_{t, 0}}^{\star}}{\sqrt{1-\tau_{t, 0}}} + \sum_{0 \le i < K} \gamma_{t, i}(\tau')(1-\tau_{t, i})^{-3/2}s_{\tau_{t, i}}^{\star}(x_{\tau_{t, i}}^{\star}) + O\Bigg(\sqrt{\frac{d(\tau'-\tau_{t, 0})^2\log T}{\tau'(1-\tau')^3}}\bigg(\frac{d(\tau_{t, 0}-\tau')\log T}{\tau'(1-\tau')}\bigg)^{K}\Bigg),\label{eq:taylor_1}
\end{align}
and
\begin{align}
	&\frac{\partial x_{\tau'}^{\star}/\sqrt{1-\tau'}}{\partial x_{\tau_{t, 0}}^{\star}/\sqrt{1-\tau_{t, 0}}}\notag\\
	&\quad= I - \int_{\tau_{t, 0}}^{\tau'} \frac{1}{2(1-\tau)^{3/2}}\frac{\partial s_{\tau}^{\star}(x_{\tau}^{\star})}{\partial x_{\tau}^{\star}/\sqrt{1-\tau}}\frac{\partial x_{\tau}^{\star}/\sqrt{1-\tau}}{\partial x_{\tau_{t, 0}}^{\star}/\sqrt{1-\tau_{t, 0}}}\mathrm{d}\tau\notag\\ 
	&\quad=I + \sum_{0 \le i < K} \gamma_{t, i}(\tau')(1-\tau_{t, i})^{-3/2}\frac{\partial s_{\tau_{t, i}}^{\star}(x_{\tau_{t, i}}^{\star})}{\partial x_{\tau_{t, i}}^{\star}/\sqrt{1-\tau_{t, i}}}\frac{\partial x_{\tau_{t, i}}^{\star}/\sqrt{1-\tau_{t, i}}}{\partial x_{\tau_{t, 0}}^{\star}/\sqrt{1-\tau_{t, 0}}} + O\Bigg(\bigg(\frac{d(\tau_{t, 0}-\tau')\log T}{\tau'(1-\tau')}\bigg)^{K+1}\Bigg).\label{eq:taylor_2}
\end{align}

\paragraph{Putting everything together.}
Combing \eqref{eq:decomposition_1}, \eqref{eq:taylor_1} and Lemma \ref{lemma:Hessian}, one has
\begin{align*}
	&\bigg\|\frac{x_{\tau'}^{(n+1)}}{\sqrt{1-\tau'}} - \frac{x_{\tau'}^{\star}}{\sqrt{1-\tau'}}\bigg\|_2\\
	&\quad\lesssim\|\xi_{\mathsf{score}, \tau'}^{(n)}\|_2 + \sum_{0 \le i < K} |\gamma_{t, i}(\tau')|(1-\tau_{t, i})^{-3/2}\big\|s_{\tau_{t, i}}^{\star}(x_{\tau_{t, i}}^{\star}) - s_{\tau_{t, i}}^{\star}(x_{\tau_{t, i}}^{(n)})\big\|_2\\&\qquad + \left\|\sum_{0 \le i < K} \gamma_{t, i}(\tau')(1-\tau_{t, i})^{-3/2} s_{\tau_{t, i}}^{\star}(x_{\tau_{t, i}}^{(n)}) - \left(-\int_{\tau_{t, 0}}^{\tau'} \frac{1}{2(1-\tau)^{3/2}}s_{\tau}^{\star}(x_{\tau}^\star)\mathrm{d}\tau\right)\right\|_2\\
	&\quad \lesssim\|\xi_{\mathsf{score}, \tau'}^{(n)}\|_2 + \sum_{0 \le i < K}2^K(\tau_{t, 0} - \tau')(1-\tau_{t, i})^{-3/2}\big\|s_{\tau_{t, i}}^{\star}(x_{\tau_{t, i}}^{\star}) - s_{\tau_{t, i}}^{\star}(x_{\tau_{t, i}}^{(n)})\big\|_2\\
	&\qquad + O\Bigg(\sqrt{\frac{d(\tau'-\tau_{t, 0})^2\log T}{\tau'(1-\tau')^3}}\bigg(\frac{d(\tau_{t, 0}-\tau')\log T}{\tau'(1-\tau')}\bigg)^{K}\Bigg) \\
	&\quad\lesssim \sum_{0 \le i < K} \frac{d(\tau_{t, 0}-\tau')\log T}{\tau_{t, i}(1-\tau_{t, i})} \bigg\|\frac{x_{\tau_{t, i}}^{(n)}}{\sqrt{1-\tau'}} - \frac{x_{\tau_{t, i}}^{\star}}{\sqrt{1-\tau'}}\bigg\|_2 \\
	&\qquad+ \|\xi_{\mathsf{score}, \tau'}^{(n)}\|_2 + O\Bigg(\sqrt{\frac{d(\tau'-\tau_{t, 0})^2\log T}{\tau'(1-\tau')^3}}\bigg(\frac{d(\tau_{t, 0}-\tau')\log T}{\tau'(1-\tau')}\bigg)^{K}\Bigg) \\
	&\quad\lesssim \frac{d\log^2 T}{T} \sum_{0 \le i < K} \bigg\|\frac{x_{\tau_{t, i}}^{(n)}}{\sqrt{1-\tau'}} - \frac{x_{\tau_{t, i}}^{\star}}{\sqrt{1-\tau'}}\bigg\|_2 
	+ \|\xi_{\mathsf{score}, \tau'}^{(n)}\|_2 + O\Bigg(\sqrt{\frac{d\tau'\log^3 T}{T^2(1-\tau')}}\bigg(\frac{d\log^2 T}{T}\bigg)^{K}\Bigg),
\end{align*}
Here, the second inequality makes use of \eqref{ineq:learning_rate_g} and Lemma \ref{lemma:learning_rate}; the third inequality is valid due to Lemma \ref{lemma:Hessian}; the last line holds because of Lemma \ref{lemma:learning_rate}. Recalling that $K$ is a constant, we know there exists constants $C > 0$ such that
\begin{align*}
	\big\|x_{\tau'}^{(n+1)} - x_{\tau'}^{\star}\big\|_2^2 &\leq C\left[\left(\frac{d\log^2 T}{T}\right)^2 \sum_{0 \le i < K} \big\|x_{\tau_{t, i}}^{(n)} - x_{\tau_{t, i}}^{\star}\big\|_2^2 + (1 - \tau')\|\xi_{\mathsf{score}, \tau'}^{(n)}\|_2^2 + \frac{d\tau'\log^3 T}{T^2}\bigg(\frac{d\log^2 T}{T}\bigg)^{2K}\right]\\
	&\stackrel{\eqref{ineq:score_estimation_1}}{\leq} \frac{1}{3}\max_{0 \leq i \leq K}\big\|x_{\tau_{t, i}}^{(n)} - x_{\tau_{t, i}}^{\star}\big\|_2^2 + C\tau'^{2}\frac{\log^2 T}{T^2}\sum_{0 \le i < K}\big(\varepsilon_{\mathsf{score}, t, i}^{(n)}(x_{\tau_{t, i}}^{(n)})\big)^2 + C\frac{d\tau'\log^3 T}{T^2}\bigg(\frac{d\log^2 T}{T}\bigg)^{2K},
\end{align*}
Then it follows by induction that, for all $N' \geq 1$,
\begin{align}
	\big\|x_{\tau'}^{(N')} - x_{\tau'}^{\star}\big\|_2^2 &\leq \frac{1}{3^{N'}}\max_{0 \leq i \leq K}\big\|x_{\tau_{t, i}}^{(0)} - x_{\tau_{t, i}}^{\star}\big\|_2^2 + C\tau'^{2}\frac{\log^2 T}{T^2}\sum_{0 \leq n \leq N'}\sum_{0 \le i < K}\big(\varepsilon_{\mathsf{score}, t, i}^{(n)}(x_{\tau_{t, i}}^{(n)})\big)^2\notag\\&\quad+ 2C\frac{d\tau'\log^3 T}{T^2}\bigg(\frac{d\log^2 T}{T}\bigg)^{2K}.\label{ineq:induction_1}
\end{align}
Recalling the definition of $x_{\tau_{t, i}}^{\star}$ and $x_{\tau_{t, i}}^{(0)}$, we invoke Lemma \ref{lemma:Hessian} to obtain
\begin{align}
	\big\|x_{\tau_{t, i}}^{\star} - x_{\tau_{t, i}}^{(0)}\big\|_2 &= \big\|x_{\tau_{t, i}}^{\star} - x_{\tau_{t, 0}}\big\|_2 = \bigg\|\int_{\tau_{t, 0}}^{\tau_{t, i}}s_{\tau}^\star(x_{\tau}^\star){\rm d}\tau\bigg\|_2 \leq (\tau_{t, 0}-\tau_{t, i})\sup_{\tau_{t, i}\leq\tau\leq\tau_{t, 0}}\left\|s_{\tau}^\star(x_{\tau}^\star)\right\|_2\notag\\
	&\lesssim (\tau_{t, 0}-\tau_{t, K-1})\sqrt{\frac{d\log T}{\tau_{t, K-1}}} = (\overline{\alpha}_{t-1} - \overline{\alpha}_t)\sqrt{\frac{d\log T}{1-\overline{\alpha}_{t-1}}} =\overline{\alpha}_{t-1}\sqrt{1-\alpha_t}\sqrt{\frac{1-\alpha_t}{1-\overline{\alpha}_{t-1}}}\sqrt{d\log T}\notag\\
	&\stackrel{\text{Lemma }\ref{lemma:learning_rate}}{\leq} 2c_1\frac{\log T}{T}\sqrt{d\log T}\label{ineq13}
\end{align}
for all $0 \leq i \leq k-1$, provided that $-\log p_{\tau}(x_{\tau}^\star) \lesssim d\log T$ for all $\tau_{t,k-1} \leq \tau \leq \tau_{t, 0}$. As a consequence, as long as $N \geq C_0\log T$ for some large enough constant $C_0 > 0$, one has
\begin{align}\label{ineq12}
	\big\|x_{\tau'}^{(N)} - x_{\tau'}^{\star}\big\|_2^2 &\leq C\tau'^{2}\frac{\log^2 T}{T^2}\sum_{0 \leq n \leq N}\sum_{0 \le i < K}\big(\varepsilon_{\mathsf{score}, t, i}^{(n)}(x_{\tau_{t, i}}^{(n)})\big)^2 + 3C\frac{d\tau'\log^3 T}{T^2}\bigg(\frac{d\log^2 T}{T}\bigg)^{2K}.
\end{align}

Similarly, \eqref{eq:decomposition_2}, \eqref{ineq3a}, \eqref{eq:taylor_2}, Lemma \ref{lemma:learning_rate} and Lemma \ref{lemma:Hessian} together imply that on $\mathcal{E}_t$,
\begin{align*}
	&\bigg\|\frac{\partial x_{\tau'}^{(n+1)}/\sqrt{1-\tau'}}{\partial x_{\tau_{t, 0}}/\sqrt{1-\tau_{t, 0}}} - \frac{\partial x_{\tau'}^{\star}/\sqrt{1-\tau'}}{\partial x_{\tau_{t, 0}}^{\star}/\sqrt{1-\tau_{t, 0}}}\bigg\|\\
	&\quad \lesssim \sum_{0 \le i < K} (\tau_{t, 0}-\tau')(1-\tau_{t, i})^{-3/2}\bigg\|\frac{\partial s_{\tau_{t, i}}^{\star}(x_{\tau_{t, i}}^{\star})}{\partial x_{\tau_{t, i}}^{\star}/\sqrt{1-\tau_{t, i}}}\frac{\partial x_{\tau_{t, i}}^{\star}/\sqrt{1-\tau_{t, i}}}{\partial x_{\tau_{t, 0}}^{\star}/\sqrt{1-\tau_{t, 0}}}
	- \frac{\partial s_{\tau_{t, i}}^{\star}(x_{\tau_{t, i}}^{(n)})}{\partial x_{\tau_{t, i}}^{(n)}/\sqrt{1-\tau_{t, i}}}\frac{\partial x_{\tau_{t, i}}^{(n)}/\sqrt{1-\tau_{t, i}}}{\partial x_{\tau_{t, 0}}/\sqrt{1-\tau_{t, 0}}}\bigg\|\\
	&\qquad+ \|\xi_{\mathsf{Jacob}, \tau'}^{(n)}\|
	+ O\Bigg(\bigg(\frac{d(\tau_{t, 0}-\tau')\log T}{\tau'(1-\tau')}\bigg)^{K+1}\Bigg) \\
	&\quad\lesssim \sum_{0 \le i < K} (\tau_{t, 0}-\tau')(1-\tau_{t, i})^{-3/2}\bigg\|\frac{\partial s_{\tau_{t, i}}^{\star}(x_{\tau_{t, i}}^{\star})}{\partial x_{\tau_{t, i}}^{\star}/\sqrt{1-\tau_{t, i}}}\bigg\|\bigg\|\frac{\partial x_{\tau_{t, i}}^{\star}/\sqrt{1-\tau_{t, i}}}{\partial x_{\tau_{t, 0}}^{\star}/\sqrt{1-\tau_{t, 0}}} - \frac{\partial x_{\tau_{t, i}}^{(n)}/\sqrt{1-\tau_{t, i}}}{\partial x_{\tau_{t, 0}}/\sqrt{1-\tau_{t, 0}}}\bigg\|\\
	&\qquad+\sum_{0 \le i < K} (\tau_{t, 0}-\tau')(1-\tau_{t, i})^{-3/2}\bigg\|\frac{\partial x_{\tau_{t, i}}^{(n)}/\sqrt{1-\tau_{t, i}}}{\partial x_{\tau_{t, 0}}/\sqrt{1-\tau_{t, 0}}}\bigg\|\bigg\|\frac{\partial s_{\tau_{t, i}}^{\star}(x_{\tau_{t, i}}^{\star})}{\partial x_{\tau_{t, i}}^{\star}/\sqrt{1-\tau_{t, i}}} - \frac{\partial s_{\tau_{t, i}}^{\star}(x_{\tau_{t, i}}^{(n)})}{\partial x_{\tau_{t, i}}^{(n)}/\sqrt{1-\tau_{t, i}}}\bigg\|\\
	&\qquad+ \|\xi_{\mathsf{Jacob}, \tau'}^{(n)}\|
	+ O\Bigg(\bigg(\frac{d(\tau_{t, 0}-\tau')\log T}{\tau'(1-\tau')}\bigg)^{K+1}\Bigg) \\
	&\quad\lesssim \sum_{0 \le i < K} (\tau_{t, 0}-\tau')(1-\tau_{t, i})^{-3/2}\sqrt{1-\tau_{t,i}}\frac{d\log T}{\tau_{t,i}}\bigg\|\frac{\partial x_{\tau_{t, i}}^{\star}/\sqrt{1-\tau_{t, i}}}{\partial x_{\tau_{t, 0}}^{\star}/\sqrt{1-\tau_{t, 0}}} - \frac{\partial x_{\tau_{t, i}}^{(n)}/\sqrt{1-\tau_{t, i}}}{\partial x_{\tau_{t, 0}}/\sqrt{1-\tau_{t, 0}}}\bigg\|\\
	&\qquad+\sum_{0 \le i < K} (\tau_{t, 0}-\tau')(1-\tau_{t, i})^{-3/2}\sqrt{1-\tau_{t, i}}\sqrt{\frac{d^3\log^3 T}{\tau_{t,i}^3}}\big\|x_{\tau_{t, i}}^{(n)} - x_{\tau_{t, i}}^{\star}\big\|_2\\
	&\qquad+ \|\xi_{\mathsf{Jacob}, \tau'}^{(n)}\|
	+ O\Bigg(\bigg(\frac{d(\tau_{t, 0}-\tau')\log T}{\tau'(1-\tau')}\bigg)^{K+1}\Bigg) \\
	&\quad\lesssim 
	\frac{d\log^2 T}{T} \sum_{0 \le i < K} \bigg\{\bigg\|\frac{\partial x_{\tau_{t, i}}^{(n)}/\sqrt{1-\tau_{t, i}}}{\partial x_{\tau_{t, 0}}/\sqrt{1-\tau_{t, 0}}} - \frac{\partial x_{\tau_{t, i}}^{\star}/\sqrt{1-\tau_{t, i}}}{\partial x_{\tau_{t, 0}}^{\star}/\sqrt{1-\tau_{t, 0}}}\bigg\|
	+\sqrt{\frac{d\log T}{\tau_{t,i}}}\big\|x_{\tau_{t, i}}^{(n)} - x_{\tau_{t, i}}^{\star}\big\|_2\bigg\} \\
	&\qquad+ \|\xi_{\mathsf{Jacob}, \tau'}^{(n)}\|
	+ O\bigg(\Big(\frac{d\log^2 T}{T}\Big)^{K+1}\bigg),
\end{align*}
where the last inequality holds since $\frac{\tau_{t,0} - \tau_{t,i}}{\tau_{t,i}(1-\tau_{t,i})} \lesssim \frac{\log T}{T}$ and $\tau_{t,0} - \tau' \asymp \tau_{t,0} - \tau_{t,i}$ for all $1 \leq i < K$.
\begin{comment}
	\begin{align*}
		&\bigg\|\frac{\partial x_{\tau'}^{(n+1)}/\sqrt{1-\tau'}}{\partial x_{\tau_{t, 0}}/\sqrt{1-\tau_{t, 0}}} - \frac{\partial x_{\tau'}^{\star}/\sqrt{1-\tau'}}{\partial x_{\tau_{t, 0}}^{\star}/\sqrt{1-\tau_{t, 0}}}\bigg\|
		\lesssim \sum_{0 \le i < k} (\tau'-\tau_{t, 0})(1-\tau_{t, i})^{-3/2}\bigg\|\frac{\partial s_{\tau_{t, i}}^{\star}(x_{\tau_{t, i}}^{\star})}{\partial x_{\tau_{t, i}}^{\star}/\sqrt{1-\tau_{t, i}}}\frac{\partial x_{\tau_{t, i}}^{\star}/\sqrt{1-\tau_{t, i}}}{\partial x_{\tau_{t, 0}}^{\star}/\sqrt{1-\tau_{t, 0}}} \\
		&\qquad\qquad\qquad\qquad\qquad
		- \frac{\partial s_{\tau_{t, i}}^{\star}(x_{\tau_{t, i}}^{(n)})}{\partial x_{\tau_{t, i}}^{(n)}/\sqrt{1-\tau_{t, i}}}\frac{\partial x_{\tau_{t, i}}^{(n)}/\sqrt{1-\tau_{t, i}}}{\partial x_{\tau_{t, 0}}/\sqrt{1-\tau_{t, 0}}}\bigg\|  
		+ \|\xi_{\mathsf{Jacob}, \tau'}^{(n)}\|
		+ O\Bigg(\bigg(\frac{d(\tau'-\tau_{t, 0})\log T}{\tau(1-\tau)}\bigg)^{k+1}\Bigg) \\
		&\qquad\qquad\lesssim 
		\frac{d\log^2 T}{T} \sum_{0 \le i < k} \bigg\{\bigg\|\frac{\partial x_{\tau_{t, i}}^{(n)}/\sqrt{1-\tau_{t, i}}}{\partial x_{\tau_{t, 0}}/\sqrt{1-\tau_{t, 0}}} - \frac{\partial x_{\tau_{t, i}}^{\star}/\sqrt{1-\tau_{t, i}}}{\partial x_{\tau_{t, 0}}^{\star}/\sqrt{1-\tau_{t, 0}}}\bigg\|
		+\sqrt{\frac{d\log T}{\tau}}\big\|x_{\tau_{t, i}}^{(n)} - x_{\tau_{t, i}}^{\star}\big\|_2\bigg\} \\
		&\qquad\qquad\qquad\qquad\qquad+ \|\xi_{\mathsf{Jacob}, \tau'}^{(n)}\|
		+ O\Bigg(\bigg(\frac{d\log^2 T}{T}\bigg)^{k+1}\Bigg),
	\end{align*}
\end{comment}

The previous inequality combined with \eqref{ineq:score_estimation_2}, \eqref{ineq:induction_1} and \eqref{ineq13} yields that on the event $\mathcal{E}_t$,
\begin{align}
	&\bigg\|\frac{\partial x_{\tau'}^{(n+1)}/\sqrt{1-\tau'}}{\partial x_{\tau_{t, 0}}/\sqrt{1-\tau_{t, 0}}} - \frac{\partial x_{\tau'}^{\star}/\sqrt{1-\tau'}}{\partial x_{\tau_{t, 0}}^{\star}/\sqrt{1-\tau_{t, 0}}}\bigg\|^2\notag\\
	&\quad \leq C\frac{d^2\log^4 T}{T^2}\sum_{0 \le i < K}\bigg\|\frac{\partial x_{\tau_{t, i}}^{(n)}/\sqrt{1-\tau_{t, i}}}{\partial x_{\tau_{t, 0}}/\sqrt{1-\tau_{t, 0}}} - \frac{\partial x_{\tau_{t, i}}^{\star}/\sqrt{1-\tau_{t, i}}}{\partial x_{\tau_{t, 0}}^{\star}/\sqrt{1-\tau_{t, 0}}}\bigg\|^2\notag\\&\qquad + C\frac{d^3\log^5 T}{T^2\tau'}\bigg(\frac{1}{3^{n+1}}\max_{0 \leq i \leq K}\big\|x_{\tau_{t, i}}^{(0)} - x_{\tau_{t, i}}^{\star}\big\|_2^2 + C\tau'^{2}\frac{\log^2 T}{T^2}\sum_{0 \leq j \leq n+1}\sum_{0 \le i < K}\big(\varepsilon_{\mathsf{score}, t, i}^{(j)}(x_{\tau_{t, i}}^{(j)})\big)^2\notag\\&\hspace{2.8cm} + 2C\frac{d\tau'\log^3 T}{T^2}\Big(\frac{d\log^2 T}{T}\Big)^{2K}\bigg)\notag\\&\qquad + C\frac{\log^2 T}{T^2}\tau'^{2}\sum_{0 \le i < K}\big(\varepsilon_{\mathsf{Jacobi}, t, i}^{(n)}(x_{\tau_{t, i}}^{(n)})\big)^2 + C\Big(\frac{d\log^2 T}{T}\Big)^{2K+2}\notag\\
	&\leq C\frac{d^2\log^4 T}{T^2}\sum_{0 \le i < K}\bigg\|\frac{\partial x_{\tau_{t, i}}^{(n)}/\sqrt{1-\tau_{t, i}}}{\partial x_{\tau_{t, 0}}/\sqrt{1-\tau_{t, 0}}} - \frac{\partial x_{\tau_{t, i}}^{\star}/\sqrt{1-\tau_{t, i}}}{\partial x_{\tau_{t, 0}}^{\star}/\sqrt{1-\tau_{t, 0}}}\bigg\|^2\notag\\&\qquad + C\frac{d^3\log^5 T}{T^2\tau'}\bigg(\frac{1}{3^{n+1}}4c_1^2\frac{d\log^3 T}{T^2} + C\tau'^{2}\frac{\log^2 T}{T^2}\sum_{0 \leq j \leq n+1}\sum_{0 \le i < K}\big(\varepsilon_{\mathsf{score}, t, i}^{(j)}(x_{\tau_{t, i}}^{(j)})\big)^2\notag\\&\hspace{2.8cm} + 2C\frac{d\tau'\log^3 T}{T^2}\Big(\frac{d\log^2 T}{T}\Big)^{2K}\bigg)\notag\\&\qquad + C\frac{\log^2 T}{T^2}\tau'^{2}\sum_{0 \le i < K}\big(\varepsilon_{\mathsf{Jacobi}, t, i}^{(n)}(x_{\tau_{t, i}}^{(n)})\big)^2 + C\Big(\frac{d\log^2 T}{T}\Big)^{2K+2}.
\end{align}

Repeating similar arguments as in \eqref{ineq12} yields
\begin{align}\label{ineq15}
	&\bigg\|\frac{\partial x_{\tau'}^{(N)}/\sqrt{1-\tau'}}{\partial x_{\tau_{t, 0}}/\sqrt{1-\tau_{t, 0}}} - \frac{\partial x_{\tau'}^{\star}/\sqrt{1-\tau'}}{\partial x_{\tau_{t, 0}}^{\star}/\sqrt{1-\tau_{t, 0}}}\bigg\|^2\notag\\ &\lesssim \frac{d^3\log^7 T}{T^4}\sum_{i,n}\big(\varepsilon_{\mathsf{score}, t, i}^{(n)}(x_{\tau_{t, i}}^{(n)})\big)^2 + \frac{\log^2 T}{T^2}\tau'^{2}\sum_{i,n}\big(\varepsilon_{\mathsf{Jacobi}, t, i}^{(n)}(x_{\tau_{t, i}}^{(n)})\big)^2 + \Big(\frac{d\log^2 T}{T}\Big)^{2K+2}.
\end{align}
In addition, recognizing that \eqref{ineq3a} implies $$\bigg\|\frac{\partial x_{\tau_{t, K-1}}^{(N)}/\sqrt{1-\tau_{t, K-1}}}{\partial x_{\tau_{t, 0}}/\sqrt{1-\tau_{t, 0}}}\bigg\|^{-1} \lesssim 1,$$
we have finished the proof of Lemma \ref{lemma:main}.

\paragraph{Proofs of \eqref{ineq3a} and \eqref{ineq3b}.} Assuming that the event $\mathcal{E}_t$ occurs, we will prove \eqref{ineq3a} and \eqref{ineq3b} by induction. For the base case $n = 0$, since $x_{\tau'}^{(0)} = x_{\tau_{t, i}}^{(0)} = x_{\tau_{t, 0}}$, we have
\begin{align*}
	1 \lesssim \sqrt{\frac{1 - \overline{\alpha}_t}{1 - \overline{\alpha}_{t+1}}} \leq \left\|\frac{\partial x_{\tau_{t, i}}^{(0)}/\sqrt{1-\tau_{t, i}}}{\partial x_{\tau_{t, 0}}/\sqrt{1-\tau_{t, 0}}}\right\| = \left\|\frac{\sqrt{1 - \tau_{t, 0}}}{\sqrt{1 - \tau_{t, i}}}I\right\| = \frac{\sqrt{1 - \tau_{t, 0}}}{\sqrt{1 - \tau_{t, i}}} \leq 1.
\end{align*}
Moreover, it can be shown that on the event $\mathcal{E}_t$,
\begin{align}
	\|\xi_{\mathsf{Jacob}, \tau'}^{(0)}\|
	&= \bigg\|\sum_{0 \le i < K} \gamma_{t, i}(\tau')(1-\tau_{t, i})^{-3/2}\bigg(\frac{\partial s_{\tau_{t, i}}(x_{\tau_{t, i}}^{(0)})}{\partial x_{\tau_{t, i}}^{(0)}/\sqrt{1-\tau_{t, i}}} - \frac{\partial s_{\tau_{t, i}}^{\star}(x_{\tau_{t, i}}^{(0)})}{\partial x_{\tau_{t, i}}^{(0)}/\sqrt{1-\tau_{t, i}}}\bigg)\frac{\partial x_{\tau_{t, i}}^{(0)}/\sqrt{1-\tau_{t, i}}}{\partial x_{\tau_{t, 0}}/\sqrt{1-\tau_{t, 0}}}\bigg\|\notag\\
	&\lesssim \sum_{0 \le i < K} 2^K (\tau_{t, 0} - \tau')(1-\tau_{t, i})^{-1}\bigg\|\frac{\partial s_{\tau_{t, i}}(x_{\tau_{t, i}}^{(0)})}{\partial x_{\tau_{t, i}}^{(0)}} - \frac{\partial s_{\tau_{t, i}}^{\star}(x_{\tau_{t, i}}^{(0)})}{\partial x_{\tau_{t, i}}^{(0)}}\bigg\|\notag\\
	&\stackrel{\eqref{ineq:learning_rate_f}}{\ll} K\cdot 2^K\cdot \tau_{t, i}\frac{\log T}{T}\cdot \frac{T}{\log T}\notag\\
	&\lesssim 1,\label{ineq4}
\end{align}
Therefore, \eqref{ineq3a} and \eqref{ineq3b} hold for $n = 0$. Supposing that \eqref{ineq3a} and \eqref{ineq3b} hold for $n$, we would like to show that these two claims still hold for $n + 1$. In view of \eqref{ineq5}, one has
%and we make use of the following induction relation that on the event $\mathcal{E}$,
\begin{align*}
	\bigg\|\frac{\partial x_{\tau'}^{(n+1)}/\sqrt{1-\tau'}}{\partial x_{\tau_{t, 0}}/\sqrt{1-\tau_{t, 0}}}\bigg\|
	\le 1 + 2^K\sum_{0 \le i < K} (\tau_{t, 0} - \tau')\frac{d\log T}{\tau_{t, i}(1-\tau_{t, i})}\bigg\|\frac{\partial x_{\tau_{t, i}}^{(n)}/\sqrt{1-\tau_{t, i}}}{\partial x_{\tau_{t, 0}}/\sqrt{1-\tau_{t, 0}}}\bigg\| + \|\xi_{\mathsf{Jacob}, \tau'}^{(n)}\| \lesssim 1,
\end{align*}
provided that $-p_{\overline{X}_{\tau_{t, i}}}(x_{\tau_{t, i}}^{(n)}) \lesssim d\log T$. Here, the first inequality makes use of Lemma \ref{lemma:Hessian} and the second inequality holds due to the induction hypotheses \eqref{ineq3a} and \eqref{ineq3b} for $n$. Similarly, one has
\begin{align*}
	\bigg\|\frac{\partial x_{\tau'}^{(n+1)}/\sqrt{1-\tau'}}{\partial x_{\tau_{t, 0}}/\sqrt{1-\tau_{t, 0}}}\bigg\|
	\geq 1 - 2^K\sum_{0 \le i < K} (\tau_{t, 0} - \tau')\frac{d\log T}{\tau_{t, i}(1-\tau_{t, i})}\bigg\|\frac{\partial x_{\tau_{t, i}}^{(n)}/\sqrt{1-\tau_{t, i}}}{\partial x_{\tau_{t, 0}}/\sqrt{1-\tau_{t, 0}}}\bigg\| - \|\xi_{\mathsf{Jacob}, \tau'}^{(n)}\| \gtrsim 1
\end{align*}
In addition, repeating a similar argument as in \eqref{ineq4} yields
\begin{align*}
	\|\xi_{\mathsf{Jacob}, \tau'}^{(n+1)}\| \leq \frac{1}{2}.
\end{align*}
Therefore, \eqref{ineq3a} and \eqref{ineq3b} hold for $n+1$, and consequently we have finished the proof.
\paragraph{Proof of Claims~\eqref{eq:derivative_1} and \eqref{eq:derivative_2}.}
It remains to verify \eqref{eq:derivative_1} and \eqref{eq:derivative_2}. We let $$u_k := \frac{\partial^k}{\partial \tau^k} \frac{s_{\tau}^{\star}(x_{\tau}^{\star})}{(1-\tau)^{3/2}}.$$
The score function can be expressed as
\begin{align}\label{eq2}
	\frac{s_{\tau}^{\star}(\sqrt{1-\tau}x)}{(1-\tau)^{3/2}}
	&= -\frac{1}{\tau(1-\tau)}\int_{x_0} p_{X_0 \mymid \overline{X}_\tau}(x_0 \mymid \sqrt{1-\tau}x)(x - x_0)\mathrm{d}x_0 \notag\\
	&= -\frac{1}{\tau(1-\tau)}\cdot\frac{\int_{x_0} p_{X_0}(x_0)p_{\overline{X}_\tau \mymid X_0}(\sqrt{1-\tau}x \mymid x_0)(x - x_0)\mathrm{d}x_0}{\int_{x_0} p_{X_0}(x_0)p_{\overline{X}_\tau \mymid X_0}(\sqrt{1-\tau}x \mymid x_0)\mathrm{d}x_0},
\end{align}
where
\begin{align*}
	(2\pi\tau)^{d/2}p_{\overline{X}_\tau \mymid X_0}(\sqrt{1-\tau}x \mymid x_0)
	= \exp\Big(-\frac{(1-\tau)\|x - x_0\|_2^2}{2\tau}\Big).
\end{align*}
Then for sufficiently small $\delta$, we have
\begin{align}
	&\frac{s_{\tau+\delta}^{\star}(x_{\tau+\delta}^{\star})}{(1-\tau-\delta)^{3/2}}\notag\\
	&\quad= -\frac{1}{(\tau+\delta)(1-\tau-\delta)}\int_{x_0} p_{X_0 \mymid \overline{X}_{\tau+\delta}}(x_0 \mymid x_{\tau+\delta}^{\star})\bigg(\frac{x_{\tau+\delta}^{\star}}{\sqrt{1-\tau-\delta}} - x_0\bigg)\mathrm{d}x_0 \notag\\
	&\quad= -\frac{1}{(\tau+\delta)(1-\tau-\delta)}\bigg(\frac{x_{\tau+\delta}^{\star}}{\sqrt{1-\tau-\delta}} - \frac{x_{\tau}^{\star}}{\sqrt{1-\tau}} + \int_{x_0} p_{X_0 \mymid \overline{X}_{\tau+\delta}}(x_0 \mymid x_{\tau+\delta}^{\star})\bigg(\frac{x_{\tau}^{\star}}{\sqrt{1-\tau}} - x_0\bigg)\mathrm{d}x_0\bigg)\notag\\
	&\quad= -\frac{1}{(\tau+\delta)(1-\tau-\delta)}\left(\frac{x_{\tau+\delta}^{\star}}{\sqrt{1-\tau-\delta}} - \frac{x_{\tau}^{\star}}{\sqrt{1-\tau}} + \frac{\int_{x_0} p_{X_0}(x_0)p_{\overline{X}_{\tau+\delta} \mymid X_0}(x_{\tau+\delta}^{\star} \mymid x_0)\left(\frac{x_{\tau}^{\star}}{\sqrt{1-\tau}} - x_0\right)\mathrm{d}x_0}{\int_{x_0} p_{X_0}(x_0)p_{\overline{X}_{\tau+\delta} \mymid X_0}(x_{\tau+\delta}^{\star} \mymid x_0)\mathrm{d}x_0}\right)\notag\\
	&\quad= -\frac{1}{(\tau+\delta)(1-\tau-\delta)}\left(\frac{x_{\tau+\delta}^{\star}}{\sqrt{1-\tau-\delta}} - \frac{x_{\tau}^{\star}}{\sqrt{1-\tau}} + \frac{\int_{x_0} p_{X_0}(x_0)p_{\overline{X}_{\tau} \mymid X_0}(x_{\tau}^{\star} \mymid x_0)\exp(\Delta)\left(\frac{x_{\tau}^{\star}}{\sqrt{1-\tau}} - x_0\right)\mathrm{d}x_0}{\int_{x_0} p_{X_0}(x_0)p_{\overline{X}_{\tau} \mymid X_0}(x_{\tau}^{\star} \mymid x_0)\exp(\Delta)\mathrm{d}x_0}\right)\notag\\
	&\quad= -\frac{1}{(\tau+\delta)(1-\tau-\delta)}\left(\frac{x_{\tau+\delta}^{\star}}{\sqrt{1-\tau-\delta}} - \frac{x_{\tau}^{\star}}{\sqrt{1-\tau}} + \frac{\int_{x_0} p_{X_0 \mymid \overline{X}_\tau}(x_0 \mymid x_{\tau}^{\star})\exp(\Delta)\big(\frac{x_{\tau}^{\star}}{\sqrt{1-\tau}} - x_0\big)\mathrm{d}x_0}
	{\int_{x_0} p_{X_0 \mymid \overline{X}_\tau}(x_0 \mymid x_{\tau}^{\star})\exp(\Delta)\mathrm{d}x_0}\right),\label{eq1}
\end{align}
%\begin{align*}
%\frac{s_{\tau+\delta}^{\star}(x_{\tau+\delta}^{\star})}{(1-\tau-\delta)^{3/2}}
%&= \frac{x_{\tau}^{\star}}{\sqrt{1-\tau}} - \frac{x_{\tau+\delta}^{\star}}{\sqrt{1-\tau-\delta}} 
%-\frac{1}{(\tau+\delta)(1-\tau-\delta)}
%\frac{\int_{x_0} p_{X_0 \mymid X_\tau}(x_0 \mymid x_{\tau}^{\star})\exp(\Delta)\big(\frac{x_{\tau}^{\star}}{\sqrt{1-\tau}} - x_0\big)\mathrm{d}x_0}
%{\int_{x_0} p_{X_0 \mymid X_\tau}(x_0 \mymid x_{\tau}^{\star})\exp(\Delta)\mathrm{d}x_0},
%\end{align*}
where
\begin{align*}
	\Delta := \frac{(1-\tau)\|x_{\tau}^{\star}/\sqrt{1-\tau} - x_0\|_2^2}{\tau} - \frac{(1-\tau-\delta)\|x_{\tau+\delta}^{\star}/\sqrt{1-\tau-\delta} - x_0\|_2^2}{\tau+\delta}
	=: \sum_{k = 1}^{\infty} \frac{\delta^k}{k!}v_k.
\end{align*}
Here, the last equation of \eqref{eq1} makes use of the Bayes rule.
We prove by induction that for any quantity $\overline{C} \geq 2$,
\begin{subequations}
	\begin{align}
		\|u_k\|_2 &\le c_k \sqrt{\frac{d\theta\log T}{\tau(1-\tau)^3}}\bigg(\frac{d\theta\log T}{\tau(1-\tau)}\bigg)^k, \quad\text{for }k = 0, 1, \ldots, K,\label{ineq9a}\\
		|v_k| &\le C_k\overline{C}^2\bigg(\frac{d\theta\log T}{\tau(1-\tau)}\bigg)^k,
		\quad\text{for }k = 1, \ldots, K, \quad\text{provided that }\Big\|\frac{x_{\tau}^{\star}}{\sqrt{1-\tau}} - x_0\Big\|_2 \leq 5\overline{C}\sqrt{\frac{d\theta\tau\log T}{1-\tau}}.\label{ineq9b}
	\end{align}
\end{subequations}
For $k = 0$, we have 
\begin{align*}
	\|u_0\|_2 &= \left\|\frac{s_{\tau}^{\star}(x_{\tau}^\star)}{(1-\tau)^{3/2}}\right\|_2\\ &= \frac{1}{\tau(1 - \tau)}\left\|\int_{x_0} p_{X_0 \mymid \overline{X}_\tau}(x_0 \mymid x_{\tau}^{\star})\left(\frac{x_{\tau}^{\star}}{\sqrt{1-\tau}} - x_0\right)\mathrm{d}x_0\right\|_2\\
	&= \frac{1}{\tau(1 - \tau)}\int_{x_0} p_{X_0 \mymid \overline{X}_\tau}(x_0 \mymid x_{\tau}^{\star})\left\|\frac{x_{\tau}^{\star}}{\sqrt{1-\tau}} - x_0\right\|\mathrm{d}x_0\\
	&= \frac{1}{\tau(1 - \tau)}\mathbb{E}\left[\left\|\frac{x_{\tau}^{\star}}{\sqrt{1-\tau}} - X_0\right\|_2\ \bigg|\ \overline{X}_{\tau} = x_{\tau}^{\star}\right]\\
	&\stackrel{\text{Lemma \ref{lemma:tail_bound}}}{\lesssim} \frac{1}{\tau(1 - \tau)}\frac{1}{\sqrt{1-\tau}}\sqrt{d\theta\tau\log T}\\
	&\asymp \sqrt{\frac{d\theta\log T}{\tau(1-\tau)^3}}.
\end{align*}
Suppose that \eqref{ineq9a} and \eqref{ineq9b} hold for $k \leq k_0$. We would like to prove \eqref{ineq9a} and \eqref{ineq9b} for $k = k_0 + 1$. Recalling that
\begin{align*}
	\frac{\partial}{\partial \tau} \frac{x_{\tau}^{\star}}{\sqrt{1-\tau}} = -\frac{1}{2(1-\tau)^{3/2}}s_{\tau}^{\star}(x_{\tau}^{\star}),
\end{align*}
we have
\begin{align*}
	\frac{\partial^k}{\partial \tau^k} \frac{x_{\tau}^{\star}}{\sqrt{1-\tau}} = -\frac{\partial^{k-1}}{\partial \tau^{k-1}}\frac{1}{2(1-\tau)^{3/2}}s_{\tau}^{\star}(x_{\tau}^{\star}).
\end{align*}
Then we know from Taylor expansion that 
\begin{align*}
	\frac{x_{\tau+\delta}^{\star}}{\sqrt{1-\tau-\delta}}-\frac{x_{\tau}^{\star}}{\sqrt{1-\tau}}
	= -\frac{1}{2}\sum_{k = 1}^{\infty} \frac{\delta^k}{k!}u_{k-1}.
\end{align*}
Furthermore, it is straightforward to verify that
\begin{align}\label{eq3}
	\frac{1}{(\tau+\delta)(1-\tau-\delta)}-\frac{1}{\tau(1-\tau)}
	= \sum_{k = 1}^{\infty} \delta^k\big[(1-\tau)^{-k-1} - (-\tau)^{-k-1}\big],
\end{align}
and
\begin{align}\label{eq4}
	&\frac{(1-\tau-\delta)\|x_{\tau+\delta}^{\star}/\sqrt{1-\tau-\delta} - x_0\|_2^2}{\tau+\delta}\notag\\
	&\quad= \Big(\frac{1-\tau}{\tau} - \sum_{k = 1}^{\infty} \delta^k(-\tau)^{-k-1}\Big)\Big\|\frac{x_{\tau}^{\star}}{\sqrt{1-\tau}} - x_0 + \frac{x_{\tau+\delta}^{\star}}{\sqrt{1-\tau-\delta}}-\frac{x_{\tau}^{\star}}{\sqrt{1-\tau}}\Big\|_2^2\notag\\
	&\quad= \Big(\frac{1-\tau}{\tau} - \sum_{k = 1}^{\infty} \delta^k(-\tau)^{-k-1}\Big)\bigg(\Big\|\frac{x_{\tau}^{\star}}{\sqrt{1-\tau}} - x_0\Big\|_2^2 - \Big(\frac{x_{\tau}^{\star}}{\sqrt{1-\tau}} - x_0\Big)^{\top}\sum_{k = 1}^{\infty} \frac{\delta^k}{k!}u_{k-1} + \frac{1}{4}\Big\|\sum_{k = 1}^{\infty} \frac{\delta^k}{k!}u_{k-1}\Big\|_2^2\bigg).
\end{align}
Then we immediately have
\begin{align*}
	\sum_{k = 1}^{\infty} \frac{\delta^k}{k!}v_k &= \sum_{k=1}^{\infty}\bigg[-(-\tau)^{-k-1}\Big\|\frac{x_{\tau}^{\star}}{\sqrt{1-\tau}} - x_0\Big\|_2^2 - \frac{1}{k!}\frac{1 - \tau}{\tau}\Big(\frac{x_{\tau}^{\star}}{\sqrt{1-\tau}} - x_0\Big)^{\top}u_{k-1}\\ &\hspace{1cm}+ \Big(\frac{x_{\tau}^{\star}}{\sqrt{1-\tau}} - x_0\Big)^{\top}\sum_{\ell = 1}^{k-1}(-\tau)^{-\ell-1}\frac{1}{(k-\ell)!}u_{k - \ell - 1} + \frac{1}{4}\frac{1-\tau}{\tau}\sum_{\ell = 1}^{k - 1}\frac{1}{\ell!(k - \ell)!}u_{\ell - 1}u_{k - \ell - 1}\\&\hspace{1cm} - \sum_{\ell = 1}^{k - 2}\sum_{j = 1}^{k - 1 - \ell}\frac{1}{j!(k-\ell-j)!}(-\tau)^{-\ell - 1}u_{j - 1}u_{k - \ell - j - 1}\bigg]\delta^k 
\end{align*}
for sufficiently small $\delta > 0$.
Comparing the coefficients of $\delta^{k_0 + 1}$ on both sides and making use of the induction assumption, one has 
\begin{align*}
	|v_{k_0 + 1}| &\leq \tau^{-k_0}\Big\|\frac{x_{\tau}^{\star}}{\sqrt{1-\tau}} - x_0\Big\|_2^2 + \frac{1}{(k_0 + 1)!}\frac{1 - \tau}{\tau}\Big\|\frac{x_{\tau}^{\star}}{\sqrt{1-\tau}} - x_0\Big\|_2\|u_{k_0}\|_2\\
	&\quad + \Big\|\frac{x_{\tau}^{\star}}{\sqrt{1-\tau}} - x_0\Big\|_2\sum_{\ell = 1}^{k_0}\tau^{-\ell-1}\frac{1}{(k_0+1-\ell)!}u_{k_0 - \ell}\\
	&\quad + \frac{1}{4}\sum_{\ell = 1}^{k_0}\frac{1}{\ell!(k_0+1 - \ell)!}u_{\ell-1}u_{k_0 - \ell} + \sum_{\ell = 1}^{k_0 - 1}\sum_{j = 1}^{k_0 - \ell}\frac{1}{j!(k_0+1-\ell-j)!}\tau^{-\ell - 1}u_{j - 1}u_{k_0 - \ell - j}\\
	&\lesssim \tau^{k_0}\overline{C}^2\frac{d\theta\tau\log T}{1 - \tau} + \frac{1 - \tau}{\tau}\overline{C}\sqrt{\frac{d\theta\tau\log T}{1-\tau}}\sqrt{\frac{d\theta\log T}{\tau(1-\tau)^3}}\bigg(\frac{d\theta\log T}{\tau(1-\tau)}\bigg)^{k_0}\\&\quad + \overline{C}\sqrt{\frac{d\theta\tau\log T}{1-\tau}}\sum_{\ell = 1}^{k_0}\tau^{-\ell-1}\sqrt{\frac{d\theta\log T}{\tau(1-\tau)^3}}\bigg(\frac{d\theta\log T}{\tau(1-\tau)}\bigg)^{k_0 - \ell}\\&\quad + \frac{1-\tau}{\tau}\sum_{\ell = 1}^{k_0}\sqrt{\frac{d\theta\log T}{\tau(1-\tau)^3}}\bigg(\frac{d\theta\log T}{\tau(1-\tau)}\bigg)^{\ell - 1}\sqrt{\frac{d\theta\log T}{\tau(1-\tau)^3}}\bigg(\frac{d\theta\log T}{\tau(1-\tau)}\bigg)^{k_0 - \ell}\\
	&\quad + \sum_{\ell = 1}^{k_0 - 1}\sum_{j = 1}^{k_0 - \ell}\tau^{-\ell - 1}\sqrt{\frac{d\theta\log T}{\tau(1-\tau)^3}}\bigg(\frac{d\theta\log T}{\tau(1-\tau)}\bigg)^{j - 1}\sqrt{\frac{d\theta\log T}{\tau(1-\tau)^3}}\bigg(\frac{d\theta\log T}{\tau(1-\tau)}\bigg)^{k_0-\ell-j}\\
	&\lesssim \overline{C}^2\bigg(\frac{d\theta\log T}{\tau(1-\tau)}\bigg)^{k_0+1},
\end{align*}
provided that $\|\frac{x_{\tau}^{\star}}{\sqrt{1-\tau}} - x_0\|_2 \leq 5\overline{C}\sqrt{\frac{d\theta\tau\log T}{1-\tau}}$. Therefore, we have verified \eqref{ineq9b} for $k = k_0 + 1$.

Then, we prove that \eqref{ineq9a} holds for $k = k_0 + 1$. We let $e^\Delta =: \sum_{k = 0}^{\infty}\frac{\delta^k}{k!}w_k = \sum_{k = 0}^{\infty}\frac{\delta^k}{k!}w_k(x_0)$ denote the Taylor expansion of $e^\Delta$. Then one can show that $w_0 = 1$ and for all $1 \leq k \leq k_0+1$,
\begin{align}\label{ineq10}
	|w_k| = \bigg|\sum_{\ell = 1}^{k}\sum_{j_1 + \cdots + j_{\ell} = k}\frac{k!}{j_1!\cdots j_\ell!}v_{j_1}\cdots v_{j_\ell}\bigg| \lesssim \overline{C}^{2k}\bigg(\frac{d\theta\log T}{\tau(1-\tau)}\bigg)^{k},
\end{align}
provided that $\|\frac{x_{\tau}^{\star}}{\sqrt{1-\tau}} - x_0\|_2 \leq 5\overline{C}\sqrt{\frac{d\theta\tau\log T}{1-\tau}}$.

We define
\begin{align*}
	\mathcal{C}_0 &= \left\{x_0 \in \mathbb{R}^d: \Big\|\frac{x_{\tau}^{\star}}{\sqrt{1-\tau}} - x_0\Big\|_2 \leq 5c_7\sqrt{\frac{d\theta\tau\log T}{1-\tau}}\right\},\\
	\mathcal{C}_\ell &= \left\{x_0 \in \mathbb{R}^d: 5\cdot 2^{\ell-1}c_7\sqrt{\frac{d\tau\log T}{1-\tau}} < \Big\|\frac{x_{\tau}^{\star}}{\sqrt{1-\tau}} - x_0\Big\|_2 \leq 5\cdot 2^{\ell}c_7\sqrt{\frac{d\theta\tau\log T}{1-\tau}}\right\},\quad k \geq 1.
\end{align*}
where $c_7$ is a sufficiently large constant.
Let $a_k$ (resp.~$b_k$) denote the $k$-th order derivative\\$\frac{\partial^k}{\partial \tau^k}\int_{x_0} p_{X_0 \mymid \overline{X}_\tau}(x_0 \mymid x_{\tau}^{\star})\exp(\Delta)\big(\frac{x_{\tau}^{\star}}{\sqrt{1-\tau}} - x_0\big)\mathrm{d}x_0$ (resp.~$\frac{\partial^k}{\partial \tau^k}\int_{x_0} p_{X_0 \mymid \overline{X}_\tau}(x_0 \mymid x_{\tau}^{\star})\exp(\Delta)\mathrm{d}x_0$). Then for $0 \leq k \leq k_0 + 1$,
\begin{align*}
	\|a_k\|_2&= \Big\|\int_{x_0} p_{X_0 \mymid \overline{X}_\tau}(x_0 \mymid x_{\tau}^{\star})\Big(\frac{x_{\tau}^{\star}}{\sqrt{1-\tau}} - x_0\Big)w_k\mathrm{d}x_0\Big\|_2\\
	&= \Big\|\sum_{\ell=0}^{\infty}\int_{x_0 \in \mathcal{C}_\ell} p_{X_0 \mymid \overline{X}_\tau}(x_0 \mymid x_{\tau}^{\star})\Big(\frac{x_{\tau}^{\star}}{\sqrt{1-\tau}} - x_0\Big)w_k\mathrm{d}x_0\Big\|_2\\
	&\leq \Big\|\int_{x_0 \in \mathcal{C}_0} p_{X_0 \mymid \overline{X}_\tau}(x_0 \mymid x_{\tau}^{\star})\Big(\frac{x_{\tau}^{\star}}{\sqrt{1-\tau}} - x_0\Big)w_k\mathrm{d}x_0\Big\|_2\\
	&\quad + \sum_{\ell=1}^{\infty}\Big\|\int_{x_0 \in \mathcal{C}_\ell} p_{X_0 \mymid \overline{X}_\tau}(x_0 \mymid x_{\tau}^{\star})\Big(\frac{x_{\tau}^{\star}}{\sqrt{1-\tau}} - x_0\Big)w_k\mathrm{d}x_0\Big\|_2\\
	&\leq \sup_{x_0 \in \mathcal{C}_0}\Big\|\frac{x_{\tau}^{\star}}{\sqrt{1-\tau}} - x_0\Big\|_2\|w_k\|_2\\&\quad + \sum_{\ell=1}^{\infty}\mathbb{P}\left(\Big\|\frac{x_{\tau}^{\star}}{\sqrt{1-\tau}} - X_0\Big\|_2 \leq 5\cdot 2^\ell c_7\sqrt{\frac{d\theta\tau\log T}{1-\tau}}\ \Big|\ \overline{X}_{\tau} = x_{\tau}^\star\right)\sup_{x_0 \in \mathcal{C}_\ell}\Big\|\frac{x_{\tau}^{\star}}{\sqrt{1-\tau}} - x_0\Big\|_2\|w_k\|_2\\
	&\stackrel{(*)}{\lesssim} 5c_7\sqrt{\frac{d\theta\tau\log T}{1-\tau}}\cdot c_7^{2k}\bigg(\frac{d\theta\log T}{\tau(1-\tau)}\bigg)^{k}\\&\quad + \sum_{\ell=1}^{\infty}\exp\left(-2^{2\ell}c_7^2d\theta\log T\right)\cdot 2^{\ell}c_7\sqrt{\frac{d\theta\tau\log T}{1-\tau}}\cdot (5\cdot2^\ell c_7)^{2k}\bigg(\frac{d\theta\log T}{\tau(1-\tau)}\bigg)^{k}\\
	&\lesssim \sqrt{\frac{d\theta\tau\log T}{1-\tau}}\bigg(\frac{d\theta\log T}{\tau(1-\tau)}\bigg)^{k} + \sum_{\ell = 1}^\infty\exp\left(-2^{\ell}c_7^2d\theta\log T\right)\sqrt{\frac{d\theta\tau\log T}{1-\tau}}\bigg(\frac{d\theta\log T}{\tau(1-\tau)}\bigg)^{k}\\
	&\asymp \sqrt{\frac{d\theta\tau\log T}{1-\tau}}\bigg(\frac{d\theta\log T}{\tau(1-\tau)}\bigg)^{k},
\end{align*}
and more specifically,
\begin{align*}
	a_0 = \int_{x_0} p_{X_0 \mymid \overline{X}_\tau}(x_0 \mymid x_{\tau}^{\star})\Big(\frac{x_{\tau}^{\star}}{\sqrt{1-\tau}} - x_0\Big)w_0\mathrm{d}x_0 = \int_{x_0} p_{X_0 \mymid \overline{X}_\tau}(x_0 \mymid x_{\tau}^{\star})\Big(\frac{x_{\tau}^{\star}}{\sqrt{1-\tau}} - x_0\Big)\mathrm{d}x_0.
\end{align*}
Here, $(*)$ is valid due to Lemma \ref{lemma:tail_bound} and \eqref{ineq10}.
Similarly, for all $1 \leq k \leq k_0+1$, one can show that
\begin{align*}
	b_0 = 1,~\quad~\text{and}~\quad~|b_k| \lesssim \bigg(\frac{d\theta\log T}{\tau(1-\tau)}\bigg)^{k},\quad \forall 1 \leq k \leq k_0 + 1.
\end{align*}
We denote by $$d_k:= \frac{\partial^k}{\partial \tau^k}\frac{\int_{x_0} p_{X_0 \mymid \overline{X}_\tau}(x_0 \mymid x_{\tau}^{\star})\exp(\Delta)\big(\frac{x_{\tau}^{\star}}{\sqrt{1-\tau}} - x_0\big)\mathrm{d}x_0}
{\int_{x_0} p_{X_0 \mymid \overline{X}_\tau}(x_0 \mymid x_{\tau}^{\star})\exp(\Delta)\mathrm{d}x_0}$$ the $k$-th derivative of $\frac{\int_{x_0} p_{X_0 \mymid \overline{X}_\tau}(x_0 \mymid x_{\tau}^{\star})\exp(\Delta)\big(\frac{x_{\tau}^{\star}}{\sqrt{1-\tau}} - x_0\big)\mathrm{d}x_0}
{\int_{x_0} p_{X_0 \mymid \overline{X}_\tau}(x_0 \mymid x_{\tau}^{\star})\exp(\Delta)\mathrm{d}x_0}$. Then one can easily verify that
\begin{subequations}
	\begin{align}
		d_0 &= \int_{x_0} p_{X_0 \mymid \overline{X}_\tau}(x_0 \mymid x_{\tau}^{\star})\Big(\frac{x_{\tau}^{\star}}{\sqrt{1-\tau}} - x_0\Big)\mathrm{d}x_0,\label{ineq11a}\\
		|d_k| &= |a_k - \sum_{\ell = 0}^{k-1}d_\ell b_{k-\ell}| \lesssim \sqrt{\frac{d\theta\tau\log T}{1-\tau}}\bigg(\frac{d\theta\log T}{\tau(1-\tau)}\bigg)^{k},\quad\text{for all}~0 \leq k \leq k_0+1.\label{ineq11b}
	\end{align}
\end{subequations}
Now, we are ready to prove \eqref{ineq9b} for $t = t_0 + 1$. In view of \eqref{eq1} and \eqref{eq2}, one has
\begin{align}
	\sum_{k = 1}^{\infty}\frac{\delta^k}{k!}u_k &=\frac{s_{\tau+\delta}^{\star}(x_{\tau+\delta}^{\star})}{(1-\tau-\delta)^{3/2}} - \frac{s_{\tau}^{\star}(x_{\tau}^{\star})}{(1-\tau)^{3/2}}\notag\\
	&= -\frac{1}{(\tau+\delta)(1-\tau-\delta)}\left(\frac{x_{\tau+\delta}^{\star}}{\sqrt{1-\tau-\delta}} - \frac{x_{\tau}^{\star}}{\sqrt{1-\tau}} + \frac{\int_{x_0} p_{X_0 \mymid \overline{X}_\tau}(x_0 \mymid x_{\tau}^{\star})\exp(\Delta)\big(\frac{x_{\tau}^{\star}}{\sqrt{1-\tau}} - x_0\big)\mathrm{d}x_0}
	{\int_{x_0} p_{X_0 \mymid \overline{X}_\tau}(x_0 \mymid x_{\tau}^{\star})\exp(\Delta)\mathrm{d}x_0}\right)\notag\\&\quad + \frac{1}{\tau(1-\tau)}\int_{x_0} p_{X_0 \mymid \overline{X}_\tau}(x_0 \mymid \sqrt{1-\tau}x)(x - x_0)\mathrm{d}x_0\notag\\
	&=-\left(\frac{1}{\tau(1-\tau)} + \sum_{k = 1}^{\infty} \delta^k\big[(1-\tau)^{-k-1} - (-\tau)^{-k-1}\big]\right)\left(-\frac{1}{2}\sum_{k = 1}^{\infty} \frac{\delta^k}{k!}u_{k-1} + \sum_{k = 0}^{\infty} \frac{\delta^k}{k!}d_{k}\right) + \frac{1}{\tau(1-\tau)}d_0\notag\\
	&= \sum_{k=1}^{\infty}\left(\frac{1}{\tau(1-\tau)}\frac{u_{k-1}-2d_k}{2k!} + \sum_{\ell=1}^{k}\big[(-\tau)^{-\ell-1}-(1-\tau)^{-\ell-1}\big]\frac{2d_{k-\ell}-u_{k-1-\ell}}{2(k-\ell)!}\right)\delta^k.
\end{align}
By virtue of the induction hypothesis \eqref{ineq9a} for $1 \leq k \leq k_0$ and \eqref{ineq11b}, we have
\begin{align*}
	|u_{k_0+1}| &= \left|\frac{1}{\tau(1-\tau)}\left(\frac{u_{k_0}}{2} - d_{k_0+1}\right) + \sum_{\ell=1}^{k_0+1}\big[(-\tau)^{-\ell-1}-(1-\tau)^{-\ell-1}\big]\frac{(k_0+1)!}{2(k_0+1-\ell)!}\left(2d_{k_0+1-\ell}-u_{k_0-\ell}\right)\right|\\
	&\lesssim \frac{1}{\tau(1-\tau)}\left(\sqrt{\frac{d\theta\log T}{\tau(1-\tau)^3}}\bigg(\frac{d\theta\log T}{\tau(1-\tau)}\bigg)^{k_0} + \sqrt{\frac{d\theta\tau\log T}{1-\tau}}\bigg(\frac{d\theta\log T}{\tau(1-\tau)}\bigg)^{k_0+1}\right)\\&\quad + \sum_{\ell = 1}^{k_0+1}\frac{1}{[\tau(1-\tau)]^{\ell+1}}\left[\sqrt{\frac{d\theta\tau\log T}{1-\tau}}\bigg(\frac{d\theta\log T}{\tau(1-\tau)}\bigg)^{k_0+1-\ell}+\sqrt{\frac{d\theta\log T}{\tau(1-\tau)^3}}\bigg(\frac{d\theta\log T}{\tau(1-\tau)}\bigg)^{k_0-\ell}\right]\\
	&\asymp \sqrt{\frac{d\theta\log T}{\tau(1-\tau)^3}}\bigg(\frac{d\theta\log T}{\tau(1-\tau)}\bigg)^{k_0+1},
\end{align*}
which has finished the proof of the induction step. Therefore, we have verified Claim \eqref{eq:derivative_1}. 

Claim \eqref{eq:derivative_2} can be proved by using similar arguments. We omit the details here for the sake of brevity.

%In addition, the remaining relation can be proved in a similar way, which is omitted here.

\section{Proofs of technical lemmas}
\subsection{Proof of Lemma \ref{lemma:learning_rate}}\label{sec:proof_lemma_learning_rate}
Equations \eqref{ineq:learning_rate_a} - \eqref{ineq:learning_rate_e} follow directly from \citet[(26a) - (26e)]{li2024sharp}.
\paragraph{Proof of \eqref{ineq:learning_rate_f}.} For any $2 \leq t \leq T, 0 \leq i_1, i_2, i_3, i_4 \leq K - 1$, one has
\begin{align*}
	\Big|\frac{\tau_{t, i_1} - \tau_{t, i_2}}{\tau_{t, i_3}(1-\tau_{t, i_4})}\Big| \leq \frac{\tau_{t,0} - \tau_{t,k-1}}{\tau_{t,k-1}\left(1 - \tau_{t,0}\right)} = \frac{\overline{\alpha}_{t-1} - \overline{\alpha}_{t}}{\overline{\alpha}_{t}\left(1 - \overline{\alpha}_{t-1}\right)} = \frac{1}{\alpha_t}\frac{1 - \alpha_t}{1 - \overline{\alpha}_{t-1}} \stackrel{\eqref{ineq:learning_rate_a}~\text{and}~\eqref{ineq:learning_rate_b}}{\leq} 2\cdot\frac{4c_1\log T}{T} = \frac{8c_1\log T}{T}.
\end{align*}
\paragraph{Proof of \eqref{ineq:learning_rate_g}.} By virtue of \eqref{def:gamma}, one has 
\begin{align*}
	\left|\gamma_{t, i}(\tau_{t,j})\right| \leq \left(\tau_{t,0} - \tau_{t,j}\right)\sup_{\tau \in [\tau_{t,K-1}, \tau_{t,0}]}\left|\psi_i(\tau)\right|.
\end{align*}
We observe that 
\begin{align*}
	\sup_{\tau \in [\tau_{t,K-1}, \tau_{t,0}]}\left|\psi_i(\tau)\right| = \sup_{\eta \in [0,K-1]}\left|\frac{\prod_{i' \neq i}(K-1-i'-\eta)}{\prod_{i' \neq i}(i-i')}\right| \leq \frac{(K-1)!}{\left(\lfloor\frac{K-1}{2}\rfloor\right)!\left(\lfloor\frac{K-1}{2}\rfloor\right)!} = \binom{K-1}{\lfloor\frac{K-1}{2}\rfloor} \leq 2^{K}.
\end{align*}
Putting the previous two inequalities together, we have completed the proof of \eqref{ineq:learning_rate_g}.
\paragraph{Proof of \eqref{ineq:learning_rate_h}.} For all $0 \leq i \leq K-1, 2 \leq t \leq T$, one has
\begin{align*}
	\overline{\alpha}_{t} = 1 - \tau_{t,0} \leq 1 - \tau_{t,i} \leq 1 - \tau_{t,K-1} = \overline{\alpha}_{t-1}~\quad~\text{and}~\quad~1 - \overline{\alpha}_{t-1} = \tau_{t,K-1} \leq \tau_{t,i} \leq \tau_{t,0} = 1 - \overline{\alpha}_{t}.
\end{align*}
Therefore, it suffices to show that 
\begin{align*}
	\overline{\alpha}_{t} \asymp \overline{\alpha}_{t-1}~\quad~\text{and}~\quad~1 - \overline{\alpha}_{t-1} \asymp 1 - \overline{\alpha}_{t},
\end{align*}
which are direct consequences of \eqref{ineq:learning_rate_a} and \eqref{ineq:learning_rate_c}, respectively.

\subsection{Proof of Lemma \ref{lemma:Hessian}}\label{sec:proof_lemma_Hessian}
Elementary calculations reveal that
\begin{align*}
	\left\|s_{\tau}^{\star}(x)\right\|_2 &= \frac{\sqrt{1 - \tau}}{\tau}\left\|\int_{x_0} p_{X_0 \mymid \overline{X}_\tau}(x_0 \mymid x)\left(\frac{x}{\sqrt{1-\tau}} - x_0\right)\mathrm{d}x_0\right\|_2\\
	&= \frac{\sqrt{1 - \tau}}{\tau}\int_{x_0} p_{X_0 \mymid \overline{X}_\tau}(x_0 \mymid x)\left\|\frac{x}{\sqrt{1-\tau}} - x_0\right\|\mathrm{d}x_0\\
	&= \frac{\sqrt{1 - \tau}}{\tau}\mathbb{E}\left[\left\|\frac{x}{\sqrt{1-\tau}} - X_0\right\|_2\ \bigg|\ \overline{X}_{\tau} = x\right]\\
	&\stackrel{\text{Lemma \ref{lemma:tail_bound}}}{\lesssim} \frac{\sqrt{1 - \tau}}{\tau}\frac{1}{\sqrt{1-\tau}}\sqrt{d\theta_{\tau}(x)\tau\log T}\\
	&\asymp \sqrt{\frac{d\theta_{\tau}(x)\log T}{\tau}},
\end{align*}
which has finished the proof of \eqref{score_1}.

Turning to \eqref{score_2}, by virtue of (24) and (25b) in \cite{li2024sharp}, we have
\begin{align}\label{ineq8}
	\frac{\partial s_{\tau}^\star(x_{\tau})}{\partial x} = -\frac{1}{\tau}I_d + \frac{1}{\tau^2}{\sf Cov}\left(\overline{X}_{\tau} - \sqrt{1 - \tau}X_0\ |\ \overline{X}_{\tau} = x\right).
\end{align}
We make the observation that
\begin{align*}
	\left\|{\sf Cov}\left(\overline{X}_{\tau} - \sqrt{1 - \tau}X_0\ |\ \overline{X}_{\tau} = x\right)\right\|
	&\quad \leq \left\|\mathbb{E}\left[\left(x - \sqrt{1 - \tau}X_0\right)\left(x - \sqrt{1 - \tau}X_0\right)^\top\ |\ \overline{X}_{\tau} = x\right]\right\|\\
	&\quad \leq \mathbb{E}\left[\left\|x - \sqrt{1 - \tau}X_0\right\|_2^2\ |\ \overline{X}_{\tau} = x\right]\\
	&\quad \lesssim d\theta_{\tau}(x)\tau\log T.
\end{align*}
Here, the last line holds due to Lemma \ref{lemma:tail_bound}. Then we immediately have
\begin{align*}
	\left\|\frac{\partial s_{\tau}^\star(x_{\tau})}{\partial x}\right\| \lesssim \frac{1}{\tau} + \frac{1}{\tau^2}d\theta_{\tau}(x)\tau\log T \asymp \frac{d\theta_{\tau}(x)\log T}{\tau}.
\end{align*}

Eqn. \eqref{score_derivative_difference} follows by an argument similar to that of \citet[Claim (89)]{li2024provable}; we omit the details for brevity.

\subsection{Proof of Lemma \ref{lemma:log_density}}\label{proof:Lemma_log_density}
We let $J_{\tau}$ denote the Jacobian matrix
\begin{align}
	J_{\tau}(\sqrt{1-\tau}x) &:= \frac{\partial}{\partial x}\frac{s_{\tau}^\star(\sqrt{1-\tau}x)}{(1-\tau)^{3/2}}\notag\\
	&= -\frac{1}{\tau(1-\tau)}I_d + \frac{1}{\tau^2}\Big(\mathbb{E}\big[(x-X_0)(x-X_0)^\top~\big |~\overline{X}_{\tau} = \sqrt{1-\tau}x\big]\notag\\&\hspace{3.5cm}- \mathbb{E}\big[x-X_0~\big |~\overline{X}_{\tau} = -\sqrt{1-\tau}x\big]\mathbb{E}\big[x-X_0~\big |~\overline{X}_{\tau} = \sqrt{1-\tau}x\big]^\top\Big).\label{def:Jacobian}
\end{align}
We define
\begin{align*}
	\theta' = \sup_{\tau': |\tau'-\widetilde\tau| \leq c_0\widetilde\tau(1-\widetilde\tau)}\frac{-\log p_{\overline{X}_{\tau'}}(x_{\tau'}^\star)}{d\log T} < \infty.
\end{align*}
By virtue of Lemma \ref{lemma:tail_bound}, we have
\begin{align}\label{ineq19}
	\mathsf{tr}\left(J_{\tau}(x_{\tau}^\star)\right) &\leq -\frac{d}{\tau(1-\tau)} + \frac{1}{\tau^2(1-\tau)}\mathsf{tr}\left(\mathbb{E}\big[(x_{\tau}^\star-\sqrt{1-\tau}X_0)(x_{\tau}^\star-\sqrt{1-\tau}X_0)^\top~\big |~\overline{X}_{\tau} = x_{\tau}^\star\big]\right)\notag\\
	&= -\frac{d}{\tau(1-\tau)} + \frac{1}{\tau^2(1-\tau)}\mathbb{E}\left[\left\|x_{\tau}^\star-\sqrt{1-\tau}X_0\right\|_2^2~\big |~\overline{X}_{\tau} = x_{\tau}^\star\right]\notag\\
	&\leq -\frac{d}{\tau(1-\tau)} + \frac{120}{\tau^2(1-\tau)}\theta'd\tau\log T \leq \frac{120}{\tau(1-\tau)}\theta'd\log T.
\end{align}
In addition, \eqref{eq:ODE}tells us that 
\begin{align}\label{eq:relation_initial}
	\frac{\partial}{\partial\tau}\frac{\partial x_{\tau}^\star/\sqrt{1-\tau}}{\partial x_{\tau'}^\star/\sqrt{1-\tau'}} &= \frac{\partial}{\partial x_{\tau'}^\star/\sqrt{1-\tau'}}\frac{\partial x_{\tau}^\star/\sqrt{1-\tau}}{\partial\tau}\notag\\
	&= \frac{\partial}{\partial x_{\tau}^\star/\sqrt{1-\tau}}\left(-\frac{1}{2(1-\tau)^{3/2}}s_{\tau}^{\star}(x_\tau^{\star})\right)\notag\\
	&= \frac{\partial}{\partial x_{\tau'}^\star/\sqrt{1-\tau'}}\left(-\frac{1}{2(1-\tau)^{3/2}}s_{\tau}^{\star}(x_\tau^{\star})\right)\frac{\partial x_{\tau}^\star/\sqrt{1-\tau}}{\partial x_{\tau'}^\star/\sqrt{1-\tau'}}\notag\\
	&= -\frac{1}{2}J_{\tau}\left(x_{\tau}^\star\right)\frac{\partial x_{\tau}^\star/\sqrt{1-\tau}}{\partial x_{\tau'}^\star/\sqrt{1-\tau'}}.
\end{align}
Applying Jacobi's formula
\begin{align*}
	\frac{{\rm d}}{{\rm d}t}\det A(t) = (\det A(t))\cdot{\rm tr}\left(A(t)^{-1}\cdot \frac{{\rm d}A(t)}{{\rm d}t}\right)
\end{align*}
yields
\begin{align}\label{eq7}
	\frac{\partial}{\partial\tau}\det\left(\frac{\partial x_{\tau}^\star/\sqrt{1-\tau}}{\partial x_{\tau'}^\star/\sqrt{1-\tau'}}\right) &= \det \left(\frac{\partial x_{\tau}^\star/\sqrt{1-\tau}}{\partial x_{\tau'}^\star/\sqrt{1-\tau'}}\right)\cdot{\rm tr}\left(\left(\frac{\partial x_{\tau}^\star/\sqrt{1-\tau}}{\partial x_{\tau'}^\star/\sqrt{1-\tau'}}\right)^{-1}\cdot	\frac{\partial}{\partial\tau}\frac{\partial x_{\tau}^\star/\sqrt{1-\tau}}{\partial x_{\tau'}^\star/\sqrt{1-\tau'}}\right)\notag\\
	&= -\frac{1}{2}{\rm tr}\left(J_{\tau}\left(x_{\tau}^\star\right)\right)\det \left(\frac{\partial x_{\tau}^\star/\sqrt{1-\tau}}{\partial x_{\tau'}^\star/\sqrt{1-\tau'}}\right).
\end{align}
By solving this equation, we know that
\begin{align}\label{eq8}
	\det\left(\frac{\partial x_{\tau''}^\star/\sqrt{1-\tau''}}{\partial x_{\tau'}^\star/\sqrt{1-\tau'}}\right) = \exp\left(-\frac{1}{2}\int_{\tau'}^{\tau''}{\rm tr}\left(J_{\tau}\left(x_{\tau}^\star\right)\right){\rm d}\tau\right)
\end{align} 
for all $\tau', \tau''$. In view of \eqref{ineq19} and \eqref{eq7}, one has
\begin{align}\label{ineq20}
	%\frac{{\rm d}}{{\rm d}\tau'}\log p_{X_{\tau'}/\sqrt{1-\tau'}}\left(x_{\tau'}^\star/\sqrt{1-\tau'}\right) &= 
	\frac{{\rm d}}{{\rm d}\tau'}\bigg(\log \frac{p_{\overline{X}_{\tau'}/\sqrt{1-\tau'}}\left(x_{\tau'}^\star/\sqrt{1-\tau'}\right)}{p_{\overline{X}_{\widetilde\tau}/\sqrt{1-\widetilde\tau}}\left(x_{\widetilde\tau}^\star/\sqrt{1-\widetilde\tau}\right)}\bigg) = -\frac{{\rm d}}{{\rm d}\tau'}\log \det\left(\frac{\partial x_{\tau'}^\star/\sqrt{1-\tau'}}{\partial x_{\widetilde\tau}^\star/\sqrt{1-\widetilde\tau}}\right) = \frac{1}{2}{\rm tr}\left(J_{\tau'}\left(x_{\tau'}^\star\right)\right).
\end{align}
Moreover, recognizing that for all $|\tau'-\widetilde{\tau}| \leq c_0\widetilde{\tau}(1-\widetilde{\tau})$,
\begin{align*}
	\left|\frac{1}{\tau'(1-\tau')} - \frac{1}{\widetilde{\tau}(1-\widetilde{\tau})}\right| = \frac{\left|(\tau'-\widetilde{\tau})(1-\tau'-\widetilde{\tau})\right|}{\tau'(1-\tau')\widetilde{\tau}(1-\widetilde{\tau})} \leq \frac{c_0\widetilde{\tau}(1-\widetilde{\tau})}{\tau'(1-\tau')\widetilde{\tau}(1-\widetilde{\tau})} = \frac{c_0}{\tau'(1-\tau')},
\end{align*}
we have
\begin{align*}
	\frac{1}{\tau'(1-\tau')} \leq \frac{1}{1-c_0}\frac{1}{\widetilde{\tau}(1-\widetilde{\tau})} \leq \frac{2}{\widetilde{\tau}(1-\widetilde{\tau})}.
\end{align*}
Combining \eqref{ineq19}, \eqref{ineq20} and the previous inequality gives us 
\begin{align*}
	-\log p_{\overline{X}_{\tau'}/\sqrt{1-\tau'}}\big(x_{\tau'}^\star/\sqrt{1-\tau'}\big) \leq -\log p_{\overline{X}_{\widetilde\tau}/\sqrt{1-\widetilde\tau}}\big(x_{\widetilde\tau}^\star/\sqrt{1-\widetilde\tau}\big) + |\tau' - \widetilde{\tau}|\frac{120}{\tau(1-\tau)}\theta'd\log T.
\end{align*}
Furthermore, by noting that $p_{\overline{X}_{\tau}/\sqrt{1-\tau}}(x_{\tau}^\star/\sqrt{1-\tau}) = (1-\tau)^{d/2}p_{\overline{X}_{\tau}}(x_{\tau}^\star)$, we have
\begin{align*}
	-\log p_{\overline{X}_{\tau'}}(x_{\tau'}^\star) &\leq -\log p_{\overline{X}_{\widetilde\tau}}(x_{\widetilde\tau}^\star) + d\log\left(\frac{1-\tau'}{1-\widetilde{\tau}}\right) + |\tau' - \widetilde{\tau}|\frac{120}{\tau(1-\tau)}\theta'd\log T\\
	&\leq -\log p_{\overline{X}_{\widetilde\tau}}(x_{\widetilde\tau}^\star) + d\frac{|\tau'-\widetilde{\tau}|}{1-\widetilde{\tau}} + 60c_0\theta'd\log T\\
	&\leq \theta d\log T + c_0d\tau + 60c_0\theta'd\log T\\
	&\leq \theta d\log T + \frac{1}{2}\theta'd\log T,
\end{align*}
provided that $|\tau'-\widetilde{\tau}| \leq c_0\widetilde{\tau}(1-\widetilde{\tau})$. Here, the second inequality makes use of $\log(1+x) \leq x$. Recalling the definition of $\theta'$, we know that
\begin{align*}
	\theta'd\log T = \sup_{\tau': |\tau'-\widetilde\tau| \leq c_0\widetilde\tau(1-\widetilde\tau)}-\log p_{\overline{X}_{\tau'}}(x_{\tau'}^\star) \leq \theta d\log T + \frac{1}{2}\theta'd\log T,
\end{align*}
which implies
\begin{align*}
	\theta' \leq 2\theta.
\end{align*}
Therefore, for all $|\tau'-\widetilde{\tau}| \leq c_0\widetilde{\tau}(1-\widetilde{\tau})$, we have
\begin{align*}
	-\log p_{\overline{X}_{\tau'}}(x_{\tau'}^\star) \leq 2\theta d\log T,
\end{align*}
which has finished the proof of Lemma \ref{lemma:log_density}.

\subsection{Proof of Lemma \ref{lemma:density_ratio_1}}\label{sec:proof_lemma_density_ratio_1}
In addition, elementary calculations show that
\begin{align*}
	\frac{p_{\sqrt{\alpha_t}X_{t-1}}(\sqrt{\alpha_t}x_{t-1})}{p_{\sqrt{\alpha_t}X_{t-1}}(\sqrt{\alpha_t}x_{t-1}^{\star})}
	&= \int_{x_0} p_{X_0\mymid \sqrt{\alpha_t}X_{t-1}}(x_0 | \sqrt{\alpha_t}x_{t-1}^{\star})
	\exp\Big(-\frac{\|x_{t-1} - \sqrt{\overline{\alpha}_{t-1}}x_0\|_2^2 - \|x_{t-1}^{\star} - \sqrt{\overline{\alpha}_{t-1}}x_0\|_2^2}{2(1 - \overline{\alpha}_{t-1})}\Big)
	\mathrm{d}x_0.
\end{align*}
We define $u := x_{t-1} - x_{t-1}^{\star}$. Then we have
\begin{align}\label{ineq17}
	\frac{p_{\sqrt{\alpha_t}X_{t-1}}(\sqrt{\alpha_t}x_{t-1})}{p_{\sqrt{\alpha_t}X_{t-1}}(\sqrt{\alpha_t}x_{t-1}^{\star})}
	&= \int_{x_0} p_{X_0\mymid \sqrt{\alpha_t}X_{t-1}}(x_0 | \sqrt{\alpha_t}x_{t-1}^{\star})\exp\Big(-\frac{\|u\|_2^2 + 2u^\top(x_{t-1}^{\star} - \sqrt{\overline{\alpha}_{t-1}}x_0)}{2(1 - \overline{\alpha}_{t-1})}\Big)\mathrm{d}x_0\notag\\
	&= \exp\Big(-\frac{\|u\|_2^2}{2(1 - \overline{\alpha}_{t-1})}\Big)\int_{x_0} p_{X_0\mymid \sqrt{\alpha_t}X_{t-1}}(x_0 | \sqrt{\alpha_t}x_{t-1}^{\star})\exp\Big(-\frac{u^\top(x_{t-1}^{\star} - \sqrt{\overline{\alpha}_{t-1}}x_0)}{1 - \overline{\alpha}_{t-1}}\Big)\mathrm{d}x_0\notag\\
	&\leq \exp\Big(-\frac{\|u\|_2^2}{2(1 - \overline{\alpha}_{t-1})}\Big)\mathbb{E}\left[\exp\Big(-\frac{u^\top(x_{t-1}^{\star} - \sqrt{\overline{\alpha}_{t-1}}X_0)}{1 - \overline{\alpha}_{t-1}}\Big)~\Big |~X_{t-1} = x_{t-1}^{\star}\right]\notag\\
	&\leq \exp\Big(-\frac{\|u\|_2^2}{2(1 - \overline{\alpha}_{t-1})}\Big)\mathbb{E}\left[\exp\Big(\frac{\|u\|_2\|x_{t-1}^{\star} - \sqrt{\overline{\alpha}_{t-1}}X_0\|_2}{1 - \overline{\alpha}_{t-1}}\Big)~\Big |~X_{t-1} = x_{t-1}^{\star}\right].
\end{align}
By virtue of Lemma \ref{lemma:tail_bound}, one has
\begin{align}\label{ineq21}
	&\mathbb{E}\left[\exp\Big(\frac{\|u\|_2\|x_{t-1}^{\star} - \sqrt{\overline{\alpha}_{t-1}}X_0\|_2}{1 - \overline{\alpha}_{t-1}}\Big)~\Big |~X_{t-1} = x_{t-1}^{\star}\right]\notag\\
	&\quad=\int_{0}^{\infty}\mathbb{P}\left(\exp\Big(\frac{\|u\|_2\|x_{t-1}^{\star} - \sqrt{\overline{\alpha}_{t-1}}X_0\|_2}{1 - \overline{\alpha}_{t-1}}\Big) > w~\Big |~X_{t-1} = x_{t-1}^{\star}\right){\rm d}w\notag\\
	&\quad=1 + \int_{1}^{\infty}\mathbb{P}\left(\exp\Big(\frac{\|u\|_2\|x_{t-1}^{\star} - \sqrt{\overline{\alpha}_{t-1}}X_0\|_2}{1 - \overline{\alpha}_{t-1}}\Big) > w~\Big |~X_{t-1} = x_{t-1}^{\star}\right){\rm d}w\notag\\
	&\quad\stackrel{v = \frac{\sqrt{1 - \overline{\alpha}_{t-1}}\log w}{\|u\|_2}}{=}1 + \int_{0}^{\infty}e^{\frac{\|u\|_2v}{\sqrt{1 - \overline{\alpha}_{t-1}}}}\frac{\|u\|_2}{\sqrt{1 - \overline{\alpha}_{t-1}}}\mathbb{P}\left(\|x_{t-1}^{\star} - \sqrt{\overline{\alpha}_{t-1}}X_0\|_2 > \sqrt{1 - \overline{\alpha}_{t-1}}v~\Big |~X_{t-1} = x_{t-1}^{\star}\right){\rm d}v\notag\\
	&\quad\leq 1 + 5c_5\sqrt{c_6d\log T}\cdot \frac{\|u\|_2}{\sqrt{1 - \overline{\alpha}_{t-1}}}\exp\left(\frac{\|u\|_2\cdot 5c_5\sqrt{c_6d\log T}}{\sqrt{1 - \overline{\alpha}_{t-1}}}\right) + \int_{5c_5\sqrt{c_6d\log T}}^{\infty}e^{\frac{\|u\|_2v}{\sqrt{1 - \overline{\alpha}_{t-1}}}}\frac{\|u\|_2}{\sqrt{1 - \overline{\alpha}_{t-1}}}e^{-\frac{v^2}{25}}{\rm d}v\notag\\
	&\quad\leq 1 + 5c_5\sqrt{c_6d\log T}\cdot \frac{\|u\|_2}{\sqrt{1 - \overline{\alpha}_{t-1}}}\exp\left(\frac{\|u\|_2\cdot 5c_5\sqrt{c_6d\log T}}{\sqrt{1 - \overline{\alpha}_{t-1}}}\right)\notag\\&\qquad + \frac{\|u\|_2}{\sqrt{1 - \overline{\alpha}_{t-1}}}\exp\left(\frac{25\|u\|_2^2}{4(1-\overline{\alpha}_{t-1})}\right)\int_{-\infty}^{\infty}\exp\bigg(-\frac{(v - \frac{25\|u\|_2}{2\sqrt{1 - \overline{\alpha}_{t-1}}})^2}{25}\bigg){\rm d}v\notag\\
	&\quad= 1 + 5c_5\sqrt{c_6d\log T}\cdot \frac{\|u\|_2}{\sqrt{1 - \overline{\alpha}_{t-1}}}\exp\left(\frac{\|u\|_2\cdot 5c_5\sqrt{c_6d\log T}}{\sqrt{1 - \overline{\alpha}_{t-1}}}\right) + 5\sqrt{2\pi}\frac{\|u\|_2}{\sqrt{1 - \overline{\alpha}_{t-1}}}\exp\left(\frac{25\|u\|_2^2}{4(1-\overline{\alpha}_{t-1})}\right)\notag\\
	&\quad\leq \exp\bigg(O\bigg(\frac{\|x_{t-1} - x_{t-1}^{\star}\|_2^2}{1 - \overline{\alpha}_{t-1}} + \sqrt{\frac{d\|x_{t-1} - x_{t-1}^{\star}\|_2^2\log T}{1 - \overline{\alpha}_{t-1}}}\bigg)\bigg).
\end{align}
where the last line holds since $1 + a\exp(b) + c\exp(d) \leq (1+a+c)\exp(\max\{b,d\}) \leq \exp(a+b+c+d)$ for all $a, b , c, d \geq 0$. Combining \eqref{ineq17} and the previous inequality, we have finished the proof of \eqref{ineq18}.

In addition, we make the reservation that $p_{\phi(Y)}(\phi(x)) = \mathsf{det}\Big(\frac{\partial \phi(x)}{\partial x}\Big)^{-1}p_{Y}(x)$. Then we can write
\begin{align*}
	\frac{p_{\sqrt{\alpha_t}Y_{t-1}}(\sqrt{\alpha_t}x_{t-1})}{p_{\sqrt{\alpha_t}Y_{t-1}^{\star}}(\sqrt{\alpha_t}x_{t-1}^{\star})}
	&= \frac{p_{\frac{1}{\sqrt{\overline{\alpha}_{t-1}}}Y_{t-1}}\big(\frac{1}{\sqrt{\overline{\alpha}_{t-1}}}x_{t-1}\big)}{p_{\frac{1}{\sqrt{\overline{\alpha}_{t-1}}}Y_{t-1}^\star}\big(\frac{1}{\sqrt{\overline{\alpha}_{t-1}}}x_{t-1}^\star\big)}\\
	&\stackrel{x_{\tau_t,0}=x_{\tau_t,0}^\star}{=} \frac{p_{\frac{1}{\sqrt{1-\tau_{t,K-1}}}Y_{t-1}}\big(\frac{1}{\sqrt{1-\tau_{t,K-1}}}x_{\tau_{t, K-1}^{(N)}}\big)/p_{\frac{1}{\sqrt{1-\tau_{t,0}}}Y_{t}}\big(\frac{1}{1-\tau_{t, 0}}x_{\tau_{t,0}}\big)}{p_{\frac{1}{\sqrt{\overline{\alpha}_{t-1}}}Y_{t-1}^\star}\big(\frac{1}{\sqrt{\overline{\alpha}_{t-1}}}x_{t-1}^\star\big)/p_{\frac{1}{\sqrt{1-\tau_{t,0}}}Y_{t}}\big(\frac{1}{1-\tau_{t, 0}}x_{\tau_{t,0}}^\star\big)}\\
	&= \mathsf{det}\bigg(\frac{\partial x_{\tau_{t, K-1}}^{\star}/\sqrt{1-\tau_{t, K-1}}}{\partial x_{\tau_{t, 0}}^{\star}/\sqrt{1-\tau_{t, 0}}}\bigg) / \mathsf{det}\bigg(\frac{\partial x_{\tau_{t, K-1}}^{(N)}/\sqrt{1-\tau_{t, K-1}}}{\partial x_{\tau_{t, 0}}/\sqrt{1-\tau_{t, 0}}}\bigg)\\
	&= 1 + \frac{\mathsf{det}\bigg(\frac{\partial x_{\tau_{t, K-1}}^{\star}/\sqrt{1-\tau_{t, K-1}}}{\partial x_{\tau_{t, 0}}^{\star}/\sqrt{1-\tau_{t, 0}}}\bigg) - \mathsf{det}\bigg(\frac{\partial x_{\tau_{t, K-1}}^{(N)}/\sqrt{1-\tau_{t, K-1}}}{\partial x_{\tau_{t, 0}}/\sqrt{1-\tau_{t, 0}}}\bigg)}{\mathsf{det}\bigg(\frac{\partial x_{\tau_{t, K-1}}^{(N)}/\sqrt{1-\tau_{t, K-1}}}{\partial x_{\tau_{t, 0}}/\sqrt{1-\tau_{t, 0}}}\bigg)}.
\end{align*}
We know from \citet[Corollary 2.14]{ipsen2008perturbation} that
\begin{align*}
	&\left|\frac{\mathsf{det}\bigg(\frac{\partial x_{\tau_{t, K-1}}^{\star}/\sqrt{1-\tau_{t, K-1}}}{\partial x_{\tau_{t, 0}}^{\star}/\sqrt{1-\tau_{t, 0}}}\bigg) - \mathsf{det}\bigg(\frac{\partial x_{\tau_{t, K-1}}^{(N)}/\sqrt{1-\tau_{t, K-1}}}{\partial x_{\tau_{t, 0}}/\sqrt{1-\tau_{t, 0}}}\bigg)}{\mathsf{det}\bigg(\frac{\partial x_{\tau_{t, K-1}}^{(N)}/\sqrt{1-\tau_{t, K-1}}}{\partial x_{\tau_{t, 0}}/\sqrt{1-\tau_{t, 0}}}\bigg)}\right|\\ &\quad\leq \bigg(\bigg\|\frac{\partial x_{\tau_{t, K-1}}^{(N)}/\sqrt{1-\tau_{t, K-1}}}{\partial x_{\tau_{t, 0}}/\sqrt{1-\tau_{t, 0}}}\bigg\|^{-1}\bigg\|\frac{\partial x_{\tau_{t, K-1}}^{\star}/\sqrt{1-\tau_{t, K-1}}}{\partial x_{\tau_{t, 0}}^{\star}/\sqrt{1-\tau_{t, 0}}}-\frac{\partial x_{\tau_{t, K-1}}^{(N)}/\sqrt{1-\tau_{t, K-1}}}{\partial x_{\tau_{t, 0}}/\sqrt{1-\tau_{t, 0}}}\bigg\| + 1\bigg)^d - 1\\
	&\quad\leq \exp\bigg(d\bigg\|\frac{\partial x_{\tau_{t, K-1}}^{(N)}/\sqrt{1-\tau_{t, K-1}}}{\partial x_{\tau_{t, 0}}/\sqrt{1-\tau_{t, 0}}}\bigg\|^{-1}\bigg\|\frac{\partial x_{\tau_{t, K-1}}^{\star}/\sqrt{1-\tau_{t, K-1}}}{\partial x_{\tau_{t, 0}}^{\star}/\sqrt{1-\tau_{t, 0}}}-\frac{\partial x_{\tau_{t, K-1}}^{(N)}/\sqrt{1-\tau_{t, K-1}}}{\partial x_{\tau_{t, 0}}/\sqrt{1-\tau_{t, 0}}}\bigg\|\bigg) - 1.
\end{align*}
The previous two inequalities together tell us that \eqref{ineq39} holds if $\bigg\|\frac{\partial x_{\tau_{t, K-1}}^{(N)}/\sqrt{1-\tau_{t, K-1}}}{\partial x_{\tau_{t, 0}}/\sqrt{1-\tau_{t, 0}}}\bigg\|^{-1} \lesssim 1$.

\subsection{Proof of Lemma \ref{lemma:log_density_ratio_1}}
We prove Lemma \ref{lemma:log_density_ratio_1} by induction. For notational convenience, we let $\theta_t := \theta_{\tau_{t,0}}(x_{\tau_{t,i}}^{(0)})$. Recalling that $x_{\tau}^\star$ is the solution of ODE \eqref{eq:ODE} at $\tau$ with the initial condition $x_{\tau_{t,0}}^\star = x_{\tau_{t,0}}$, we know from Lemma \ref{lemma:log_density} that $-\log p_{\overline{X}_\tau}(x_{\tau}^\star) \leq 2\theta_td\log T$ for all $\tau_{t,K-1} \leq \tau \leq \tau_{t, 0}$. 
For all $\lambda \in [0,1]$, one can derive
\begin{align*}
	\frac{p_{\overline{X}_{\tau_{t,i}}}\big(\lambda x_{\tau_{t,i}}^{(0)} + (1-\lambda)x_{\tau_{t,i}}^\star\big)}{p_{\overline{X}_{\tau_{t,i}}}(x_{\tau_{t,i}}^\star)}
	&= \exp\Big(-\frac{\|u\|_2^2}{2\tau_{t, i}}\Big)\int_{x_0} p_{X_0 \mymid \overline{X}_{\tau_{t,i}}}\left(x_0 \mymid x_{\tau_{t,i}}^{\star}\right)\exp\Big(-\frac{u^\top(x_{\tau_{t,i}}^\star - \sqrt{1-\tau_{t, i}}x_0)}{\tau_{t, i}}\Big)\mathrm{d}x_0\\
	&\geq \exp\Big(-\frac{\|u\|_2^2}{2\tau_{t, i}}\Big)\mathbb{E}\left[\exp\Big(-\frac{\|u\|_2\|x_{\tau_{t,i}}^\star - \sqrt{1-\tau_{t, i}}X_0\|_2}{\tau_{t, i}}\Big)~\Big |~\overline{X}_{\tau_{t,i}} = x_{\tau_{t,i}}^{\star}\right],
	%\\\quad&= \frac{p_{\sqrt{\frac{1-\tau_{t,0}}{1-\tau_{t,i}}}X_{\tau_{t,i}}}\Big(\sqrt{\frac{1-\tau_{t,0}}{1-\tau_{t,i}}}x_{\tau_{t,i}}^{(0)}\Big)}{p_{\sqrt{\frac{1-\tau_{t,0}}{1-\tau_{t,i}}}X_{\tau_{t,i}}}\Big(\sqrt{\frac{1-\tau_{t,0}}{1-\tau_{t,i}}}x_{\tau_{t,i}}^\star\Big)}\\
	%\quad&= \exp\Big(-\frac{\|u\|_2^2}{2\tau_{t, i}}\Big)\int_{x_0} p_{X_0~\big |~ \sqrt{\frac{1-\tau_{t,0}}{1-\tau_{t,i}}}X_{\tau_{t,i}}}\bigg(x_0~\bigg |~ \sqrt{\frac{1-\tau_{t,0}}{1-\tau_{t,i}}}x_{\tau_{t,i}}^{\star}\bigg)\exp\Big(-\frac{u^\top(x_{t-1}^{\star} - \sqrt{\overline{\alpha}_{t-1}}x_0)}{\tau_{t, i}}\Big)\mathrm{d}x_0
\end{align*}
where $u = \lambda(x_{\tau_{t,i}}^{(0)} - x_{\tau_{t,i}}^\star)$.
In view of \eqref{ineq21} and Jensen's inequality, we know that
\begin{align*}
	&\mathbb{E}\left[\exp\Big(-\frac{\|u\|_2\|x_{\tau_{t,i}}^\star - \sqrt{1-\tau_{t, i}}X_0\|_2}{\tau_{t, i}}\Big)~\Big |~\overline{X}_{\tau_{t,i}} = x_{\tau_{t,i}}^{\star}\right]\\ &\quad \geq \left(\mathbb{E}\left[\exp\Big(\frac{\|u\|_2\|x_{\tau_{t,i}}^\star - \sqrt{1-\tau_{t, i}}X_0\|_2}{\tau_{t, i}}\Big)~\Big |~\overline{X}_{\tau_{t,i}} = x_{\tau_{t,i}}^{\star}\right]\right)^{-1}\\
	&\quad \geq \exp\left(-O\left(\frac{\big\|x_{\tau_{t,i}}^{(0)} - x_{\tau_{t,i}}^\star\big\|_2^2}{\tau_{t,i}} + \sqrt{\frac{d\big\|x_{\tau_{t,i}}^{(0)} - x_{\tau_{t,i}}^\star\big\|_2^2\log T}{\tau_{t,i}}}\right)\right)\\
	&\quad \geq \exp\left(-O\left(\frac{\big\|x_{\tau_{t,i}}^{(0)} - x_{\tau_{t,i}}^\star\big\|_2^2}{1-\overline{\alpha}_{t-1}} + \sqrt{\frac{d\big\|x_{\tau_{t,i}}^{(0)} - x_{\tau_{t,i}}^\star\big\|_2^2\log T}{1 - \overline{\alpha}_{t-1}}}\right)\right).
\end{align*}
where the last line makes use of $1 - \overline{\alpha}_{t-1} \asymp \tau_{t,i}$.
%Repeating similar arguments as in \eqref{ineq18}, one has
%\begin{align}
%	\frac{p_{X_{\tau_{t,i}}}(x_{\tau_{t,i}}^{(0)})}{p_{X_{\tau_{t,i}}}(x_{\tau_{t,i}}^\star)} = \exp\left(O\left(\frac{\big\|x_{\tau_{t,i}}^{(0)} - x_{\tau_{t,i}}^\star\big\|_2^2}{1-\overline{\alpha}_{t-1}} + \sqrt{\frac{d\big\|x_{\tau_{t,i}}^{(0)} - x_{\tau_{t,i}}^\star\big\|_2^2\log T}{1 - \overline{\alpha}_{t-1}}}\right)\right).
%\end{align}
By virtue of \eqref{ineq13}, one has
\begin{align}\label{ineq22}
	\big\|x_{\tau_{t,i}}^{(0)} - x_{\tau_{t,i}}^\star\big\|_2 \lesssim \overline{\alpha}_{t-1}(1-\alpha_t)\frac{1}{\sqrt{1-\overline{\alpha}_{t-1}}}\sqrt{\theta_t d\log T}.
\end{align}
The previous two inequalities together with Lemma \ref{lemma:learning_rate} and \eqref{ineq18} imply that
\begin{align*}
	\left|\log\frac{p_{\overline{X}_{\tau_{t,i}}}\big(\lambda x_{\tau_{t,i}}^{(0)} + (1-\lambda)x_{\tau_{t,i}}^\star\big)}{p_{\overline{X}_{\tau_{t,i}}}(x_{\tau_{t,i}}^\star)}\right| = O\left(\frac{\theta_td\log^3 T}{T^2} + \frac{\sqrt{\theta_t}d\log^2 T}{T}\right) = O\left(\frac{\sqrt{\theta_t}d\log^2 T}{T}\right).
\end{align*}
Therefore, for any $i$ and $\lambda \in [0,1]$, we have
\begin{align}\label{ineq23}
	-\log p_{\overline{X}_{\tau_{t,i}}}\big(\lambda x_{\tau_{t,i}}^{(0)} + (1-\lambda)x_{\tau_{t,i}}^\star\big) \leq 2.1d\theta_t\log T.
\end{align}
Then Lemma \ref{lemma:Hessian} tells us that
\begin{align}\label{ineq24}
	\big\|s_{\tau_{t,i}}^\star\big(\lambda x_{\tau_{t,i}}^{(0)} + (1-\lambda)x_{\tau_{t,i}}^\star\big)\big\| \lesssim \sqrt{\frac{d\theta_t\log T}{\tau_{t,i}}},~\quad~\forall 0 \leq i \leq K-1.
\end{align}
We define
\begin{align*}
	u_{t,i}^{(n)} &:= \sqrt{\frac{1-\tau_{t,0}}{1-\tau_{t,i}}}x_{\tau_{t,i}}^{(n)} - x_{\tau_{t,0}}\\
	&= \sqrt{1-\tau_{t, 0}}\sum_{0 \le j < K}\gamma_{t, j}(\tau_{t,i})(1-\tau_{t, j})^{-3/2}s_{\tau_{t, j}}(x_{\tau_{t, j}}^{(n-1)}).
\end{align*}
For $n = 1$, recognizing that 
\begin{align*}
	\sqrt{\frac{1-\tau_{t,0}}{1-\tau_{t,i}}}\overline{X}_{\tau_{t,i}} \stackrel{{\rm d.}}{=} \sqrt{1-\tau_{t,0}}X_0 + \sqrt{\frac{(1-\tau_{t,0})\tau_{t,i}}{1-\tau_{t,i}}}Z,~\quad~\text{where}~Z \sim N(0, I_d),
\end{align*}
one can apply the Bayesian rule to derive
\begin{align*}
	&p_{\sqrt{\frac{1-\tau_{t,0}}{1-\tau_{t,i}}}\overline{X}_{\tau_{t,i}}}\left(\sqrt{\frac{1-\tau_{t,0}}{1-\tau_{t,i}}}x_{\tau_{t,i}}^{(1)}\right)\\&\quad = \int_{x_0}p_{X_0}(x_0)p_{\sqrt{\frac{1-\tau_{t,0}}{1-\tau_{t,i}}}\overline{X}_{\tau_{t,i}}\mymid X_0}\left(\sqrt{\frac{1-\tau_{t,0}}{1-\tau_{t,i}}}x_{\tau_{t,i}}^{(1)}~\Big |~x_0\right){\rm  d}x_0\\
	&\quad= \int_{x_0}p_{X_0}(x_0)\frac{1}{\left(2\pi\frac{(1-\tau_{t,0})\tau_{t,i}}{1-\tau_{t,i}} \right)^{d/2}}\exp\left(-\frac{\left\|\sqrt{\frac{1-\tau_{t,0}}{1-\tau_{t,i}}}x_{\tau_{t,i}}^{(1)} - \sqrt{1-\tau_{t,0}}x_0\right\|_2^2}{2\frac{(1-\tau_{t,0})\tau_{t,i}}{1-\tau_{t,i}}}\right){\rm  d}x_0\\
	&\quad= \int_{x_0}p_{X_0}(x_0)\frac{1}{\left(2\pi\frac{(1-\tau_{t,0})\tau_{t,i}}{1-\tau_{t,i}} \right)^{d/2}}\exp\left(-\frac{\left\|\sqrt{\frac{1-\tau_{t,0}}{1-\tau_{t,i}}}x_{\tau_{t,i}}^{(1)} - \sqrt{1-\tau_{t,0}}x_0\right\|_2^2}{2\frac{(1-\tau_{t,0})\tau_{t,i}}{1-\tau_{t,i}}}\right){\rm  d}x_0\\
	&\quad= \int_{x_0}p_{X_0}(x_0)\frac{1}{\left(2\pi\frac{(1-\tau_{t,0})\tau_{t,i}}{1-\tau_{t,i}} \right)^{d/2}}\exp\left(-\frac{\left\|x_{\tau_{t,0}} - \sqrt{1-\tau_{t,0}}x_0\right\|_2^2}{2\tau_{t,0}}\right)\\
	&\qquad \cdot\exp\left(-\frac{(\tau_{t,0} - \tau_{t, i})\left\|x_{\tau_{t,0}} - \sqrt{1-\tau_{t,0}}x_0\right\|_2^2}{2(1-\tau_{t,0})\tau_{t,0}\tau_{t,i}} -\frac{\big\|u_{t,i}^{(1)}\big\|_2^2 + 2\big<u_{t,i}^{(1)}, x_{\tau_{t,0}} - \sqrt{1-\tau_{t,0}}x_0\big>}{2\frac{(1-\tau_{t,0})\tau_{t,i}}{1-\tau_{t,i}}}\right){\rm  d}x_0.
\end{align*}
Therefore, we can rewrite
\begin{align}\label{ineq25}
	&\frac{p_{\sqrt{\frac{1-\tau_{t,0}}{1-\tau_{t,i}}}\overline{X}_{\tau_{t,i}}}\left(\sqrt{\frac{1-\tau_{t,0}}{1-\tau_{t,i}}}x_{\tau_{t,i}}^{(1)}\right)}{p_{\overline{X}_{\tau_{t,0}}}(x_{\tau_{t,0}})}\notag\\ &\quad = \frac{1}{p_{\overline{X}_{\tau_{t,0}}}(x_{\tau_{t,0}})}\int_{x_0}p_{X_0}(x_0)p_{\sqrt{\frac{1-\tau_{t,0}}{1-\tau_{t,i}}}\overline{X}_{\tau_{t,i}}\mymid X_0}\left(\sqrt{\frac{1-\tau_{t,0}}{1-\tau_{t,i}}}x_{\tau_{t,i}}^{(1)}~\Big |~x_0\right){\rm  d}x_0\notag\\
	&\quad= \left(\frac{(1-\tau_{t,i})\tau_{t,0}}{(1-\tau_{t,0})\tau_{t,i}}\right)^{d/2}\int_{x_0}p_{X_0\mymid \overline{X}_{\tau_{t, 0}}}(x_0\mymid x_{\tau_{t,0}})\notag\\
	&\qquad \cdot\exp\left(-\frac{(\tau_{t,i} - \tau_{t, 0})\left\|x_{\tau_{t,0}} - \sqrt{1-\tau_{t,0}}x_0\right\|_2^2}{2(1-\tau_{t,0})\tau_{t,0}\tau_{t,i}} -\frac{\big\|u_{t,i}^{(1)}\big\|_2^2 + 2\big<u_{t,i}^{(1)}, x_{\tau_{t,0}} - \sqrt{1-\tau_{t,0}}x_0\big>}{2\frac{(1-\tau_{t,0})\tau_{t,i}}{1-\tau_{t,i}}}\right){\rm  d}x_0\notag\\
	&\quad = \left(1 + \frac{d}{2}\frac{\tau_{t, 0} - \tau_{t, i}}{(1-\tau_{t,0})\tau_{t,i}} + O\left(d^2\left(\frac{\tau_{t, 0} - \tau_{t, i}}{(1-\tau_{t,0})\tau_{t,i}}\right)^2\right)\right)\notag\\
	&\qquad \cdot\int_{x_0}p_{X_0\mymid \overline{X}_{\tau_{t, 0}}}(x_0\mymid x_{\tau_{t,0}})\notag\\&\hspace{2cm}\cdot\exp\left(-\frac{(\tau_{t,0} - \tau_{t, i})\left\|x_{\tau_{t,0}} - \sqrt{1-\tau_{t,0}}x_0\right\|_2^2}{2(1-\tau_{t,0})\tau_{t,0}\tau_{t,i}} -\frac{\big\|u_{t,i}^{(1)}\big\|_2^2 + 2\big<u_{t,i}^{(1)}, x_{\tau_{t,0}} - \sqrt{1-\tau_{t,0}}x_0\big>}{2\frac{(1-\tau_{t,0})\tau_{t,i}}{1-\tau_{t,i}}}\right){\rm  d}x_0.
\end{align}
By virtue of \eqref{ineq24}, we can derive
\begin{align}\label{ineq30}
	&\bigg\|u_{t,i}^{(1)} - \bigg(-\sqrt{1-\tau_{t, 0}}\int_{\tau_{t, 0}}^{\tau_{t,i}} \frac{1}{2(1-\tau)^{3/2}}s_{\tau}^{\star}(x_{\tau}^\star)\mathrm{d}\tau\bigg)\bigg\|_2\notag\\
	&\quad= \sqrt{1-\tau_{t, 0}}\Bigg\|\sum_{0 \le j < K} \gamma_{t, j}(\tau_{t,i})(1-\tau_{t, j})^{-3/2}\big(s_{\tau_{t, j}}(x_{\tau_{t, j}}^{(0)}) - s_{\tau_{t, j}}^\star(x_{\tau_{t, j}}^{(0)})\big)\notag\\&\hspace{2.5cm} + \bigg[\sum_{0 \le j < k} \gamma_{t, j}(\tau_{t,i})(1-\tau_{t, j})^{-3/2} s_{\tau_{t, j}}^{\star}(x_{\tau_{t, j}}^{\star}) - \bigg(-\int_{\tau_{t, 0}}^{\tau'} \frac{1}{2(1-\tau)^{3/2}}s_{\tau}^{\star}(x_{\tau}^\star)\mathrm{d}\tau\bigg)\bigg]\Bigg\|_2\notag\\
	&\quad\leq\sqrt{1-\tau_{t, 0}}\sum_{0 \le j < K} \left|\gamma_{t, j}(\tau_{t,i})\right|(1-\tau_{t, j})^{-3/2}\big\|s_{\tau_{t, j}}(x_{\tau_{t, j}}^{(0)}) - s_{\tau_{t, j}}^\star(x_{\tau_{t, j}}^{(0)})\big\|\notag\\
	&\qquad + \sqrt{1-\tau_{t, 0}}\bigg\|\sum_{0 \le j < K} \gamma_{t, j}(\tau_{t,i})(1-\tau_{t, j})^{-3/2} s_{\tau_{t, j}}^{\star}(x_{\tau_{t, j}}^{\star}) - \bigg(-\int_{\tau_{t, 0}}^{\tau'} \frac{1}{2(1-\tau)^{3/2}}s_{\tau}^{\star}(x_{\tau}^\star)\mathrm{d}\tau\bigg)\bigg\|\notag\\
	&\quad\asymp \sum_{0 \le j < K}(\tau_{t,0} - \tau_{t,i})(1-\tau_{t, 0})^{-1}\big\|s_{\tau_{t, j}}(x_{\tau_{t, j}}^{(0)}) - s_{\tau_{t, j}}^\star(x_{\tau_{t, j}}^{(0)})\big\|_2\notag\\&\qquad + \sqrt{1-\tau_{t, 0}}\sqrt{\frac{d\theta_t(\tau_{t,i}-\tau_{t, 0})^2\log T}{\tau_{t,0}(1-\tau_{t,0})^3}}\bigg(\frac{d\theta_t(\tau_{t, 0}-\tau_{t,i})\log T}{\tau_{t,0}(1-\tau_{t,0})}\bigg)^{K}\notag\\
	&\quad\lesssim \frac{\tau_{t, 0}\log T}{T}\sum_{0 \le j < K}\big\|s_{\tau_{t, j}}(x_{\tau_{t, j}}^{(0)}) - s_{\tau_{t, j}}^\star(x_{\tau_{t, j}}^{(0)})\big\|_2 +  \sqrt{\frac{d\theta_t\tau_{t, 0}\log^3 T}{T^2}}\left(\frac{d\theta_t\log^2T}{T}\right)^K\notag\\
	&\quad\lesssim \frac{\tau_{t, 0}\log T}{T}\sqrt{K\sum_{i}\big(\varepsilon_{\mathsf{score}, t, i}^{(n)}(x_{\tau_{t, i}}^{(0)})\big)^2} +  \sqrt{\frac{d\theta_t\tau_{t, 0}\log^3 T}{T^2}}\left(\frac{d\theta_t\log^2T}{T}\right)^K\notag\\
	&\quad\asymp \frac{\tau_{t, 0}\log T}{T}\sqrt{\sum_{i}\big(\varepsilon_{\mathsf{score}, t, i}^{(0)}(x_{\tau_{t, i}}^{(0)})\big)^2} +  \sqrt{\frac{d\theta_t\tau_{t, 0}\log^3 T}{T^2}}\left(\frac{d\theta_t\log^2T}{T}\right)^K.
\end{align}
Here, the fourth line makes use of \eqref{ineq:learning_rate_g}, \eqref{ineq:learning_rate_h}, \eqref{eq:derivative_1} and Taylor's Theorem; the fifth line holds due to \eqref{ineq:learning_rate_f}; the fifth line comes from Cauchy-Schwarz inequality and the last line is valid since $K$ is a constant. By virtue of Lemmas \ref{lemma:learning_rate}, \ref{lemma:Hessian} and \ref{lemma:log_density}, one has
\begin{align}
	\bigg\|-\sqrt{1-\tau_{t, 0}}\int_{\tau_{t, 0}}^{\tau_{t,i}} \frac{1}{2(1-\tau)^{3/2}}s_{\tau}^{\star}(x_{\tau}^\star)\mathrm{d}\tau\bigg\|_2 &\lesssim \sqrt{1-\tau_{t, 0}}\int_{\tau_{t, i}}^{\tau_{t,0}}\frac{1}{2(1-\tau)^{3/2}}\sqrt{\frac{d\theta_t\log T}{\tau}}d\tau\notag\\
	&\lesssim \sqrt{1-\tau_{t, 0}}\frac{\tau_{t, 0}-\tau_{t, i}}{2(1-\tau_{t,0})^{3/2}}\sqrt{\frac{d\theta_t\log T}{\tau_{t,0}}}\notag\\
	&=\frac{\tau_{t, 0}-\tau_{t, i}}{2(1-\tau_{t,0})\tau_{t,0}}\sqrt{\tau_{t, 0}d\theta_t\log T}\\
	&\lesssim \frac{\sqrt{\tau_{t, 0}d\theta_t\log^3 T}}{T}.\label{ineq34}
\end{align}
Combining the previous two inequalities, we obtain
\begin{align}\label{ineq26}
	\big\|u_{t,i}^{(1)}\big\|_2
	&\lesssim \frac{\sqrt{\tau_{t, 0}d
			\theta_t\log^3 T}}{T} + \frac{\tau_{t, 0}\log T}{T}\sqrt{\sum_{i}\big(\varepsilon_{\mathsf{score}, t, i}^{(0)}(x_{\tau_{t, i}}^{(0)})\big)^2}.
\end{align}

We denote
\begin{align*}
	\mathcal{E}_\ell^{\mathsf{typical}} := \{x_0: \|x_{\tau_{t,0}} - \sqrt{1-\tau_{t,0}}x_0\|_2 \leq 5\ell\sqrt{d\theta_t\tau_{t,0}\log T}\},~\quad~\ell = 1, 2, \dots
\end{align*}
Then for any $x_0 \in \mathcal{E}_\ell^{\text{typical}}$, we have
\begin{align}
	\frac{(\tau_{t,0} - \tau_{t, i})\left\|x_{\tau_{t,0}} - \sqrt{1-\tau_{t,0}}x_0\right\|_2^2}{2(1-\tau_{t,0})\tau_{t,0}\tau_{t,i}} \leq \frac{25\ell^2}{2}\frac{(\tau_{t,0} - \tau_{t, i})d\theta_t\log T}{(1-\tau_{t,0})\tau_{t,i}} &\stackrel{\text{Lemma}~\ref{lemma:learning_rate}}{\lesssim} \ell^2\frac{d\theta_t\log^2 T}{T},\label{ineq27}
\end{align}
\begin{align}
	&\left|\frac{\big\|u_{t,i}^{(1)}\big\|_2^2 + 2\big<u_{t,i}^{(1)}, x_{\tau_{t,0}} - \sqrt{1-\tau_{t,0}}x_0\big>}{2\frac{(1-\tau_{t,0})\tau_{t,i}}{1-\tau_{t,i}}}\right|\notag\\&\quad\leq \frac{\big\|u_{t,i}^{(1)}\big\|_2^2 + 2\big\|u_{t,i}^{(1)}\big\|_2\|x_{\tau_{t,0}} - \sqrt{1-\tau_{t,0}}x_0\|_2}{2\frac{(1-\tau_{t,0})\tau_{t,i}}{1-\tau_{t,i}}}\notag\\
	&\quad\stackrel{\eqref{ineq26}~\text{and Lemma}~\ref{lemma:learning_rate}}{\lesssim} \frac{d\theta_t\log^3 T}{T^2} + \frac{\tau_{t, 0}\log^2 T}{T^2}\sum_{i}\big(\varepsilon_{\mathsf{score}, t, i}^{(0)}(x_{\tau_{t, i}}^{(0)})\big)^2 + \ell\frac{d\theta_t\log^2 T}{T}\notag\\&\hspace{3cm} + \frac{\ell\sqrt{d\theta_t\tau_{t, 0}\log^3 T}}{T}\sqrt{\sum_{i}\big(\varepsilon_{\mathsf{score}, t, i}^{(0)}(x_{\tau_{t, i}}^{(0)})\big)^2}\notag\\
	&\quad\lesssim \ell\frac{d\theta_t\log^2 T}{T} + \frac{\ell\sqrt{d\theta_t\tau_{t, 0}\log^3 T}}{T}\sqrt{\sum_{i}\big(\varepsilon_{\mathsf{score}, t, i}^{(0)}(x_{\tau_{t, i}}^{(0)})\big)^2}.\label{ineq31}
\end{align}
Combining the previous two inequalities and the fact $\exp(-z) = 1 - z + O(z^2)$, we know that for any $x_0 \in \mathcal{E}_2^{\mathsf{typical}}$, 
\begin{align}\label{ineq32}
	&\exp\left(-\frac{(\tau_{t,0} - \tau_{t, i})\left\|x_{\tau_{t,0}} - \sqrt{1-\tau_{t,0}}x_0\right\|_2^2}{2(1-\tau_{t,0})\tau_{t,0}\tau_{t,i}} -\frac{\big\|u_{t,i}^{(1)}\big\|_2^2 + 2\big<u_{t,i}^{(1)}, x_{\tau_{t,0}} - \sqrt{1-\tau_{t,0}}x_0\big>}{2\frac{(1-\tau_{t,0})\tau_{t,i}}{1-\tau_{t,i}}}\right)\notag\\
	&\quad= 1 -\frac{(\tau_{t,0} - \tau_{t, i})\left\|x_{\tau_{t,0}} - \sqrt{1-\tau_{t,0}}x_0\right\|_2^2}{2(1-\tau_{t,0})\tau_{t,0}\tau_{t,i}} -\frac{\big\|u_{t,i}^{(1)}\big\|_2^2 + 2\big<u_{t,i}^{(1)}, x_{\tau_{t,0}} - \sqrt{1-\tau_{t,0}}x_0\big>}{2\frac{(1-\tau_{t,0})\tau_{t,i}}{1-\tau_{t,i}}}\notag\\&\qquad + O\left(\frac{d^2\theta_t^2\log^4 T}{T^2} + \frac{d\theta_t\tau_{t, 0}\log^3 T}{T^2}\sum_{i}\big(\varepsilon_{\mathsf{score}, t, i}^{(0)}(x_{\tau_{t, i}}^{(0)})\big)^2\right)\notag\\
	&\quad= 1 -\frac{(\tau_{t,0} - \tau_{t, i})\left\|x_{\tau_{t,0}} - \sqrt{1-\tau_{t,0}}x_0\right\|_2^2}{2(1-\tau_{t,0})\tau_{t,0}\tau_{t,i}} - \frac{2\big<u_{t,i}^{(1)}, x_{\tau_{t,0}} - \sqrt{1-\tau_{t,0}}x_0\big>}{2\frac{(1-\tau_{t,0})\tau_{t,i}}{1-\tau_{t,i}}}\notag\\&\qquad + O\left(\frac{d^2\theta_t^2\log^4 T}{T^2} + \frac{d\theta_t\tau_{t, 0}\log^3 T}{T^2}\sum_{i}\big(\varepsilon_{\mathsf{score}, t, i}^{(0)}(x_{\tau_{t, i}}^{(0)})\big)^2\right),
\end{align}
and for any $x_0 \in \mathcal{E}_\ell^{\mathsf{typical}}$, 
\begin{align}\label{ineq33}
	-\frac{(\tau_{t,0} - \tau_{t, i})\left\|x_{\tau_{t,0}} - \sqrt{1-\tau_{t,0}}x_0\right\|_2^2}{2(1-\tau_{t,0})\tau_{t,0}\tau_{t,i}} -\frac{\big\|u_{t,i}^{(1)}\big\|_2^2 + 2\big<u_{t,i}^{(1)}, x_{\tau_{t,0}} - \sqrt{1-\tau_{t,0}}x_0\big>}{2\frac{(1-\tau_{t,0})\tau_{t,i}}{1-\tau_{t,i}}} \leq \ell d\theta_t.
\end{align}

%Repeating similar arguments as in \citet[(42a)]{li2024sharp} yields
We make the observation that
\begin{align}
	&\int_{x_0}p_{X_0\mymid \overline{X}_{\tau_{t, 0}}}(x_0\mymid x_{\tau_{t,0}})\notag\\&\hspace{3cm}\cdot\exp\left(-\frac{(\tau_{t,0} - \tau_{t, i})\left\|x_{\tau_{t,0}} - \sqrt{1-\tau_{t,0}}x_0\right\|_2^2}{2(1-\tau_{t,0})\tau_{t,0}\tau_{t,i}} -\frac{\big\|u_{t,i}^{(1)}\big\|_2^2 + 2\big<u_{t,i}^{(1)}, x_{\tau_{t,0}} - \sqrt{1-\tau_{t,0}}x_0\big>}{2\frac{(1-\tau_{t,0})\tau_{t,i}}{1-\tau_{t,i}}}\right){\rm  d}x_0\notag\\
	&= \int_{x_0 \in \mathcal{E}_2^{\mathsf{typical}}}p_{X_0\mymid \overline{X}_{\tau_{t, 0}}}(x_0\mymid x_{\tau_{t,0}})\notag\\&\hspace{3cm}\cdot\exp\left(-\frac{(\tau_{t,0} - \tau_{t, i})\left\|x_{\tau_{t,0}} - \sqrt{1-\tau_{t,0}}x_0\right\|_2^2}{2(1-\tau_{t,0})\tau_{t,0}\tau_{t,i}} -\frac{\big\|u_{t,i}^{(1)}\big\|_2^2 + 2\big<u_{t,i}^{(1)}, x_{\tau_{t,0}} - \sqrt{1-\tau_{t,0}}x_0\big>}{2\frac{(1-\tau_{t,0})\tau_{t,i}}{1-\tau_{t,i}}}\right){\rm  d}x_0\notag\\
	&\quad + \sum_{\ell = 3}^{\infty}\int_{x_0 \in \mathcal{E}_\ell^{\mathsf{typical}}\backslash \mathcal{E}_{\ell-1}^{\mathsf{typical}}}p_{X_0\mymid \overline{X}_{\tau_{t, 0}}}(x_0\mymid x_{\tau_{t,0}})\notag\\&\hspace{2cm}\cdot\exp\left(-\frac{(\tau_{t,0} - \tau_{t, i})\left\|x_{\tau_{t,0}} - \sqrt{1-\tau_{t,0}}x_0\right\|_2^2}{2(1-\tau_{t,0})\tau_{t,0}\tau_{t,i}} -\frac{\big\|u_{t,i}^{(1)}\big\|_2^2 + 2\big<u_{t,i}^{(1)}, x_{\tau_{t,0}} - \sqrt{1-\tau_{t,0}}x_0\big>}{2\frac{(1-\tau_{t,0})\tau_{t,i}}{1-\tau_{t,i}}}\right){\rm  d}x_0\notag\\
	&\stackrel{\eqref{ineq32}~\text{and}~\eqref{ineq33}}{=} \int_{x_0 \in \mathcal{E}_2^{\mathsf{typical}}}p_{X_0\mymid \overline{X}_{\tau_{t, 0}}}(x_0\mymid x_{\tau_{t,0}})\notag\\&\hspace{3cm}\cdot\left(1 -\frac{(\tau_{t,0} - \tau_{t, i})\left\|x_{\tau_{t,0}} - \sqrt{1-\tau_{t,0}}x_0\right\|_2^2}{2(1-\tau_{t,0})\tau_{t,0}\tau_{t,i}} - \frac{2\big<u_{t,i}^{(1)}, x_{\tau_{t,0}} - \sqrt{1-\tau_{t,0}}x_0\big>}{2\frac{(1-\tau_{t,0})\tau_{t,i}}{1-\tau_{t,i}}}\right){\rm  d}x_0\notag\\
	&\quad + O\left(\frac{d^2\theta_t^2\log^4 T}{T^2} + \frac{d\theta_t\tau_{t, 0}\log^3 T}{T^2}\sum_{i}\big(\varepsilon_{\mathsf{score}, t, i}^{(0)}(x_{\tau_{t, i}}^{(0)})\big)^2\right)\notag\\
	&\quad + O\left(\sum_{\ell = 3}^{\infty}\int_{x_0 \in \mathcal{E}_\ell^{\mathsf{typical}}\backslash \mathcal{E}_{\ell-1}^{\mathsf{typical}}}p_{X_0\mymid \overline{X}_{\tau_{t, 0}}}(x_0\mymid x_{\tau_{t,0}}){\rm  d}x_0\cdot\exp\left(\ell d\theta_t\right)\right)\\&\stackrel{\eqref{ineq30}~\text{and Lemma}~\ref{lemma:tail_bound}}{=}\int_{x_0 \in \mathcal{E}_2^{\mathsf{typical}}}p_{X_0\mymid \overline{X}_{\tau_{t, 0}}}(x_0\mymid x_{\tau_{t,0}})\left(1 -\frac{(\tau_{t,0} - \tau_{t, i})\left\|x_{\tau_{t,0}} - \sqrt{1-\tau_{t,0}}x_0\right\|_2^2}{2(1-\tau_{t,0})\tau_{t,0}\tau_{t,i}}\right){\rm  d}x_0\notag\\&\quad -\int_{x_0 \in \mathcal{E}_2^{\mathsf{typical}}}p_{X_0\mymid \overline{X}_{\tau_{t, 0}}}(x_0\mymid x_{\tau_{t,0}})\frac{2\big<-\sqrt{1-\tau_{t, 0}}\int_{\tau_{t, 0}}^{\tau_{t,i}} \frac{1}{2(1-\tau)^{3/2}}s_{\tau}^{\star}(x_{\tau}^\star)\mathrm{d}\tau, x_{\tau_{t,0}} - \sqrt{1-\tau_{t,0}}x_0\big>}{2\frac{(1-\tau_{t,0})\tau_{t,i}}{1-\tau_{t,i}}}{\rm  d}x_0\notag\\
	&\quad+ O\left(\bigg(\frac{\tau_{t, 0}\log T}{T}\sqrt{\sum_{i}\big(\varepsilon_{\mathsf{score}, t, i}^{(0)}(x_{\tau_{t, i}}^{(0)})\big)^2} +  \sqrt{\frac{d\theta_t\tau_{t, 0}\log^3 T}{T^2}}\left(\frac{d\theta_t\log^2T}{T}\right)^K\bigg)\frac{\sqrt{d\theta_t\tau_{t,0}\log T}}{\frac{(1-\tau_{t,0})\tau_{t,i}}{1-\tau_{t,i}}}\right)\notag\\
	&\quad + O\left(\frac{d^2\theta_t^2\log^4 T}{T^2} + \frac{d\theta_t\tau_{t, 0}\log^3 T}{T^2}\sum_{i}\big(\varepsilon_{\mathsf{score}, t, i}^{(0)}(x_{\tau_{t, i}}^{(0)})\big)^2\right) + O\left(\sum_{\ell = 3}^{\infty}\exp\left(-(\ell-1)^2d\theta\log T\right)\cdot\exp\left(\ell d\theta_t\right)\right).\label{ineq63}
\end{align}
Furthermore, \eqref{ineq34}, \eqref{ineq27}, \eqref{ineq31} and Lemma~\ref{lemma:tail_bound} together imply that
\begin{align}
	&\text{RHS of}~\eqref{ineq63}\notag\\
	&=1 - \frac{(\tau_{t,0} - \tau_{t, i})\int_{x_0}p_{X_0\mymid \overline{X}_{\tau_{t, 0}}}(x_0\mymid x_{\tau_{t,0}})\left\|x_{\tau_{t,0}} - \sqrt{1-\tau_{t,0}}x_0\right\|_2^2{\rm  d}x_0}{2(1-\tau_{t,0})\tau_{t,0}\tau_{t,i}}\notag\\
	&\quad - \frac{2\big<-\sqrt{1-\tau_{t, 0}}\int_{\tau_{t, 0}}^{\tau_{t,i}} \frac{1}{2(1-\tau)^{3/2}}s_{\tau}^{\star}(x_{\tau}^\star)\mathrm{d}\tau, \int_{x_0}p_{X_0\mymid \overline{X}_{\tau_{t, 0}}}(x_0\mymid x_{\tau_{t,0}})(x_{\tau_{t,0}} - \sqrt{1-\tau_{t,0}}x_0){\rm  d}x_0\big>}{2\frac{(1-\tau_{t,0})\tau_{t,i}}{1-\tau_{t,i}}}\notag\\
	&\quad - \int_{x_0 \notin \mathcal{E}_2^{\mathsf{typical}}}p_{X_0\mymid \overline{X}_{\tau_{t, 0}}}(x_0\mymid x_{\tau_{t,0}}){\rm  d}x_0\notag\\&\quad + \sum_{\ell = 3}^{\infty}\int_{x_0 \in \mathcal{E}_\ell^{\mathsf{typical}}\backslash \mathcal{E}_{\ell-1}^{\mathsf{typical}}}p_{X_0\mymid \overline{X}_{\tau_{t, 0}}}(x_0\mymid x_{\tau_{t,0}})\frac{(\tau_{t,0} - \tau_{t, i})\left\|x_{\tau_{t,0}} - \sqrt{1-\tau_{t,0}}x_0\right\|_2^2}{2(1-\tau_{t,0})\tau_{t,0}\tau_{t,i}}{\rm  d}x_0\notag\\
	&\quad + O\Bigg(\bigg\|-\sqrt{1-\tau_{t, 0}}\int_{\tau_{t, 0}}^{\tau_{t,i}} \frac{1}{2(1-\tau)^{3/2}}s_{\tau}^{\star}(x_{\tau}^\star)\mathrm{d}\tau\bigg\|_2\notag\\&\hspace{2cm}\sum_{\ell = 3}^{\infty}\int_{x_0 \in \mathcal{E}_\ell^{\mathsf{typical}}\backslash \mathcal{E}_{\ell-1}^{\mathsf{typical}}}p_{X_0\mymid \overline{X}_{\tau_{t, 0}}}(x_0\mymid x_{\tau_{t,0}})\left|\frac{\big\|u_{t,i}^{(1)}\big\|_2^2 + 2\big<u_{t,i}^{(1)}, x_{\tau_{t,0}} - \sqrt{1-\tau_{t,0}}x_0\big>}{2\frac{(1-\tau_{t,0})\tau_{t,i}}{1-\tau_{t,i}}}\right|{\rm  d}x_0\Bigg)\notag\\
	&\quad + O\left(\frac{d^2\theta_t^2\log^4 T}{T^2} + \frac{\sqrt{d\theta_t\tau_{t, 0}\log^3 T}}{T}\sqrt{\sum_{i}\big(\varepsilon_{\mathsf{score}, t, i}^{(0)}(x_{\tau_{t, i}}^{(0)})\big)^2}\right)\notag\\
	&= 1 - \frac{(\tau_{t,0} - \tau_{t, i})\int_{x_0}p_{X_0\mymid \overline{X}_{\tau_{t, 0}}}(x_0\mymid x_{\tau_{t,0}})\left\|x_{\tau_{t,0}} - \sqrt{1-\tau_{t,0}}x_0\right\|_2^2{\rm  d}x_0}{2(1-\tau_{t,0})\tau_{t,0}\tau_{t,i}}\notag\\
	&\quad - \frac{2\big<-\sqrt{1-\tau_{t, 0}}\int_{\tau_{t, 0}}^{\tau_{t,i}} \frac{1}{2(1-\tau)^{3/2}}s_{\tau}^{\star}(x_{\tau}^\star)\mathrm{d}\tau, \int_{x_0}p_{X_0\mymid \overline{X}_{\tau_{t, 0}}}(x_0\mymid x_{\tau_{t,0}})(x_{\tau_{t,0}} - \sqrt{1-\tau_{t,0}}x_0){\rm  d}x_0\big>}{2\frac{(1-\tau_{t,0})\tau_{t,i}}{1-\tau_{t,i}}}\notag\\
	&\quad + O\left(\frac{d^2\theta_t^2\log^4 T}{T^2} + \frac{\sqrt{d\theta_t\tau_{t, 0}\log^3 T}}{T}\sqrt{\sum_{i}\big(\varepsilon_{\mathsf{score}, t, i}^{(0)}(x_{\tau_{t, i}}^{(0)})\big)^2}\right).\label{ineq36}
\end{align}
In addition, one can show that
\begin{align*}
	&\bigg\|2\sqrt{1-\tau_{t, 0}}\int_{\tau_{t, i}}^{\tau_{t,0}} \frac{1}{2(1-\tau)^{3/2}}s_{\tau}^{\star}(x_{\tau}^\star)\mathrm{d}\tau - \frac{\tau_{t, 0} - \tau_{t,i}}{1-\tau_{t, i}}s_{\tau_{t,0}}^{\star}(x_{\tau_{t,0}})\bigg\|_2\\
	&\quad\leq \bigg\|2\sqrt{1-\tau_{t, 0}}\int_{\tau_{t, i}}^{\tau_{t,0}} \bigg(\frac{1}{2(1-\tau)^{3/2}}s_{\tau}^{\star}(x_{\tau}^\star) - \frac{1}{2(1-\tau_{t,0})^{3/2}}s_{\tau_{t,0}}^{\star}(x_{\tau_{t,0}}^\star)\bigg)\mathrm{d}\tau\bigg\|_2\\
	&\qquad + \left(\frac{\tau_{t, 0} - \tau_{t,i}}{1-\tau_{t, i}} - \frac{\tau_{t, 0} - \tau_{t,i}}{1-\tau_{t, 0}}\right)\big\|s_{\tau_{t,0}}^{\star}(x_{\tau_{t,0}})\big\|_2\\
	&\quad\lesssim 2\sqrt{1-\tau_{t, 0}}\int_{\tau_{t, i}}^{\tau_{t,0}}(\tau_{t,0} - \tau)\sup_{\tau' \in [\tau, \tau_{t, 0}]}\sqrt{\frac{d\theta_t\log T}{\tau'(1-\tau')^3}}\frac{d\theta_t\log T}{\tau'(1-\tau')}\mathrm{d}\tau + \frac{(\tau_{t, 0} - \tau_{t,i})^2}{(1-\tau_{t, i})(1-\tau_{t, 0})}\sqrt{\frac{d\theta_t\log T}{\tau_{t, 0}}}\\
	&\quad \lesssim \sqrt{1-\tau_{t, 0}}(\tau_{t,0} - \tau_{t,i})^2\frac{d^{3/2}\theta_t^{3/2}\log^{3/2}T}{\tau_{t, 0}^{3/2}(1 - \tau_{t, 0})^{5/2}} + \frac{\tau_{t, 0} - \tau_{t,i}}{1-\tau_{t, i}}\frac{\tau_{t, 0} - \tau_{t,i}}{1-\tau_{t, 0}}\sqrt{\frac{d\theta_t\log T}{\tau_{t, 0}}}\\
	&\quad \lesssim \tau_{t, 0}^{1/2}\frac{d^{3/2}\theta_t^{3/2}\log^{7/2}T}{T^2} + \tau_{t, 0}^{3/2}\frac{d^{1/2}\theta_t^{1/2}\log^{5/2}T}{T^2} \asymp \tau_{t, 0}^{1/2}\frac{d^{3/2}\theta_t^{3/2}\log^{7/2}T}{T^2}.
\end{align*}
Here, the third line is valid due to \eqref{eq:derivative_1} and Lemma \ref{lemma:Hessian}; the penultimate line and the last line result from Lemma \ref{lemma:learning_rate}. Then we have
\begin{align}\label{ineq35}
	&\frac{\left|\big<-2\sqrt{1-\tau_{t, 0}}\int_{\tau_{t, 0}}^{\tau_{t,i}} \frac{1}{2(1-\tau)^{3/2}}s_{\tau}^{\star}(x_{\tau}^\star)\mathrm{d}\tau - \frac{\tau_{t, 0} - \tau_{t,i}}{1-\tau_{t, i}}s_{\tau_{t,0}}^{\star}(x_{\tau_{t,0}}), \int_{x_0}p_{X_0\mymid \overline{X}_{\tau_{t, 0}}}(x_0\mymid x_{\tau_{t,0}})(x_{\tau_{t,0}} - \sqrt{1-\tau_{t,0}}x_0){\rm  d}x_0\big>\right|}{2\frac{(1-\tau_{t,0})\tau_{t,i}}{1-\tau_{t,i}}}\notag\\
	&\quad \stackrel{\text{Lemma}~\ref{lemma:learning_rate}}{\lesssim} \frac{\big\|2\sqrt{1-\tau_{t, 0}}\int_{\tau_{t, i}}^{\tau_{t,0}} \frac{1}{2(1-\tau)^{3/2}}s_{\tau}^{\star}(x_{\tau}^\star)\mathrm{d}\tau - \frac{\tau_{t, 0} - \tau_{t,i}}{1-\tau_{t, i}}s_{\tau_{t,0}}^{\star}(x_{\tau_{t,0}})\big\|_2}{\tau_{t,0}}\notag\\&\hspace{2cm}\cdot\big\|\int_{x_0}p_{X_0\mymid \overline{X}_{\tau_{t, 0}}}(x_0\mymid x_{\tau_{t,0}})(x_{\tau_{t,0}} - \sqrt{1-\tau_{t,0}}x_0){\rm  d}x_0\big\|_2\notag\\
	&\quad \stackrel{\text{Lemma}~\ref{lemma:tail_bound}}{\lesssim} \frac{\tau_{t, 0}^{1/2}\frac{d^{3/2}\theta_t^{3/2}\log^{7/2}T}{T^2}\cdot \sqrt{\theta_td\tau_{t,0}\log T}}{\tau_{t, 0}}\notag\\
	&\quad = \frac{d^2\theta_t^2\log^4 T}{T^2}.
\end{align}
Combining \eqref{ineq63}, \eqref{ineq36} and \eqref{ineq35}, one has
\begin{align*}
	&\int_{x_0}p_{X_0\mymid \overline{X}_{\tau_{t, 0}}}(x_0\mymid x_{\tau_{t,0}})\\&\quad\cdot\exp\left(-\frac{(\tau_{t,0} - \tau_{t, i})\left\|x_{\tau_{t,0}} - \sqrt{1-\tau_{t,0}}x_0\right\|_2^2}{2(1-\tau_{t,0})\tau_{t,0}\tau_{t,i}} -\frac{\big\|u_{t,i}^{(1)}\big\|_2^2 + 2\big<u_{t,i}^{(1)}, x_{\tau_{t,0}} - \sqrt{1-\tau_{t,0}}x_0\big>}{2\frac{(1-\tau_{t,0})\tau_{t,i}}{1-\tau_{t,i}}}\right){\rm  d}x_0\\
	&\quad = 1 - \frac{(\tau_{t,0} - \tau_{t, i})\int_{x_0}p_{X_0\mymid \overline{X}_{\tau_{t, 0}}}(x_0\mymid x_{\tau_{t,0}})\left\|x_{\tau_{t,0}} - \sqrt{1-\tau_{t,0}}x_0\right\|_2^2{\rm  d}x_0}{2(1-\tau_{t,0})\tau_{t,0}\tau_{t,i}}\\
	&\qquad - \frac{\big<\frac{\tau_{t, 0} - \tau_{t,i}}{1-\tau_{t, i}}s_{\tau_{t,0}}^{\star}(x_{\tau_{t,0}}), \int_{x_0}p_{X_0\mymid \overline{X}_{\tau_{t, 0}}}(x_0\mymid x_{\tau_{t,0}})(x_{\tau_{t,0}} - \sqrt{1-\tau_{t,0}}x_0){\rm  d}x_0\big>}{2\frac{(1-\tau_{t,0})\tau_{t,i}}{1-\tau_{t,i}}}\\
	&\qquad + O\left(\frac{d^2\theta_t^2\log^4 T}{T^2} + \frac{\sqrt{d\theta_t\tau_{t, 0}\log^3 T}}{T}\sqrt{\sum_{i}\big(\varepsilon_{\mathsf{score}, t, i}^{(0)}(x_{\tau_{t, i}}^{(0)})\big)^2}\right)\\
	&\quad \stackrel{\eqref{eq:score_form}}{=} 1 - \frac{\tau_{t,0} - \tau_{t, i}}{2(1-\tau_{t,0})\tau_{t,0}\tau_{t,i}}\bigg[\int_{x_0}p_{X_0\mymid \overline{X}_{\tau_{t, 0}}}(x_0\mymid x_{\tau_{t,0}})\left\|x_{\tau_{t,0}} - \sqrt{1-\tau_{t,0}}x_0\right\|_2^2{\rm  d}x_0\\&\hspace{4.5cm} - \big\|\int_{x_0}p_{X_0\mymid \overline{X}_{\tau_{t, 0}}}(x_0\mymid x_{\tau_{t,0}})(x_{\tau_{t,0}} - \sqrt{1-\tau_{t,0}}x_0){\rm  d}x_0\big\|_2^2\bigg]\\&\qquad + O\left(\frac{d^2\theta_t^2\log^4 T}{T^2} + \frac{\sqrt{d\theta_t\tau_{t, 0}\log^3 T}}{T}\sqrt{\sum_{i}\big(\varepsilon_{\mathsf{score}, t, i}^{(0)}(x_{\tau_{t, i}}^{(0)})\big)^2}\right)\\
	&\quad \stackrel{\text{Jensen's inequality}}{\leq} 1 + O\left(\frac{d^2\theta_t^2\log^4 T}{T^2} + \frac{\sqrt{d\theta_t\tau_{t, 0}\log^3 T}}{T}\sqrt{\sum_{i}\big(\varepsilon_{\mathsf{score}, t, i}^{(0)}(x_{\tau_{t, i}}^{(0)})\big)^2}\right).
\end{align*}
Equation \eqref{ineq25} together with the previous inequality and Lemma \ref{lemma:learning_rate} implies that
\begin{align}\label{ineq28}
	\log\frac{p_{\sqrt{\frac{1-\tau_{t,0}}{1-\tau_{t,i}}}\overline{X}_{\tau_{t,i}}}\left(\sqrt{\frac{1-\tau_{t,0}}{1-\tau_{t,i}}}x_{\tau_{t,i}}^{(1)}\right)}{p_{\overline{X}_{\tau_{t,0}}}(x_{\tau_{t,0}})} \leq \frac{4c_1d\log T}{T} + C_{10}\left\{\frac{d^2\theta_t^2\log^4 T}{T^2} + \frac{\sqrt{d\theta_t\log^3 T}}{T}\sqrt{\sum_{i}\big(\varepsilon_{\mathsf{score}, t, i}^{(0)}(x_{\tau_{t, i}}^{(0)})\big)^2}\right\}.
\end{align}
Similarly, for any $\lambda \in [0,1]$, one has
\begin{align}\label{ineq29}
	&\left|\log\frac{p_{\sqrt{\frac{1-\tau_{t,0}}{1-\tau_{t,i}}}\overline{X}_{\tau_{t,i}}}\left(\sqrt{\frac{1-\tau_{t,0}}{1-\tau_{t,i}}}\left(\lambda x_{\tau_{t,i}}^\star + (1-\lambda)x_{\tau_{t,i}}^{(1)}\right)\right)}{p_{\overline{X}_{\tau_{t,0}}}(x_{\tau_{t,0}})}\right|\notag\\ \quad &\leq C_{10}\left\{\frac{d\theta_t\log^2 T}{T} + \frac{\ell\sqrt{d\theta_t\log^3 T}}{T}\sqrt{\sum_{i}\big(\varepsilon_{\mathsf{score}, t, i}^{(0)}(x_{\tau_{t, i}}^{(0)})\big)^2}\right\}
\end{align}
Repeating similar arguments and using induction yields that for all $0 \leq n \leq N-1$, $0 \leq i \leq K-1$,
\begin{subequations}
	\begin{align}
		-\log p_{\overline{X}_{\tau_{t,i}}}\big(\lambda x_{\tau_{t,i}}^{(n+1)} + (1-\lambda)x_{\tau_{t,i}}^\star\big) &\leq 2.1d\theta_t\log T,\label{ineq37a}\\
		\log\frac{p_{\sqrt{\frac{1-\tau_{t,0}}{1-\tau_{t,i}}}\overline{X}_{\tau_{t,i}}}\left(\sqrt{\frac{1-\tau_{t,0}}{1-\tau_{t,i}}}x_{\tau_{t,i}}^{(n+1)}\right)}{p_{\overline{X}_{\tau_{t,0}}}(x_{\tau_{t,0}})} &\leq \frac{4c_1d\log T}{T} + C_{10}\left\{\frac{d^2\theta_t^2\log^4 T}{T^2} + \frac{\sqrt{d\theta_t\log^3 T}}{T}\sqrt{\sum_{i}\big(\varepsilon_{\mathsf{score}, t, i}^{(n)}(x_{\tau_{t, i}}^{(n)})\big)^2}\right\}.\label{ineq37b}
	\end{align}
\end{subequations}

\subsection{Proof of Lemma \ref{lemma:log_density_bound}}\label{sec:proof_lemma_log_density_bound}
We prove Lemma \ref{lemma:log_density_bound} by contradiction. Suppose that there exists $\ell \in [1, \tau(x_T))$ such that $-\log q_{\ell}(x_{\ell}) > 2c_6d\log T$. We let $1 < t \leq \ell$ denote the smallest time step satisfying
\begin{align}
	\theta_t(x_t) = \max\left\{-\frac{\log q_t(x_t)}{d\log T}, c_6\right\} > 2c_6.\label{ineq53}
\end{align}
Since $-\log q_t(x_t) \leq c_6d\log T$, we have 
\begin{align}
	\theta_t(x_t) \geq 2\theta_1(x_1).\label{ineq51}
\end{align}
Morevoer, by repeating similar arguments as in \citet[Eqn. (129)]{li2024sharp}, one can derive
\begin{align}\label{ineq50}
	\theta_1(x_1), \dots, \theta_t(x_t) \in [c_6, 4c_6]. 
\end{align}
By virtue of Lemma \ref{lemma:log_density_ratio_1}, for any $2 \leq j \leq t$, one has
\begin{align}
	&\log q_{j-1}(x_{j-1}) - \log q_j(x_j)\notag\\ &\quad= \log\frac{p_{\overline{X}_{\tau_{j, K-1}}}\big(x_{\tau_{j, K-1}}^{(n)}\big)}{p_{\overline{X}_{\tau_{j, 0}}}\big(x_{\tau_{j, 0}}\big)}\notag\\ &\quad= \log\frac{p_{\sqrt{\frac{1 - \tau_{j,0}}{1 - \tau_{j,K-1}}}\overline{X}_{\tau_{j, K-1}}}\big(\sqrt{\frac{1 - \tau_{j,0}}{1 - \tau_{j,K-1}}}x_{\tau_{j, K-1}}^{(n)}\big)}{p_{\overline{X}_{\tau_{j, 0}}}\big(x_{\tau_{j, 0}}\big)} + \log\left[\left(\frac{1 - \tau_{j,0}}{1 - \tau_{j,K-1}}\right)^{d/2}\right]\notag\\
	&\quad\leq \frac{4c_1d\log T}{T} + C_{10}\left\{\frac{d^2\theta_t^2\log^4 T}{T^2} + \frac{\sqrt{d\theta_t\log^3 T}}{T}\sqrt{\sum_{i}\big(\varepsilon_{\mathsf{score}, t, i}^{(n)}(x_{\tau_{t, i}}^{(n)})\big)^2}\right\} + \frac{d}{2}\log\alpha_j\notag\\
	&\quad\leq \frac{4c_1d\log T}{T} + C_{10}\left\{\frac{d^2\theta_t^2\log^4 T}{T^2} + \frac{\sqrt{d\theta_t\log^3 T}}{T}\sqrt{\sum_{i}\big(\varepsilon_{\mathsf{score}, t, i}^{(n)}(x_{\tau_{t, i}}^{(n)})\big)^2}\right\},\label{ineq52}
\end{align}
where we makes use of the fact $p_{\overline{X}_{\tau_{j, K-1}}}\big(x_{\tau_{j, K-1}}^{(n)}\big)  = \left(\frac{1 - \tau_{j,0}}{1 - \tau_{j,K-1}}\right)^{d/2}p_{\sqrt{\frac{1 - \tau_{j,0}}{1 - \tau_{j,K-1}}}\overline{X}_{\tau_{j, K-1}}}\big(\sqrt{\frac{1 - \tau_{j,0}}{1 - \tau_{j,K-1}}}x_{\tau_{j, K-1}}^{(n)}\big)$ in the second line.
Putting \eqref{ineq53}, \eqref{ineq51}, \eqref{ineq50} and \eqref{ineq52} together leads to
\begin{align}
	c_6 = \theta_1(x_1) &\leq \theta_t(x_t) - \theta_1(x_1) = -\frac{\log q_t(x_t)}{d\log T} - \theta_1(x_1) \leq \frac{-\log q_t(x_t) + \log q_1(x_1)}{d\log T}\notag\\
	&= \frac{1}{d\log T}\sum_{j = 2}^{t}\left(\log q_{j-1}(x_{j-1}) - \log q_{j}(x_j)\right)\notag\\
	&\leq 4c_1 + C_{11}\left(\frac{d\log^3 T}{T} + \frac{S_{\tau(x_T)-1}(x_T)}{d\log T}\right)\notag\\ &\leq 5c_1.
\end{align}
Here, the last inequality holds due to \eqref{def:tau}. This contradicts our assumption $c_6 > 5c_1$. Therefore, we know that \eqref{ineq:log_density_bound} holds for all $1 \leq \ell < \tau(x_T)$.

\subsection{Proof of Lemma \ref{lemma:log_density_ratio_2}}\label{sec:proof_lemma_log_density_ratio_2}
First, we prove that the event $\{S_t(x_T) \leq c_3\}$ implies $\mathcal{E}_t \cap \{\xi_t(x_t) \leq c_3\}$ for all $t < \tau$. We know from Lemma \ref{lemma:log_density_bound} that 
\begin{align}
	-\log q_{\ell}(x_\ell) \leq 2c_6d\log T,~\quad~\forall 1 \leq \ell \leq t.\label{ineq61}
\end{align}
Then Lemma \ref{lemma:log_density} and Lemma \ref{lemma:learning_rate} (more precisely, \eqref{ineq:learning_rate_f}) together imply that for any $\tau \in [\tau_{t,K-1}, \tau_{t,0}]$, one has
\begin{align}
	-\log p_{\overline{X}_{\tau}}(x_{\tau}^\star) \leq 4c_6d\log T.\label{ineq54}
\end{align}
Furthermore, it is straightforward to verify that
\begin{align}
	&C_{10}\frac{\theta_{\tau_{t,0}}(x_{\tau_{t,0}})d\log^2 T + \sqrt{\theta_{\tau_{t,0}}(x_{\tau_{t,0}})\sum_{i,n}\big(\varepsilon_{\mathsf{score}, t, i}^{(n)}(x_{\tau_{t, i}}^{(n)})\big)^2d\log^3 T}}{T}\notag\\
	&\quad \asymp \frac{d\log^2 T + \sqrt{\sum_{i,n}\big(\varepsilon_{\mathsf{score}, t, i}^{(n)}(x_{\tau_{t, i}}^{(n)})\big)^2d\log^3 T}}{T}\notag\\
	&\quad \leq \frac{d\log^2 T}{T} + \xi_t(x_t)\notag\\
	&\quad \leq \frac{d\log^2 T}{T} + S_t(x_T)\notag\\
	&\quad \ll 1.\label{ineq55}
\end{align}
By virtue of Lemma \ref{lemma:log_density_ratio_1}, we know that for all $0 \leq i \leq K-1, 0 \leq n \leq N$,
\begin{align}
	-\log p_{\overline{X}_{\tau_{t,i}}}\big(\lambda x_{\tau_{t,i}}^{(n)} + (1-\lambda)x_{\tau_{t,i}}^\star\big) &\leq 2.1d\theta_t\log T.\label{ineq56}
\end{align}
In addition, for all $0 \leq i \leq K-1, 0 \leq n \leq N$, we have
\begin{align}
	\frac{\varepsilon_{\mathsf{Jacob}, t, i}^{(n)}\big(x_{\tau_{t, i}}^{(n)}\big)\log T}{T} \leq \xi_t(x_t) \leq S_t(x_T) \leq c_3 \ll 1. \label{ineq57}
\end{align}
Combining \eqref{ineq54}, \eqref{ineq56} and \eqref{ineq57} together yields
\begin{align}
	\{S_t(x_T) \leq c_3\} \subseteq \mathcal{E}_t \cap \{\xi_t(x_t) \leq c_3\}.\label{ineq58}
\end{align}
Observing that $S_t(x_T) \leq c_3$ for all $t < \tau$, \eqref{ineq42} tells us that 
\begin{align}
	\frac{p_{Y_{t-1}}(x_{t-1})}{p_{X_{t-1}}(x_{t-1})} = \exp\left(O\left(\xi_t(x_t) + d\bigg(\frac{d\log^2 T}{T}\bigg)^{K+1}\right)\right)\frac{p_{Y_{t}}(x_{t})}{p_{X_{t}}(x_{t})},~\quad~\forall 2 \leq t < \tau(x_T).\label{ineq59}
\end{align}
As a result, one has
\begin{align}
	\frac{q_1(x_1)}{p_1(x_1)} = \prod_{t=2}^{\tau-1}\exp\left(O\left(\xi_t(x_t) + d\bigg(\frac{d\log^2 T}{T}\bigg)^{K+1}\right)\right)\cdot \frac{q_{\tau-1}(x_{\tau-1})}{p_{\tau-1}(x_{\tau-1})}\notag\\
	= \exp\left(O\left(\sum_{t < \tau}\xi_t(x_t) + d^2\log^2 T\bigg(\frac{d\log^2 T}{T}\bigg)^{K}\right)\right)\frac{q_{\tau-1}(x_{\tau-1})}{p_{\tau-1}(x_{\tau-1})}\notag\\
	= \left(1 + O\left(\sum_{t < \tau}\xi_t(x_t) + d^2\log^2 T\bigg(\frac{d\log^2 T}{T}\bigg)^{K}\right)\right)\frac{q_{\tau-1}(x_{\tau-1})}{p_{\tau-1}(x_{\tau-1})},\label{ineq60}
\end{align}
which has finished the proof of \eqref{ineq45a}. Here, the last equation holds since $\sum_{t < \tau}\xi_t(x_t) + d^2\log^2 T\bigg(\frac{d\log^2 T}{T}\bigg)^{K} \leq c_3 + d^2\log^2 T\bigg(\frac{d\log^2 T}{T}\bigg)^{K} \ll 1$ and $\exp(z) = 1 + O(z)$ for all $|z| < 1$. Similarly, we can also prove \eqref{ineq45b}.

\subsection{Proof of Lemma \ref{lemma:other_set}}\label{sec:proof_Lemma_other_set}
We can prove Lemma \ref{lemma:other_set} by using same arguments in the proof of \citet[Lemma 8]{li2024sharp} and we omit the details here for the sake of brevity.

\bibliographystyle{apalike}
\bibliography{reference}

\end{document}